\documentclass[opre,final]{informs3}

\DoubleSpacedXI
\usepackage{endnotes}
\let\footnote=\endnote

\usepackage{natbib}
 \bibpunct[, ]{(}{)}{,}{a}{}{,}%
 \def\bibfont{\small}%
 \TheoremsNumberedThrough
\ECRepeatTheorems
\EquationsNumberedThrough

\usepackage{microtype}
\usepackage{graphicx}
\usepackage{mathtools}
\usepackage{subcaption}
\usepackage{booktabs}
\usepackage{bm}
\usepackage{multicol,multirow}
\usepackage{comment}
\usepackage{booktabs}
\newcommand{\ie}{\textit{i.e.}}
\newcommand{\eg}{\textit{e.g.}}

\usepackage{enumerate}   
\usepackage{amsmath}
\usepackage{tikz}
\usetikzlibrary{positioning,arrows,decorations,decorations.markings,shadows,positioning,arrows.meta,matrix,fit}

\usepackage{amssymb}

\usepackage[capitalise]{cleveref}
\Crefname{assumption}{Assumption}{Assumptions}
\Crefname{appsec}{Appendix}{Appendices}

\usepackage{xr}
\DeclareMathOperator*{\plim}{plim}

\newcommand{\bh}{\mathbf{H}}
\newcommand{\smallo}{{\scriptscriptstyle\mathcal{O}}} %

\newcommand{\bG}{\mathbb{G}}

\newcommand{\Fcal}{\mathcal{F}}
\newcommand{\cm}{\mathcal{M}}
\newcommand{\dm}{\mathcal{D}}
\newcommand{\Tcal}{\mathcal{T}}
\newcommand{\Dcal}{\mathcal{D}}
\newcommand{\ch}{\mathcal{H}}
\newcommand{\ct}{\mathcal{T}}
\newcommand{\cj}{\mathcal{J}}
\newcommand{\bigO}{\mathcal{O}} %

\renewcommand{\P}{\mathbb{P}}
\newcommand{\var}{\mathrm{var}}
\newcommand{\op}{\mathrm{o}_{p}}
\newcommand{\Op}{\bigO_{p}}
\newcommand{\E}{\mathbb{E}}

\newcommand{\rE}{\mathrm{E}}

\newcommand{\prns}[1]{\left(#1\right)}
\newcommand{\magd}[1]{\left\|#1\right\|}
\newcommand{\braces}[1]{\left\{#1\right\}}
\newcommand{\bracks}[1]{\left[#1\right]}
\newcommand{\sumT}{\sum_{t=0}^T}
\newcommand{\abs}[1]{\left|#1\right|}
\newcommand{\Rl}{\mathbb{R}}

\newcommand{\epol}{\pi_{e}}
\newcommand{\bpol}{\pi_{b}}

\newcommand{\RN}[1]{%
  \textup{\uppercase\expandafter{\romannumeral#1}}%
}

\usepackage{graphicx}

\def\boxit#1{\vbox{\hrule\hbox{\vrule\kern6pt\vbox{\kern6pt#1\kern6pt}\kern6pt\vrule}\hrule}}

\usepackage{xcolor}
\def\blockedit{}
\def\newedit{}
\def\neweditf{}
\def\edit{}

\newenvironment{myproof}[1][Proof]{\proof{#1}}{\endproof}
\newcommand{\qedhere}{}

\definecolor{mygreen}{rgb}{0,0.6,0}

\begin{document}

\RUNAUTHOR{Kallus and Uehara}
\RUNTITLE{\emph{Efficiently} Breaking the Curse of Horizon}
\TITLE{\emph{Efficiently} Breaking the Curse of Horizon in Off-Policy Evaluation with Double Reinforcement Learning}
\ARTICLEAUTHORS{%
\AUTHOR{Nathan Kallus}
\AFF{Cornell University, New York, NY 10044, \EMAIL{kallus@cornell.edu}}
\AUTHOR{Masatoshi Uehara}
\AFF{Cornell University, New York, NY 10044, \EMAIL{mu223@cornell.edu}}
}

\ABSTRACT{Off-policy evaluation (OPE) in reinforcement learning is notoriously difficult in long- and infinite-horizon settings due to diminishing overlap between behavior and target policies.
In this paper, we study the role of Markovian and time-invariant structure in efficient OPE. 
We first derive the efficiency {\newedit bounds and efficient influence functions} for OPE when one assumes each of these structures.
This precisely characterizes the curse of horizon: in time-variant processes, OPE is only feasible in the near-on-policy setting, where behavior and target policies are sufficiently similar.
But, in time-invariant Markov decision processes, our bounds show that truly-off-policy evaluation is feasible, even with only just one dependent trajectory, and provide the limits of how well we could hope to do. We develop a new estimator based on Double Reinforcement Learning (DRL) that leverages this structure for OPE. 
Our DRL estimator simultaneously uses estimated stationary density ratios and $q$-functions and remains efficient when both are estimated at slow, nonparametric rates and remains consistent when either is estimated consistently. We investigate these properties and the performance benefits of leveraging the problem structure for more efficient OPE.
}


\maketitle

\section{Introduction}

{\neweditf Reinforcement learning (RL) in settings such as healthcare \citep{MurphyS.A.2003Odtr} and education \citep{Mandel2014} is often limited to the offline or off-policy setting, where we only use existing observed data, due to the inability to simulate and the costliness of exploration. 
One important task in this setting is off-policy evaluation (OPE), where we want to estimate the mean reward of a candidate decision policy, known as the target policy using observed data generated by the log of another policy, known as the behavior policy \citep{Precup2000,Mahmood20014,Li2015,thomas2016,jiang,Munos2016,NIPS2018_7530,XieTengyang2019OOEf}.\footnote{\neweditf OPE can also sometimes refer to estimating the whole value or quality function of a policy; here he focus on estimating the mean reward.}
OPE, in particular, is a building block toward policy optimization from observational data \citep{huang2020importance,kallus2020statistically}.
OPE, however, becomes increasingly difficult for problems with long and infinitely-long horizons \citep{Liu2018}. As the horizon grows, the overlap (\ie, density ratios) between trajectories generated by the target and behavior policies diminishes exponentially. This issue has in particular been noted as one of the key limitations for the applicability of RL in medical settings \citep{gottesman2019guidelines}.
}

\begin{table}[t!]%
\SingleSpacedXI\centering\def\arraystretch{2}%
\begin{tabular}{cccc}\toprule
Model &  Characteristics & MSE Scaling &  {\newedit Required Conditions} \\\midrule
\edit{NDMP}  & Non-Markov, Time-variant & $\mathcal{O}(1/N)$  & \parbox[m]{4.65cm}{\centering$N\to\infty,\,T=\omega(\log N),$\\$\magd{\nu_{t}}_\infty=\smallo(\gamma^{-t})$} \\
{\edit{TMDP}}  & 
{Markov, Time-variant} &$\mathcal{O}(1/N)$  &\parbox[m]{4.65cm}{\centering$N\to\infty,\,T=\omega(\log N),$\\$\magd{\mu_{t}}_\infty=\smallo(\gamma^{-t})$}  \\[1em]
 \edit{MDP}  &  Markov, Time-invariant & $\mathcal{O}(1/(NT))$   &  
 \parbox[m]{4.65cm}{\centering$T\to\infty,\,N\geq1,$\\Mixing, {\newedit $\|w\|_{\infty}=O(1)$}}
 \\\bottomrule
\end{tabular}%
\vspace{1em}\captionof{table}{Asymptotic order of the best-achievable MSE in each model when observing $N$ length-$(T+1)$ trajectories. The variables $\eta_t,\nu_t,\,\mu_t,\,w$ are the instantaneous, cumulative, marginal, {\newedit and stationary density ratios, respectively (see \cref{sec: problem setup} for definitions).}}%
\label{tab:my_label}%
\centering%
\begin{minipage}[m]{0.75\textwidth}\scalebox{0.85}{\begin{tikzpicture}[%
>=latex',node distance=2cm, minimum height=0.75cm, minimum width=0.75cm,
state/.style={draw, shape=circle, draw=mygreen, fill=mygreen!10, line width=0.5pt},
action/.style={draw, shape=rectangle, draw=red, fill=red!10, line width=0.5pt},
reward/.style={draw, shape=rectangle, draw=blue, fill=blue!10, line width=0.5pt}
]
\node[state] (S0) at (0,0) {$s_0$};
\node[action,right of=S0] (A0) {$a_0$};
\node[reward,right of=A0] (R0) {$r_0$};
\node[state,right of=R0] (S1) {$s_1$};
\node[action,right of=S1] (A1) {$a_1$};
\node[reward,right of=A1] (R1) {$r_1$};
\node[state,right of=R1] (S2) {$s_2$};
\draw[->] (S0) -- (A0);
\draw[->] (S0) edge[bend left=30] (R0);
\draw[->] (A0) -- (R0);
\draw[->] (S0) edge[bend left=30] (S1);
\draw[->] (A0) edge[bend left=30] (S1);
\draw[->] (S1) -- (A1);
\draw[->] (S1) edge[bend left=30] (R1);
\draw[->] (A1) -- (R1);
\draw[->] (S1) edge[bend left=30] (S2);
\draw[->] (A1) edge[bend left=30] (S2);
\draw[->] (S0) edge[bend left=30] (A1);
\draw[->] (S0) edge[bend left=30] (R1);
\draw[->] (A0) edge[bend left=30] (R1);
\draw[->] (S0) edge[bend left=30] (S2);
\draw[->] (A0) edge[bend left=30] (S2);
\end{tikzpicture}}\end{minipage}\begin{minipage}[t]{0.245\textwidth}\captionof{figure}{\edit{NMDP}}\end{minipage}
\begin{minipage}[m]{0.75\textwidth}\scalebox{0.85}{\begin{tikzpicture}[%
>=latex',node distance=2cm, minimum height=0.75cm, minimum width=0.75cm,
state/.style={draw, shape=circle, draw=mygreen, fill=mygreen!10, line width=0.5pt},
action/.style={draw, shape=rectangle, draw=red, fill=red!10, line width=0.5pt},
reward/.style={draw, shape=rectangle, draw=blue, fill=blue!10, line width=0.5pt}
]
\node[state] (S0) at (0,0) {$s_0$};
\node[action,right of=S0] (A0) {$a_0$};
\node[reward,right of=A0] (R0) {$r_0$};
\node[state,right of=R0] (S1) {$s_1$};
\node[action,right of=S1] (A1) {$a_1$};
\node[reward,right of=A1] (R1) {$r_1$};
\node[state,right of=R1] (S2) {$s_2$};
\draw[->] (S0) -- (A0);
\draw[->] (S0) edge[bend left=30] (R0);
\draw[->] (A0) -- (R0);
\draw[->] (S0) edge[bend left=30] (S1);
\draw[->] (A0) edge[bend left=30] (S1);
\draw[->] (S1) -- (A1);
\draw[->] (S1) edge[bend left=30] (R1);
\draw[->] (A1) -- (R1);
\draw[->] (S1) edge[bend left=30] (S2);
\draw[->] (A1) edge[bend left=30] (S2);
\end{tikzpicture}}\end{minipage}\begin{minipage}[t]{0.245\textwidth}\captionof{figure}{\edit{TMDP}}\end{minipage}
\begin{minipage}[m]{0.75\textwidth}\scalebox{0.85}{\begin{tikzpicture}[%
>=latex',node distance=2cm, minimum height=0.75cm, minimum width=0.75cm,
state/.style={draw, shape=circle, draw=mygreen, fill=mygreen!10, line width=0.5pt},
action/.style={draw, shape=rectangle, draw=red, fill=red!10, line width=0.5pt},
reward/.style={draw, shape=rectangle, draw=blue, fill=blue!10, line width=0.5pt}
]
\node[state] (S0) at (0,0) {$s_0$};
\node[action,right of=S0] (A0) {$a_0$};
\node[reward,right of=A0] (R0) {$r_0$};
\node[state,right of=R0] (S1) {$s_1$};
\node[action,right of=S1] (A1) {$a_1$};
\node[reward,right of=A1] (R1) {$r_1$};
\node[state,right of=R1] (S2) {$s_2$};
\draw[->] (S0) -- (A0);
\draw[->] (S0) edge[bend left=30] (R0);
\draw[->] (A0) -- (R0);
\draw[->] (S0) edge[bend left=30] (S1);
\draw[->] (A0) edge[bend left=30] (S1);
\draw[->] (S1) -- (A1);
\draw[->] (S1) edge[bend left=30] (R1);
\draw[->] (A1) -- (R1);
\draw[->] (S1) edge[bend left=30] (S2);
\draw[->] (A1) edge[bend left=30] (S2);
\draw[draw,decoration={markings, mark=at position .5  with {
\node[transform shape,rotate=-45] {\footnotesize$|$};
}}] 
(S0) edge[bend left=30,decorate,color=mygreen] (S1)
(A0) edge[bend left=30,decorate,color=mygreen] (S1)
(S1) edge[bend left=30,decorate,color=mygreen] (S2)
(A1) edge[bend left=30,decorate,color=mygreen] (S2);
\draw[draw,decoration={markings, mark=at position .5  with {
\node[transform shape,rotate=-45] {\footnotesize$|\hspace{-.5em}|$};
}}] 
(S0) edge[bend left=30,decorate,color=blue] (R0)
(A0) edge[decorate,color=blue] (R0)
(S1) edge[bend left=30,decorate,color=blue] (R1)
(A1) edge[decorate,color=blue] (R1);
\draw[draw,decoration={markings, mark=at position .5  with {
\node[transform shape,rotate=-45] {\footnotesize$|\hspace{-.5em}|\hspace{-.5em}|$};
}}] 
(S0) edge[decorate,color=red] (A0)
(S1) edge[decorate,color=red] (A1);
\end{tikzpicture}}\end{minipage}\begin{minipage}[t]{0.245\textwidth}\captionof{figure}{\edit{MDP}}\end{minipage}
\vspace{1em}\captionof{figure}{Bayes net representation of the independence structure of the truncated trajectory ending with $s_2$, $\cj_{s_2}$, under the three models: NDMP, TMDP, and MDP. Conditional on its parents, a node is independent of all other nodes. The congruency sign \protect\rotatebox[origin=c]{-45}{$|\hspace{-.5em}|$} indicates that the conditional probability function given parent nodes is equal.}\label{fig:models}
\end{table}

In this paper we study the fundamental estimation limits for OPE in infinite-horizon settings, and we develop new estimators that leverage special problem structures to achieve these limits and enable efficient and effective OPE in these problem settings. Specifically, we first derive what is the \emph{best-possible} asymptotic mean squared error (MSE) that one can hope for in OPE in this setting, that is, {\neweditf we derive the efficiency bounds \citep{VaartA.W.vander1998As}, which characterize the minimum limit of the square-root-scaled MSE (as we define in \cref{sec:semiparam}).} 
In order to study the effect of problem structure, we separately consider three different models: non-Markov decision processes (NMDP), time-varying Markov decision processes (TMDP), and time-invariant Markov decision processes (MDP). These models are illustrated in \cref{fig:models} and precisely defined \cref{sec: problem setup}.
{\neweditf Specifically, we focus on discounted bounded rewards.}
The differences between these bounds exactly characterize the effect of taking into consideration additional problem structure on the feasibility of OPE.

Our bounds in the NMDP and TMDP models reveal an important phase transition: if the target and behavior policies are sufficiently similar (relative to the discount factor) then consistent estimation is feasible. Otherwise, there exist examples where it is infeasible. This can be understood as a phrase transition between being sufficiently close to on-policy that OPE is feasible even in infinite horizons and being sufficiently off-policy that it is hopeless. We show that adaptations of the doubly robust (DR) estimator in NMDPs \citep{jiang} and in MDPs \citep{NathanUehara2019} to the infinite horizon case achieve these bounds, \ie, are efficient in the near-on-policy setting.

Our bounds in the MDP models, on the other hand, give hope for OPE in the truly off-policy setting. They show that by leveraging Markovian and time-invariant structure in RL problems, we can overcome the \emph{curse of horizon} and indicate what it would mean to do so \emph{efficiently}, \ie, using all the data available optimally. The question is then how to achieve these bounds for efficient OPE. We propose an approach based on double reinforcement learning \citep{NathanUehara2019} and on simultaneously learning average visitation distributions and $q$-functions. And, we show that, unlike importance-sampling-based estimators \citep{Liu2018}, our DRL estimator achieves the efficiency bound under certain mixing conditions. Thus, by carefully leveraging problem structure we show how to \emph{efficiently} break the curse of horizon in RL OPE.

\subsection{Organization}
The organization of papers is as follows. 
In \cref{sec: problem setup}, we define the decision process models and set up the OPE problem formally.
In \cref{sec:semiparam}, we define the efficiency bounds formally, briefly reviewing semiparametric inference as it relates to our results.
In \cref{sec:litope}, we review the relevant literature on OPE.

In \cref{sec:timevariantmdp}, we derive the efficiency bounds under each of the models under consideration, NMDP, TMDP, and MDP. In \cref{sec:m1and2}, we analyze the asymptotic properties when we extend standard DR and DRL OPE estimators to infinite horizons and provide conditions for their efficiency in the NMDP and TMDP models. We note, however, that they are not efficient under the MDP model and have the wrong MSE scaling.
In \cref{sec:dm_infinite}, we propose the first efficient estimator for OPE under the MDP model and analyze its asymptotic properties as $T\to\infty$, including when our observations consist of a single trajectory, $N=1$. This estimator is based on simultaneously learning $q$-functions and the ratio of average visitation distributions. In \cref{sec:density}, we therefore discuss how to estimate the density ratio of average visitation distributions in an off-policy manner from a single (finite) trajectory. And, in \cref{sec:q-learning}, we discuss how to estimate $q$-functions in an off-policy manner from a single (finite) trajectory. In \cref{sec:numerical}, we provide a numerical experiment to study the effects of leveraging problem structure efficiently.  Finally, we conclude in \cref{sec: conclusions}.

\subsection{Problem Setup and Notation}\label{sec: problem setup}

\edit{We consider a state space $\mathcal{S}$, action space $\mathcal{A}$, and reward space $\mathcal R\subset[0,R_{\mathrm{max}}]$, each a measurable space that may be continuous, discrete, or mixed{\neweditf.\footnote{\neweditf While in control settings one often needs to restrict to standard Borel measurable spaces due to measurability issues when optimizing \citep{hernandez2012discrete}, since we are only considering evaluation, we do not require such a restriction.}}
We fix a base measure for each, $\lambda_{\mathcal S},\lambda_{\mathcal A},\lambda_{\mathcal R}$ (\eg, Lebesgue, counting, or other), focus on distributions on these spaces that are absolutely continuous with respect to (wrt) these, and identify them with their densities (Radon-Nikodym derivative wrt the base measure).
A (time-invariant) Markov decision process (MDP) on $(\mathcal S,\mathcal A,\mathcal R)$ is given by a reward distribution $p(r\mid s,a)$ for the immediate reward after taking action $a$ in state $s$ and a transition distribution $p(s'\mid s,a)$ for the new state after taking action $a$ in state $s$.
A policy is a distribution $\pi(a\mid s)$ for the action to take in state $s$. We also associate with $\pi$ an initial state distribution, $p^{(0)}_\pi(s_0)${\neweditf.\footnote{\neweditf In greatest generality, MDPs need not restrict the conditional new state $s'$ distributions to be absolutely continuous with respect to the same base measure for all state-action pairs $s,a$. And, the same for reward and policy distributions. We make this restriction here to be able to easily consider perturbations to the MDP distributions in a semiparametric framework.}} 
{\newedit Recall we identify distributions with densities so $p(r\mid s,a),p(s'\mid s,a),\pi(a\mid s),p^{(0)}_\pi(s_0)$ are densities with respect to $\lambda_{\mathcal R},\lambda_{\mathcal S},\lambda_{\mathcal A},\lambda_{\mathcal S}$, respectively.}
{\neweditf Together, an MDP and a policy define a joint distribution over trajectories $\cj=(s_0,a_0,r_0,s_1,a_1,r_1,\cdots)$. Namely, letting $\cj_{s_{T+1}}=(s_0,a_0,r_0,\cdots,s_{T},a_{T},r_{T},s_{T+1})$ be the length-$(T+1)$ trajectory up to $s_{T+1}$, we have that for any $T$, $\cj_{s_{T+1}}$ has density $p^{(0)}_\pi(s_0)\pi(a_0\mid s_0)p(r_0\mid s_0,a_0)p(s_1\mid s_0,a_0)\pi(a_1\mid s_1)p(r_1\mid s_1,a_1)\cdots p(s_{T+1}\mid s_T,a_T)$. We also define $\ch_{s_{T+1}}=(s_0,a_0,\cdots,s_{T},a_{T},s_{T+1})$ as the same length-$(T+1)$ trajectory but excluding reward variables, which has density $p^{(0)}_\pi(s_0)\pi(a_0\mid s_0)p(s_1\mid s_0,a_0)\cdots p(s_{T+1}\mid s_T,a_T)$, and we similarly denote by $\ch_{a_{T}}$ the trajectory up to and including the variable $a_T$, excluding rewards. (We formally define MDP as a statistical model for the data-generating process in \cref{def:mdpmodel}.)}
We denote by $p_\pi^{(t)}(s_t)$ or $p_\pi^{(t)}(s_t,a_t,r_t,s_{t+1})$ the marginal distribution of $s_t$ or of $(s_t,a_t,r_t,s_{t+1})$ (etc.) under $p_\pi$.}
\edit{We further define the $\gamma$-discounted average visitation frequency as}
$$\edit{
p^{(\infty)}_{\pi,\gamma}(s)=\lim_{T\to \infty}\frac1{\sum_{t=0}^{T}\gamma^{t}}\sum_{t=0}^{T}\gamma^{t}p_{\pi}^{(t)}(s).}
$$

Our ultimate goal is to estimate the average cumulative reward of the known target evaluation policy (and known initial state distribution), $\epol$, \edit{for a given discount factor $\gamma\in [0,1)$}:
$$
\rho^{\epol}=\lim_{T\to\infty}\,\rho^{\epol}_{T},\quad\text{where}\quad \rho^{\pi}_{T}=c_T(\gamma)\;\rE_{p_\pi}\bracks{\sumT \gamma^{t} r_t},\  c_T(\gamma)
=\prns{\sum_{t=0}^{T}\gamma^{t}}^{-1}.$$
\edit{In particular, we wish to estimate $\rho^{\epol}$ based on data generated by a \emph{different} policy, $\bpol$, known as the behavior policy and which may be known or unknown. In this work, we assume an evaluation policy $\epol$ and the initial distribution $p^{(0)}_{\epol}(s_0)$ we want to evaluate are known. 
(For brevity, we often use the subscript $e$ or $b$ to mean the subscript $\epol$ or $\bpol$, respectively.)}

\edit{We will consider two data-generation settings.
\paragraph*{\bf{Transition-sampling setting.}}\label{para:tran} In the \emph{transition-sampling setting}, the data consists of $n$ independent and identically distributed (iid) draws of state-action-reward-state quadruplets, $\mathcal{D}=\{(s^{\langle i \rangle },a^{\langle i \rangle },r^{\langle i \rangle },s'^{\langle i \rangle })\}_{i=1}^n$, each drawn from $p_{\pi_b}(\cj_{s_{1}})$. {\newedit Note we do \emph{not} assume stationarity in this setting, \ie,  the marginal densities of $p_{\pi_b}(\cj_{s_{1}})$ wrt $s$ and wrt $s'$ can be different. We denote the marginal distribution over $s$ by $p^{(0)}_{\pi_b}(s)$. }
\paragraph*{\bf{Trajectory-sampling setting.}} 
In the \emph{trajectory-sampling setting}, the data consists of $N$ observations of length-$(T+1)$ trajectories, $\mathcal{D}=\{(\cj_{s_{T+1}}^{\langle i \rangle }\}_{i=1}^{N}$, each drawn from $p_{\pi_b}(\cj_{s_{T+1}})$.
Here, we set $n=NT$ as we have $n$ transitions, and also identify $\mathcal{D}=\{(s^{\langle j \rangle }_t,a^{\langle j \rangle }_t,r^{\langle j \rangle }_t,s'^{\langle j \rangle }_{t+1})\}_{j=1,t=0}^{N,T}$.
Crucially, in this setting the transitions may be \emph{dependent}.
{\newedit Unlike the transition-sampling setting,} here we assume that the data are stationary: $p_{\pi_b}^{(t)}=p_{\pi_b}^{(t')}$ for any $t,t'$. That is, $p_{\pi_b}^{(0)}(s)$ is an invariant distribution under the state-transition kernel induced by the MDP and $\bpol$. This appears strong but can be easily relaxed if assume certain ergodicity so that the initial distribution does not in fact matter and we can allow any $p_{\pi_b}^{(0)}(s)$; we discuss this in \cref{remark:stationarity}.} %

\edit{The quality and value functions ($q$- and $v$-functions) are defined as the following conditional averages of the cumulative reward to go (under $\epol$), respectively:
\begin{align*}
q(s_0,a_0)=\rE_{p_{\epol}}\bracks{\sum_{k=0}^\infty \gamma^{k} r_k\mid s_0,a_0},\qquad
v(s_0)=\rE_{p_{\epol}}\bracks{\sum_{k=0}^\infty \gamma^{k} r_k\mid s_0}=\rE_{a \sim p_{\epol}(s_0)}\bracks{q(a,s_0)\mid s_0}.
\end{align*}
Note that the very last expectation is taken only over $a_0\sim\epol(a_0\mid s_0)$.
We define the policy, cumulative, marginal, and stationary density ratios, respectively, as
$$
\eta(s,a)=\frac{\epol(a\mid s)}{\bpol(a\mid s)},\quad \nu_t(\cj_{a_{t}})=\frac{p^{(0)}_{\epol}(s_0)}{p^{(0)}_{\pi^b}(s_0)}\prod_{k=0}^t\eta_k(s_{k},a_{k}),\quad \mu_t(s_t,a_t)=\frac{p_{\pi_e}^{(t)}(s_t,a_t)}{p_{\pi_b}^{(t)}(s_t,a_t)},\quad w(s)=\frac{p^{(\infty)}_{\epol,\gamma}(s)}{p^{(0)}_{\bpol}(s)}.
$$
We add two remarks. Firstly, $\nu_t$ includes $\frac{p^{(0)}_{\epol}(s_0)}{p^{(0)}_{\pi^b}(s_0)}$ to take the difference of initial distributions into account. Secondly, in $w(s)$, notice that we divide a $\gamma$-discounted average visitation frequency by an \emph{undiscounted} marginal one.
(In \cref{remark:stationarity} we discuss assuming ergodicity instead of stationarity in the trajectory-sampling setting, in which we case we replace the denominator of $w(s)$ with the undiscounted stationary state distribution under $p_{\bpol}(\cj)$.)
}

\edit{We can generalize the MDP setting in two ways. In TMDP, the reward, transition, and policy distributions can all depend on $t$, while the Markov assumption is still retained. Adding a $t$ subscript to denote this, under TMDP, $p_\pi$ is given by $p^{(0)}_\pi(s_0)\pi_0(a_0\mid s_0)p_0(r_0\mid s_0,a_0)p_1(s_1\mid s_0,a_0)\cdots$. In NMDP, the reward, transition, and policy distributions can additionally all depend on the history of states and actions so that $p_\pi$ is given by $p^{(0)}_\pi(s_0)\pi_0(a_0\mid s_0)p_0(r_0\mid s_0,a_0)p_1(s_1\mid s_0,a_0)\pi_1(a_1\mid \cj_{s_1})p_1(r_1\mid \cj_{a_1})\cdots$. 
In TMDP, $q$- and $v$-functions depend on $t$ and are defined as the conditional expectations of $\sum_{k=0}^\infty \gamma^{k} r_{k+t}$ given $s_t,a_t$ and $s_t$, respectively, under $p_{\pi_e}$. {\newedit In NMDP, we condition instead on $\cj_{a_t}$ and $\cj_{s_t}$, respectively.} In both TMDP and NDMP, $\eta$ is also $t$-dependent since the policies are. We only consider the trajectory-sampling setting under either TMDP and NDMP since, due to the time dependence, just observing length-1 trajectories would not be enough. {\neweditf(We formally define TMDP and NMDP as a statistical model for the data-generating process in \cref{def:nmdpmodel,def:tmdpmodel}.)}} %

\edit{To streamline notation, when no subscript is denoted, all  expectations $\mathrm{E}[\cdot]$ and variances $\mathrm{var}[\cdot]$ are taken wrt the behavior policy $\bpol$, that is, $p_{\bpol}$. 
At the same time, recall that $v$- and $q$-functions are for the target policy, $\epol$.
For a function $f$ of (parts of) a trajectory we often write $f$ to mean the random variable $f(\cj)$. 
For example, we write $\nu_t=\nu_t(\cj_{a_t})$, $\mu_t=\mu_t(s_t,a_t)$, etc. The $L^{p}$-norm is defined as $\|f\|_{p}=\rE[\abs{f}^{p}]^{1/p}$.
In the transition-sampling setting, for any function of $s,a,r,s'$, we define its empirical average as
$$
\P_nf=\P_n[f(s,a,r,s')]=n^{-1}\sum_{i=1}^n f(s^{\langle i \rangle },a^{\langle i \rangle },r^{\langle i \rangle },s'^{\langle i \rangle }).
$$
When $f$ also depends on the index $i$, we write $\P_n f(s,a,r,s',i)=n^{-1}\sum_{i=1}^n f(s^{\langle i \rangle },a^{\langle i \rangle },r^{\langle i \rangle },s'^{\langle i \rangle },i)$.
In the trajectory-sampling setting, we define the time average as
$$
\P_Tf=\P_T[f(s,a,r,s')]=(T+1)^{-1}\sum_{t=0}^Tf(s_t,a_t,r_t,s_{t+1}),
$$
and for any function of a trajectory, we define the empirical average as
$$\P_{N}f=\P_{N}[f(\cj)]=N^{-1}\sum_{i=1}^N f(\cj^{\langle i \rangle }).$$
Thus, for a function of $(s,a,r,s')$, we have:  
$$\P_N\P_T f= N^{-1}(T+1)^{-1}\sum_{t=0}^{T}\sum_{j=1}^Nf(s^{\langle j \rangle }_t,a^{\langle j \rangle }_t,r^{\langle j \rangle }_t,s^{\langle j \rangle }_{t+1}),$$
which we also denote by $\P_n=\P_N\P_T$ and also allow functions that depend on the index $i=(j,t)$. 
\Cref{tab:pre} in the appendix summarizes our notation.}

\subsection{Efficiency Bounds}\label{sec:semiparam}
{\blockedit 
In this section, we define formally what we mean by the best-possible asymptotic MSE. We focus on computing efficiency bounds in settings where the data is iid and its distribution fully identifying of the estimand (transition sampling for MDP and infinitely-long-trajectory sampling for TMDP and NMDP) so that we can apply standard semiparametric theory \citep{KosorokMichaelR2008ItEP,TsiatisAnastasiosA2006STaM,bickel98}.
After establishing these bounds, we will actually show they can be achieved by estimators both in these ideal settings and even in more complex sampling settings, such as a single growing trajectory.
We here give a general overview of semiparametric theory as it pertains to our results and provide more technical detail and precise definitions in \cref{appendix:semiparam}.

Suppose our data consists of $n$ iid observations each drawn from a distribution $p$, $O_1,\dots,O_n\sim p$. 
Let us fix $p_0$ as the true, unknown distribution.
While we do not know $p_0$, we assume it belongs to a model $\cm$, \ie, a set of possible data-generating process. 
Given a parameter of interest $R:\mathcal M\to\Rl$, we want to estimate $R(p_0)$ using some estimator $\hat R(O_1,\dots,O_{n})$.
For example, in the transition-sampling setting under MDP, we will let $\cm$ be all distributions $p_{\pi_b}(\cj_1)$ for any choice of MDP and behavior policy, subject to certain minimal regularity and identifiability conditions that ensure the policy value is in fact a function of $p_{\pi_b}(\cj_1)$.

The limiting law of $\hat R$ is the distributional limit of $\sqrt{n}(\hat R-R(p_0))$ and the asymptotic mean-squared error (AMSE) is the second moment of the limiting law, which in turn lower bounds the scaled limit infimum of the mean-squared error (MSE), $\liminf n\rE[(\hat R-R(p_0))^2]$, by the portmanteau lemma.
Roughly, we say $\hat R$ is \emph{regular} wrt $\sqrt{n}$ if its limiting law is invariant to vanishing perturbations to $p_0$ that remain inside $\cm$ (see \cref{def:regular} for precise definition).
This type of regularity is common and is often considered desirable, as otherwise the estimator may behave erratically under completely undetectable changes \citep[see][Sec. 8.1]{VaartA.W.vander1998As}. 
If $\sqrt{n}(\hat R-R(p_0))=\frac1{\sqrt{n}}\sum_{i=1}^n\phi(O_i)+o_p(1/\sqrt{n})$ with $\E\phi(O)=0$ then $\hat R$ is said to be asymptotically linear (AL) with influence function $\phi$ and it follows its limiting law is $\mathcal N(0,\,\E\phi^2(O))$ at $p_0$.

{\newedit Every gradient of $R$ wrt $\cm$ at $p=p_0$ is a Gateaux derivative for all paths through $p_0$ that remain in $\cm$, which is a $p_0$-measurable random variable $\phi(O)$. See \cref{def:pathwise} for precise definition.}
The influence function of any regular AL (RAL) estimator is such a gradient (\cref{thm:gradients}).
The gradient $\phi_{\mathrm{eff}}$ with the least second moment (if such exists) is called the \emph{efficient influence function} (EIF).
This motivated by the fact (\cref{thm:lam}) that 
$$\operatorname{EffBd}(\mathcal M)=\rE_{p_0}{\phi_{\mathrm{eff}}^2},$$
which we call the \emph{efficiency bound}, lower bounds the AMSE of \emph{any} estimator that is regular wrt $\cm$. 
An \emph{efficient estimator} (at $p_0$) is a regular estimator (at $p_0$) with AMSE equal to $\operatorname{EffBd}(\mathcal M)$.\label{page:efficientdef}

If we have $\operatorname{EffBd}(\mathcal M)<\infty$ (\ie, the estimand is differentiable) and an estimator is shown to be AL with the EIF as its influence function then in addition it is also regular and hence RAL and efficient, and conversely every efficient estimator is RAL \citep[Lemma 25.23]{VaartA.W.vander1998As}.\label{page:RALefficient}
This also suggests an estimation strategy: try to approximate $\hat\psi(O)\approx\phi_{\mathrm{eff}}(O)+R(p)$ and use $\hat R=\frac1n\sum_{i=1}^n\hat\psi(O_i)$. Done appropriately, this can provide an efficient estimate. Therefore, deriving the efficient influence function is important both for computing the semiparametric efficiency bound and for coming up with good estimators.

Notice the efficiency bound depends on both $p_0$ and $\cm$. We use $\operatorname{EffBd}(\mathcal M)$ to highlight the latter dependence. Indeed, if $p_0\in\cm\subseteq \cm'$ then $\operatorname{EffBd}(\mathcal M)\leq \operatorname{EffBd}(\cm')$ since estimators that are regular in $\cm'$ are also regular in $\cm$, even though $p_0$ is the \emph{same}.\label{pageref:modelvsp0}
Standard results \citep[\eg,][Thm.~25.21]{VaartA.W.vander1998As} further establish that the efficiency lower bound also applies to \emph{all} estimators (not just regular ones) in a local minimax fashion, where the local worst-case neighborhoods of $p_0$ are restricted to remain in $\cm$.
The efficiency bound is infinite when the estimand is not pathwise differentiable wrt $\cm$, in which case no regular estimators exist \citep{WhitneyK.Newey1990SEB}.

}

\subsection{Summary of Literature on OPE}\label{sec:litope}

OPE is a central problem in both RL and in the closely related dynamic treatment regimes \citep[DTR;][]{MurphySA2001MMMf}. 
OPE is also equivalent to estimating the total treatment effect of some dynamic policy in a causal inference setting. Although we do not explicitly use counterfactual notation (potential outcomes or \emph{do}-calculus), if we assume the usual sequential ignorability conditions \citep{ErtefaieAshkan2014CDTR}, the estimands are the same and our results immediately apply.

In RL, one usually assumes that the (time-invariant) MDP model $\cm_3$ holds. 
Nonetheless, with some exceptions that we review below, OPE methods in RL have largely not leveraged the additional independence and time-invariance structure of $\cm_3$ to improve estimation, and in particular, the effect of this structure on efficiency has not previously been studied and no efficient evaluation method has been proposed.

Methods for OPE can be roughly categorized into three types. The first approach is the \emph{direct method} (DM), wherein we directly estimate the $q$-function and use it to directly estimate the value of the target evaluation policy. One can estimate the $q$-function by a value iteration in a finite-state-and-action-space setting utilizing an approximated MDP based on the empirical distribution \citep{BertsekasDimitriP2012Dpao}. 
More generally, modeling the transition and reward probabilities and using the MDP approximated by the estimates is called the model-based approach \citep{SuttonRichardS1998Rl:a}. When the sample space and action space are continuous, we can apply some functional approximation to $q$-function modeling and use the temporal-difference method \citep{LagoudakisMichail2004LPI} or fitted Q--iteration \citep{Antos2008}. Once we have an estimate $\hat q$, the DM estimate is simply $$\hat\rho_{\mathrm{DM}}=(1-\gamma)\; \rE_{a_0\sim \epol(s_0),s_0 \sim p^{(0)}_{\epol}(s_0)}\bracks{\hat q_0(a_0,s_0)}.$$
Here, recall that we assume the initial distribution $p^{(0)}_{\epol}(s_0)$ is known.
For DM, we can leverage the structure of $\cm_3$ by simply restricting the $q$-function we learn to be the same for all $t$ and solving the fixed point of the Bellman equation. However, DM can fail to be efficient and is also not robust in that, if $q$-functions are inconsistently estimated, the estimate will be inconsistent.

The second approach is \emph{importance sampling} (IS), which averages the data weighted by the density ratio of the evaluation and behavior policies. Given estimates $\hat{\nu}_t$ of $\nu_t$ (or, $\hat{\nu}_t=\nu_t$ if the behavior policy and initial distribution of the offline data $p^{(0)}_{\pi^e}(s_0)$ are known), the IS estimate is simply $$\hat\rho_{\mathrm{IS}}=c_T(\gamma)\;\P_{N}\bracks{\sum_{t=0}^T\gamma^{t}\hat{\nu}_t(\cj_{a_t}) r_t}.$$
A common variant is the self-normalized IS (SNIS) where we divide the $t^\text{th}$ summand by ${\P_{N}  \bracks{\gamma^{t} \hat{\nu}_t}}$.
Recall that $T$ here denotes the \emph{finite} length of the $N$ trajectories in our data.
In finite-horizon problems (\ie, when the estimand is $\rho^{\epol}_{T}$), when the behavior policy is known, IS is unbiased and consistent but its variance tends to be large and it is inefficient \citep{hirano03}. In infinite-horizon problems, we need $T$ to grow for consistent estimation. But even if $T=\infty$ (\ie, our data consists of \emph{full} trajectories), IS can have \emph{infinite} variance because of diminishing overlap, known as the curse of horizon \citep{Liu2018}. Our results (\cref{tab:my_label}) in $\cm_1,\,\cm_2$ characterize more precisely when this curse applies or not.

The third approach is the \emph{doubly robust} (DR) method, which combines DM and IS and is given by adding the estimated $q$-function as a control variate \citep{scharfstein99,DudikMiroslav2014DRPE,zhang2013robust,jiang}. Under $\cm_1$, the DR estimate has the form 
{
\begin{align*}
    \hat\rho_{\mathrm{DR}}&=c_T(\gamma)\; \; \left[ \rE_{a_0\sim \epol(s_0),s_0 \sim p^{(0)}_{\epol}(s_0)}\bracks{\hat q_0(a_0,s_0)}+\right. \\
&\left. + \P_{N}\bracks{\sum_{t=0}^T \gamma^{t}\hat{\nu}_t(\cj_{a_t}) \prns{r_t-\hat q_t(s_t,a_t)+\gamma  \rE_{a_{t+1} \sim \epol(s_{t+1})}\bracks{\hat q_{t+1}(a_{t+1},s_{t+1})|s_{t+1}}}} \right ].
\end{align*}
}
In finite-horizon problems, DR is known to be efficient under $\cm_1$ \citep{NathanUehara2019}.
In infinite horizons, we derive the additional conditions needed for efficiency in $\cm_1$ in \cref{sec:m1and2}. 

Many variations of DR have been proposed. \citet{thomas2016} propose both a self-normalized variant of DR and a variant blending DR with DM when density ratios are extreme. \citet{Chow2018} propose to optimize the choice of $\hat q(s,a)$ to minimize variance rather than use a plug-in. \citet{Kallus2019IntrinsicallyES} propose a variant that is similarly locally efficient but further ensures asymptotic MSE no worse than DR, IS, and SNIS under misspecification and stability properties similar to self-normalized IS.

However, all of the aforementioned IS and DR estimators do not leverage Markov structure and \emph{fail} to be efficient under $\cm_2$. 
Recently, in finite horizons, \cite{NathanUehara2019} derived the efficiency bound of $\rho^{\epol}_{T}$ under $\cm_2$ and provided an efficient estimator termed Double Reinforcement Learning (DRL), taking the form
\begin{align*}
\hat\rho_{\mathrm{DRL(\cm_2)}} &=c_T(\gamma)\left \{ \frac{1}{N}\sum_{i=1}^N \rE_{a_0\sim \epol(s_0),s_0 \sim p^{(0)}_{\epol}(s_0)}\bracks{\hat q^{\langle i \rangle }_0(a_0,s_0)}\right. \\
&\left. + \frac{1}{N}\sum_{i=1}^N \bracks{\sum_{t=0}^T \gamma^{t}\hat{\mu}^{\langle i \rangle }_t(s^{\langle i \rangle }_t,a^{\langle i \rangle }_t)\prns{r^{\langle i \rangle }_t-\hat q_t^{\langle i \rangle }(s^{\langle i \rangle }_t,a^{\langle i \rangle }_t)+\gamma \rE_{a_{t+1} \sim \epol(s^{\langle i \rangle }_{t+1})}\bracks{\hat q_t^{\langle i \rangle }(s^{\langle i \rangle }_{t+1},a_{t+1})|s^{\langle i \rangle }_{t+1}}}} \right \},
\end{align*}
where $\hat{\mu}^{\langle i \rangle },\,\hat q^{\langle i \rangle }$ can either be estimated in-sample ($\hat q_t^{\langle i \rangle }=\hat q_t$ and assuming a Donsker condition) or cross-fitting (the sample is split and $\hat q_t^{\langle i \rangle }$ is fit on the fold that excludes $i$). DRL's efficiency depends only on the rates of convergence of these estimates, which can be as slow as $N^{-1/4}$ thus enabling the use of blackbox machine learning methods.
In infinite horizons, we derive the additional conditions needed for efficiency in $\cm_2$ in \cref{sec:m1and2}.

However, again, all of the aforementioned IS, DR, and DRL estimators do not leverage time-invariance and \emph{fail} to be efficient under $\cm_3$. 
Our results extend the notion of the curse of dimension and demonstrate that even estimators in $\cm_2$, such as the efficient $\hat\rho_{\mathrm{DRL(\cm_2)}}$, can fail to be consistent as $\mu_t$ can also explode just like $\nu_t$.
In contrast, in $\cm_3$, regardless of the rate of growth of $\mu_t,\,\nu_t$, consistent evaluation is possible from even just a single trajectory and knowledge of the initial distribution. 

Recently, \citet{Liu2018} proposed a variant of the IS estimator for $\cm_3$ that uses the ratio of the stationary distributions in hopes of overcoming the curse of horizon. We describe this estimator in detail in \cref{sec:isestmdp}. Its asymptotic MSE was not previously studied. We provide some results in the parametric setting. The properties in the nonparametric setting are not known. In particular, as we discuss in \cref{sec:isestmdp}, its lack of doubly robust structure and its not being an empirical average of martingale differences make analysis particularly challenging. At the same time, these issues also suggest that the estimator is inefficient.

\section{Efficiency Bounds in Infinite Horizons}\label{sec:timevariantmdp}
{\blockedit 
The efficiency bounds for $\rho^{\epol}_T$ in finite horizons under NMDP and TMDP are derived in \cite{NathanUehara2019}. First, we extend these results to infinite horizons, focusing in particular on when the bounds are \emph{infinite}. Then -- and more importantly -- we study the efficiency bound in MDP.}

\subsection{Efficiency Bounds in Non-Markov and Time-Variant Markov Decision Processes}\label{sec:cm2}

{\neweditf First, we formally define NMDP and TMDP as statistical models for our data-generating process. As data, we consider observing $N$ (infinitely long) trajectories $\cj$ from the behavior-policy-induced distribution $p_{\pi_b}(\cj)$. The model is the set of possibilities for $p_{\pi_b}(\cj)$. The NMDP model is given by (almost) all arbitrary distributions on the sequence $\cj$. Remark that these models here are different from the transition-sampling setting and trajectory-sampling setting. 

\begin{definition}[NMDP models $\cm_1,\cm_{1,b}$]\label{def:nmdpmodel}
The NMDP model $\cm_1$ is defined by all distributions $p_{\pi_b}(\cj)$ such that the conditional distribution of each variable in $\cj$ given the past is absolutely continuous wrt the respective base measure (so it has a density) and the conditional distribution of action given history, $\pi_{b,t}$, is such that the (known and fixed) evaluation policy, $\pi_{e,t}$, is absolutely continuous wrt it. Hence, we can write in the form: 
\begin{align*}
    p_{\pi_b}(\cj) =   p^{(0)}_{\pi_b}(s_0)\pi_{b,0}(a_0\mid s_0)p(r_0\mid s_0,a_0)p(s_1\mid \cj_{r_0})\pi_{b,1}(a_1\mid \cj_{s_1})p(r_1\mid \cj_{a_1})\cdots . 
\end{align*}
We also define the model $\cm_{1,b}$ where we assume the behavior policy is known; that is, $\pi_{b,t}$ and $p_{\pi_b}^{(0)}$ are fixed at their known value and not allowed to vary. 
\end{definition}
The last restriction in the definition of $\cm_1$ is known as \emph{weak overlap} and it is equivalent to saying $\nu_t$ exists.
It is necessary so to ensure that $\rho^{\epol}$ is a function of $p_{\pi_b}(\cj)$, that is, is identifiable from the data \citep{ShakeebKhan2010IISC}.
Observing infinitely-long trajectories is also necessary for identifiability, but when constructing estimators we will show it suffices to observe trajectories of modestly growing length.
Then, $\rho^{\epol}$ is a functional of $p_{\pi_b}(\cj)$ given by $(1-\gamma)\E[\sum_{t=0}^T\gamma^{t}\nu_tr_t]$, that is, it is a well-defined map $\cm_1\to\Rl$.

The TMDP model is obtained by restricting the NMDP model to satisfy the Markovian condition.
\begin{definition}[TMDP models $\cm_2,\cm_{2,b}$]\label{def:tmdpmodel}
The TMDP model $\cm_2$ is defined by restricting the model $\cm_1$ so that $s_{t+1}$ is conditionally independent of $\cj_{r_{t-1}}$ given $s_t,a_t$, and $r_{t}$ is conditionally independent of $\cj_{r_{t-1}}$ given $s_t,a_t$, and $a_{t}$ is conditionally independent of $\cj_{r_{t-1}}$ given $s_t$. Hence, we can write in the form: 
\begin{align*}
    p_{\pi_b}(\cj) =   p^{(0)}_{\pi_b}(s_0)\pi_{b,0}(a_0\mid s_0)p(r_0\mid s_0,a_0)p(s_1\mid s_0,a_0)\pi_{b,1}(a_1\mid s_1)p(r_1\mid s_1,a_1)\cdots . 
\end{align*}
Similarly, we define $\cm_{2,b}$ by fixing $\pi_{b,t}$ and $p_{\pi_b}^{(0)}$ at their known value.
\end{definition}

All of our models are \emph{nonparametric} in the sense that we do not further restrict these distributions in any way beyond requiring densities and overlap.}

\edit{We now proceed to compute the efficiency bounds for $\rho^{\epol}$ in these models.}
By slightly modifying the results of \cite{NathanUehara2019}, we obtain the following theorems.

\begin{theorem}[EB under NMDP]\label{thm:m1_bound}
\begin{align}
\label{eq:m1_bound}
\mathrm{EB}({\cm_1})=\edit{\mathrm{EB}({\cm_{1,b}})}=(1-\gamma)^{2}\;\sum_{k=1}^{\infty}\mathrm{E}[\gamma^{2(k-1)}\nu_{k-1}^{2}(\cj_{a_{k-1}}) \mathrm{var}\prns{r_{k-1}+\gamma v_k(\cj_{s_k})\mid\cj_{a_{k-1}}} ]. 
\end{align}
\end{theorem}

\begin{theorem}[EB under TMDP]\label{thm:m2_bound}
\begin{align}
\label{eq:m2_bound}
\mathrm{EB}({\cm_2})=\edit{\mathrm{EB}({\cm_{2,b}})}=(1-\gamma)^{2}\;\sum_{k=1}^{\infty}\mathrm{E}[\gamma^{2(k-1)}\mu_{k-1}^{2}(s_{k-1},a_{k-1}) \mathrm{var}\prns{r_{k-1}+\gamma v_k(a_k,s_k)\mid a_{k-1},s_{k-1}} ]. 
\end{align}
\end{theorem}

\begin{remark}
\Cref{eq:m1_bound,eq:m2_bound} are almost the same as the limit as $T\to\infty$ of $c^2_T(\gamma)$ times the \emph{finite}-horizon efficiency bounds derived by \citet{NathanUehara2019}.
They are the same if we replace the lower summation limit with $k=0$ instead of $k=1$ in \cref{eq:m1_bound,eq:m2_bound}. This is because we here assume $p^{(0)}_{\epol}$ is known while in \citet{NathanUehara2019} the assumption is that $p^{(0)}_{\bpol}=p^{(0)}_{\epol}$ are unknown, the uncertainty due to which increases the efficiency bound.
\end{remark}

{\blockedit
\Cref{thm:m1_bound,thm:m2_bound} show that, when \cref{eq:m1_bound,eq:m2_bound} are finite, the best-achievable leading term in the MSE of any regular estimator in NMDP or TMDP is $\mathrm{EB}(\cm_1)/N$ or $\mathrm{EB}(\cm_2)/N$, respectively. It also shows that the knowledge of $\bpol,p^{(0)}_{\bpol}$ does not improve the bound. The intuitive reason for this is that $\rho^{\epol}$ is only a function of the transition- and reward-distribution parts of $p_{\pi_b}(\cj)$ so that $\bpol,p^{(0)}_{\bpol}$ are ancillary. 
When the efficiency bound takes an infinite value, the estimand is not pathwise differentiable wrt the model and no regular $\sqrt{n}$-consistent estimator exists \citep{WhitneyK.Newey1990SEB}.
}

\begin{corollary}[Sufficient conditions for existence of efficiency bounds]\label{cor:m1m2ebexist}
If $\magd{\nu_k}_\infty=o(\gamma^{-k})$, then $\mathrm{EB}(\cm_1)<\infty$. If $\magd{\mu_k}_\infty=o(\gamma^{-k})$, then $\mathrm{EB}(\cm_2)<\infty$. Moreover, if $p_{\bpol}\in\cm_2$ and $\mathrm{EB}(\cm_1)<\infty$, then $\mathrm{EB}(\cm_2)<\infty$.
\end{corollary}

\begin{remark}[The curse of horizon in $\cm_1$, extended]
To demonstrate the curse of horizon, \citet{Liu2018} gave an example where the IS estimator has a diverging variance as horizon grows.
But it is not clear if -- and without assuming MDP structure -- there might be another estimator that would not suffer from this. Our results show that in fact there is not.
If we take \emph{any} example where $\mathrm{var}\prns{r_{k-1}+\gamma v_k|\cj_{a_{k-1}}}$ are uniformly lower bounded (\ie, state transitions and reward emissions are non-degenerate), then as long as $\rE[\log(\eta_k)]\geq-\log(\gamma)$ for all $k$, we will necessarily have that $\mathrm{EB}(\cm_1)=\infty$. (Notice that $\rE[\log(\eta_k)]$ is exactly the \edit{expected} Kullback-Leibler divergence.) In this case, as long as we are not restricting the model beyond $\cm_1$, we simply \emph{cannot} break the curse of horizon and it affects \emph{all} (regular) estimators, not just IS.
\end{remark}

\begin{remark}[The curse of horizon in $\cm_2$, a milder version of the original]
Our results further extend the curse of horizon to $\cm_2$, providing another refinement of the notion. 
The curse is milder in $\cm_2$ than in $\cm_1$, since the EBs are necessarily ordered.
It is, in fact, much milder.
In particular, rather than involve the growth of the cumulative density ratios, whether $\mathrm{EB}(\cm_2)$ converges or diverges depends on the growth of the \emph{marginal} density ratios. These, of course, can also grow and $\mathrm{EB}(\cm_2)$ can diverge. However, while we can easily make $\mathrm{EB}(\cm_1)=\infty$ even with a simple MDP example, to make $\mathrm{EB}(\cm_2)$ diverge we need a more pathological example. It can be verified that if $p_{\pi_b}$ is actually stationary (or, nonstationary but ergodic) and the stationary distributions overlap, then we will necessarily have $\magd{\mu_k}_\infty=O(1)$.

This means that, for an MDP, we can overcome the curse of horizon that affects estimators like $\hat\rho_{\mathrm{DR}}$ and $\hat\rho_{\mathrm{IS}}$ by using estimators that are efficient under $\cm_2$, the first of which was proposed by \citet{NathanUehara2019}, \ie, $\hat\rho_{\mathrm{DRL(\cm_2)}}$. However, this is still \emph{not} efficient in an MDP case.  %
In fact, this is not just a matter of constants: this will not even yield the right scaling of the MSE. 
\end{remark}

\subsection{Efficiency Bounds in Time-Invariant Markov Decision Processes}
\label{sec:M3}

Next, we consider the MDP model. For the efficiency bound computation, we focus on the transition sampling setting, where we observe $n$ draws from $p_{\pi_b}(\cj_{s_1})$. {\neweditf We next formally define an MDP as a statistical model for our data. 
\begin{definition}[MDP models $\cm_3,\cm_{3,b}$]\label{def:mdpmodel}
The MDP model, $\cm_3$, is given by all distributions $p_{\pi_b}(\cj_{s_1})$ on $(s,a,r,s')$ such that the distribution of $s'$ is independent of $r$ given $s,a$, the conditional distribution of each variable given the past is absolutely continuous wrt the respective base measure (so it has a density), and further the base measure is absolutely continuous wrt the distribution of $s$. As before, we define $\cm_{3,b}$ by fixing $p_{\pi_b}^{(0)}$ and $\bpol$ at their known value.
\end{definition}
The last restriction ensures $w(s)$ exists without putting additional restrictions on the MDP itself. The existence of $w(s)$ is the analogue of overlap for the MDP setting and is necessary for identifiability. Then, $\rho^{\epol}$ is a functional of $p_{\pi_b}(\cj_{s_1})$ given by $\E[w(s)\eta(s,a)r]$, that is, it is a well-defined map $\cm_3\to\Rl$.}

\begin{theorem}[EB under MDP]
\label{thm:m3_bound} 
The EIF in either of $\cm_{3}$ or $\cm_{3,b}$ is 
\begin{align*}
    \phi_\text{eff}(s,a,r,s')=w(s)\eta(s,a)(r+\gamma v(s')-q(s,a)).
\end{align*}
The efficiency bound in either model is therefore
\begin{align}\label{eq:ergo}
\mathrm{EB}(\cm_3) &=\rE\left[w^{2}(s)\eta^{2}(s,a)\prns{r+\gamma v(s')-q(s,a)}^{2}\right].
\end{align}
\end{theorem}

\Cref{thm:m3_bound} shows that the lower bound of the first order asymptotic MSE is $\mathrm{EB}(\cm_3)/n$. It also shows that the knowledge of $\bpol,p^{(0)}_{\bpol}$ does not improve the bound. This suggests that in MDP, the MSE should scale inversely with the number of transitions ($n$) we observe, \emph{not} the number of trajectories ($T$). 
While standard efficiency analysis does not apply to the trajectory-sampling setting in MDP since the transitions are dependent, we will show in \cref{sec:dm_infinite} that we can nonetheless achieve the \emph{same} efficiency bound with a scaling of $n=N(T+1)$ under certain mixing assumptions. Thus, the achievable MSE under MDP is a factor of $T$ faster than under NMDP and TMDP.
In this sense, efficiency in $\cm_3$ corresponds to an improvement in the \emph{rate}, not just the constant, relative to efficiency in $\cm_1$ or $\cm_2$. This is in contrast to the comparison between $\cm_1$ and $\cm_2$, which have efficiency bounds that are on the same scale and only differ in the leading coefficient.

\begin{remark}[Unknown $p_{\pi_e}^{(0)}$]
In our set up we assumed $p_{\pi_e}^{(0)}$ is known, but our results can be extended to the case where $p_{\pi_e}^{(0)}$ is unknown but we see samples from it. In particular, suppose $p_{\pi_e}^{(0)}$ is allowed to vary arbitrarily in the model (but remains a density wrt the state base measure) and our data consists of $n$ iid draws of $(s_0,s,a,r,s')$ from $p_{\pi_e}^{(0)}(s_0)p_{\pi_b}(\cj_1)$. Then a modification of \cref{thm:m3_bound} shows that the EB (whether we know the behavior policy or not) is 
\begin{align*}
{\newedit \var_{p_{\pi_e}^{(0)}}[v(s_0)]+
\rE\left[w^{2}(s)\eta^{2}(a,s)\prns{r+\gamma v(s')-q(s,a)}^{2}\right].}
\end{align*}
Compared to \cref{eq:ergo}, we have an additional term corresponding to the variance wrt $p_{\pi_e}^{(0)}$.

{\newedit When $\gamma=0$, this reduces to the bound in the no-horizon bandit OPE setting \citep{robins94}:
\begin{align*}
\var_{p_{\pi_e}^{(0)}}[v(s_0)]+
\rE\left[\eta^{2}(a,s)\prns{r-q(s,a)}^{2}\right],
\end{align*}
where here $q(s,a)=\E[r\mid s,a]$ becomes simply the outcome regression function.
}
\end{remark}

\section{Efficient Estimators for Infinite Horizons under NMDP and TMDP}\label{sec:m1and2}

Before turning to developing an efficient estimator under the MDP model, we briefly review how we can extend the efficient finite-horizon DRL estimators of \citet{NathanUehara2019} to be efficient in the infinite-horizon NMDP and TMDP settings. In these settings, we have acess to the data $\{\cj^{\langle i \rangle }\}_{i=1}^N$ where each $\cj$ follows $\cm_1$ in an NMDP, and $\cm_2$ in a TMDP, respectively. Note we do not need any stationarity in the offline data in this section.

DRL is a meta-estimator: it takes in as input estimators for $q$-functions and density ratios and combines them in a particular manner that ensures efficiency even when the input estimators may not be well behaved. For example, metric entropy or Donsker assumptions can be avoided by using a cross-fitting strategy \citep{ZhengWenjing2011CTME,ChernozhukovVictor2018Dmlf,1987CEot}.
We proceed by presenting the infinite-horizon extensions of the DRL estimators of \citet{NathanUehara2019} and their properties. Again, the two DRL estimators present here are \emph{not} efficient under $\cm_3$.

\subsection{Non-Markov Decision Process}\label{sec:nmdp}

The infinite-horizon extension of the DRL estimator under $\cm_1$ is as follows. 
{\blockedit\edit{We consider the trajectory sampling setting where we observe $N$ trajectories.}
Fix some horizon truncation $\omega_N$. 
\edit{Let $q^{\omega_N}_t=\E_{\epol}[\sum_{k=t}^{\omega_N} \gamma^{t-k} r_t \mid \cj_{a_t}],\,v^{\omega_N}_t=\E_{\epol}[\sum_{k=t}^{\omega_N} \gamma^{t-k} r_t \mid \cj_{s_t}]$.}
Then the estimator is given by
{\newedit 
$$
\hat\rho_{\mathrm{DRL(\cm_1)}}=
c_{w_N}(\gamma)
\;\bracks{ \frac{1}{N}\sum_{i=1}^{N}
\sum_{t=0}^{\omega_N}\rE_{s_0\sim p^{(0)}_{\epol}}[\hat v^{\langle i \rangle }_0(s_0^{})]+ \gamma^t\hat \nu^{\langle i \rangle }_t(\cj_{a_t}^{\langle i \rangle }) \prns{r^{\langle i \rangle }_t-\hat q^{\langle i \rangle }_t(\cj_{a_t}^{\langle i \rangle })+\gamma 
\hat v^{\langle i \rangle }_{t+1}(\cj_{s_{t+1}}^{\langle i \rangle })}
},
$$}
where $\hat \nu^{\langle i \rangle }_t,\,\hat q^{\langle i \rangle }_t$ are some plug-in estimates of $\nu_t,\,\edit{q^{\omega_N}_t}$ to be used for the $i^\text{th}$ observation and $\hat v^{\langle i \rangle }_t(\cj_{s_t})=\rE_{a_t\sim\epol(\cdot\mid\cj_{s_t})}\bracks{\hat q^{\langle i \rangle }_t(\cj_{a_t}) \mid \cj_{s_t}}$ is the corresponding $v$-estimate.
Notice that $\hat v^{\langle i \rangle }_t$ is computable as it is an integral wrt the \emph{known} measure $\epol(\cdot\mid\cj_{s_t})$ (\eg, it is a simple a sum if $\mathcal A$ is finite).

We can consider two cases. In the \emph{adaptive version}, we construct functional estimators $\hat \nu_t,\hat q_t$ based on the whole data and then set $\hat \nu^{\langle i \rangle }_t=\hat\nu_t,\,\hat q^{\langle i \rangle }_t=\hat q_t$.
The adaptive version of $\hat\rho_{\mathrm{DRL(\cm_1)}}$ is exactly the DR estimator, $\hat\rho_{\mathrm{DR}}$.

In the \emph{cross-fitting version}, the sample is evenly split into two folds and $\hat \nu^{\langle i \rangle }_t,\,\hat q^{\langle i \rangle }_t,\,\,\hat v^{\langle i \rangle }_t$ are computed on estimates fit on the opposite fold so that they are independent of data point $i$. {\newedit Namely, the cross-fitting procedure is:
\begin{itemize}
    \item Split the dataset into two disjoint datasets $\Dcal_0$ and $\Dcal_1$. Let $j\langle i \rangle $ be such that $\cj^{\langle i \rangle }\in \Dcal_{j\langle i \rangle }$.
    \item Using only the trajectories in $\Dcal_0$, construct the functional estimators $\hat \nu^{[0]}_t,\hat q^{[0]}_t$ for $t\leq \omega_N$. And, using only the trajectories in $\Dcal_1$, construct the functional estimators $\hat \nu^{[1]}_t,\hat q^{[1]}_t$ for $t\leq \omega_N$. 
    \item Set $\hat \nu^{\langle i \rangle }_t=\hat\nu^{[j\langle i \rangle ]}_t,\,\hat q^{\langle i \rangle }_t=\hat q^{[j\langle i \rangle ]}_t$.
\end{itemize}
}
\citet[Section 6]{NathanUehara2019} discusses the estimation of ${\nu}_t,q_t^{\omega_N}$, that is, $q$-functions for finite-horizon problems. In particular, if the behavior policy is known we can simply let $\hat\nu_t^{\langle i \rangle }=\nu_t$. 
\edit{As we make formal below, our $q$-estimates need only estimate $q_t^{\omega_N}$ and not $q_t$, which depends on all future rewards ad infinitum. This can be done using only the truncated trajectory $\cj_{r_{\omega_N}}^{\langle i \rangle }$ (\eg, using regression), and given $q$-estimates, the estimator similarly only depends on $\cj_{r_{\omega_N}}^{\langle i \rangle }$. Therefore, while we can consider it as an estimator in the $\cm_1$ model where we observe the infinitely long $\cj^{\langle i \rangle }$, it is in fact implementable even if we just see finite trajectories of length at least $\omega_N$.}}

We can now state a straightforward infinite-horizon extension of the efficiency result of \citet[Theorems 4 and 6]{NathanUehara2019} under $\cm_1$ in finite-horizons. Essentially, we just need to be careful about choosing $\omega_N$.
{\newedit We focus on the analysis of the cross-fitting version. Recall the L2 errors are defined on the offline data distribution. }

\begin{theorem}[Asymptotic property of $\hat{\rho}_{\mathrm{DRL(\cm_1)}}$]\label{thm:m1_inf}
 Define $\kappa^\nu_N$, $\kappa^q_N$ such that $\|\hat{\nu}^{[j]}_{t}-\nu_{t}\|_{2}\leq\kappa^\nu_N$, $\|\hat{q}^{[j]}_t-q^{\omega_N}_t\|_{2}\leq\kappa^q_N$ for  $0\leq t\leq \omega_N,\,j=0,1$ {\neweditf Assume (\ref{thm:m1_inf}a) ${\nu}_{t}\leq C^{t}$ and $\gamma C<1$ for some $C>0$, (\ref{thm:m1_inf}b) $0\leq\hat{q}^{[j]}_t\leq (1-\gamma)^{-1}R_{\mathrm{max}}$ and  $0\leq\hat{\nu}^{[j]}_{t}\leq C^{t}$ for the above-mentioned $C$ and $0\leq t\leq \omega_N,\,j=0,1$,} (\ref{thm:m1_inf}c) $(\kappa^\nu_N\vee\kappa^q_N)\omega_N=\op(1)$,
(\ref{thm:m1_inf}d) $\omega_N=\omega(\frac{\log N}{1-\gamma})$,
(\ref{thm:m1_inf}e) $\kappa^\nu_N\kappa^q_N\omega_{N}=\op(N^{-1/2})$.  
Then, $\hat{\rho}_{\mathrm{DRL(\cm_1)}}$ is RAL and efficient; in particular, $\sqrt{N}(\hat{\rho}_{\mathrm{DRL(\cm_1)}}-\rho^{\epol})\stackrel{d}{\rightarrow} \mathcal{N}(0,\mathrm{EB}({\cm_1}))$. %

\end{theorem}

Each assumption has the following interpretation. 
\edit{Condition (\ref{thm:m1_inf}a) is sufficient to guarantee that the EB is finite (see \cref{cor:m1m2ebexist}).}
Conditions (\ref{thm:m1_inf}b), (\ref{thm:m1_inf}c) are required to control a term related to a stochastic equicontinutiy condition. In particular, even if we observe infinitely long trajectories ($T=\infty$) we cannot set $\omega_N=\infty$.
{\newedit Notably, with cross-fitting, we make no assumptions about our nuisance estimates except for rates, meaning we can use blackbox machine learning methods that may not satisfy strong metric entropy conditions. Without cross-fitting, the same theorem would hold if we additionally impose a Donsker condition on $\hat \nu_t,\hat q_t$ but such would be restrictive on the types of estimators allowed (see \cref{def:donsker} for definition of Donsker).}\label{pageref:donskermention}
Condition (\ref{thm:m1_inf}d) is needed so that $\rho^{\epol}_{\omega_N}=\rho^{\epol}+o(1/\sqrt{N})$.
The condition (\ref{thm:m1_inf}e) is needed to show the inflation in variance due to using plug-in estimates is $\op(N^{-1/2})$, that is, the asymptotic variance is not changed because of the plug-in. Because of the mixed bias property \citep{RotnitzkyAndrea2019Copw} of the influence function, the rate is multiplicative in the two estimators' convergence rate.
Finally, note that if we know the behavior policy we can take $\kappa^\nu_N=0$ so 
the conditions on $\kappa^q_N$ are very lax. {\blockedit If the behavior policy is not known, we can still allow very slow rates; for example, if $\omega_N=\log^{1+\epsilon} N,\,\kappa^\nu_N=N^{-\zeta_\nu},\,\kappa^q_N=N^{-\zeta_q}$ then we only need the rates to satisfy $\zeta_\nu+\zeta_q>\frac12,\,\zeta_\nu\vee \zeta_q>0,\,\epsilon>0$.}

\subsection{Time-Variant Markov Decision Process}\label{sec:tmdp}

In finite-horizons, \citet{NathanUehara2019} proposed the first efficient OPE estimator under TMDP. We now repeat the process in the previous section and show the results can be easily extended to the infinite-horizon case.
Fix some horizon truncation $\omega_N$.
{\blockedit Let $q^{\omega_N}_t(s_t,a_t)=\E_{\epol}[\sum_{k=t}^{\omega_N} \gamma^{t-k} r_t \mid s_t,a_t],\,v^{\omega_N}_t(s_t)=\E_{\epol}[\sum_{k=t}^{\omega_N} \gamma^{t-k} r_t \mid s_t]$.
The estimator is given by
$$
\hat\rho_{\mathrm{DRL(\cm_2)}}=
c_{w_N}(\gamma)
\;\frac{1}{N}\sum_{i=1}^{N}
{
\sum_{t=0}^{\omega_N}\rE_{s_0\sim p^{(0)}_{\epol}}[\hat v^{\langle i \rangle }_0(s_0^{})] + \gamma^t \hat \mu^{\langle i \rangle }_t(s_t^{\langle i \rangle },a_t^{\langle i \rangle }) \prns{r_t^{\langle i \rangle }-\hat q^{\langle i \rangle }_t(s_t^{\langle i \rangle },a_t^{\langle i \rangle })+\gamma \hat v^{\langle i \rangle }_{t+1}(s_{t+1}^{\langle i \rangle })}},
$$
where $\hat \mu^{\langle i \rangle }_t,\,\hat q^{\langle i \rangle }_t$ are some plug-in estimates of $\mu_t,\,\edit{q^{\omega_N}_t}$ to be used for the $i^\text{th}$ observation and $\hat v^{\langle i \rangle }_t(s_t)=\rE_{a_t \sim \epol(\cdot\mid s_t)}\bracks{\hat q^{\langle i \rangle}_t(s_t,a_t) \mid {s_t}}$, which is an integral over $a\sim \epol(\cdot\mid s_t^{\langle i \rangle })$, which is known.
Again, $\mu^{\langle i \rangle }_t,\,q^{\langle i \rangle }_t$ can be estimated adaptively or using cross-fitting as in \cref{sec:nmdp}.
\citet[Section 6]{NathanUehara2019} discusses strategies for estimating ${\mu}_t,\,q_t^{\omega_N}$, that is, $q$-functions for finite-horizon problems.}

We can again state a straightforward infinite-horizon extension of the efficiency result of \citet[Theorem 9]{NathanUehara2019} under $\cm_2$ in finite-horizons.
{\newedit We focus on the analysis of the cross-fitting version. }
\begin{theorem}[Asymptotic property of $\hat{\rho}_{\mathrm{DRL(\cm_2)}}$]\label{thm:m2_inf}
{\neweditf Define $\kappa^\mu_N$, $\kappa^q_N$ such that $\|\hat{\mu}^{[j]}_{t}-\mu_{t}\|_{2}\leq\kappa^\mu_N$, $\|\hat{q}^{[j]}_t-\edit{q^{\omega_N}_t}\|_{2}\leq\kappa^q_N$ for $0\leq t\leq \omega_N,\,j=0,1$.  Assume (\ref{thm:m1_inf}a) ${\mu}_{t}\leq C'^{t}$ and $\gamma C'<1$ for some $C'>0$, (\ref{thm:m1_inf}b) $0\leq\hat{q}^{[j]}_t\leq (1-\gamma)^{-1}R_{\mathrm{max}}$ and $0\leq\hat{\mu}^{[j]}_{t}\leq C'^{t}$ for the above-mentioned $C'$ and $0\leq t\leq \omega_N,\,j=0,1$. (\ref{thm:m1_inf}c) $(\kappa^\mu_N\vee\kappa^q_N)\omega_N=\op(1)$,
(\ref{thm:m1_inf}d) \edit{$\omega_N=\omega(\frac{\log N}{1-\gamma})$},
(\ref{thm:m1_inf}e) $\kappa^\mu_N\kappa^q_N\omega_{N}=\op(N^{-1/2})$. 
Then, \edit{$\hat{\rho}_{\mathrm{DRL(\cm_2)}}$ is RAL and efficient; in particular,} $\sqrt{N}(\hat{\rho}_{\mathrm{DRL(\cm_2)}}-\rho^{\epol})\stackrel{d}{\rightarrow} \mathcal{N}(0,\mathrm{EB}({\cm_2}))$.} %
\end{theorem}

\edit{Again, the estimate is feasible as long as we observe trajectories of length $\omega(\log N)$, and the cross-fitted version makes no assumption on nuisance estimates except rates. And, again, we can allow very slow rates: if $\omega_N=\log^{1+\epsilon} N,\,\kappa^\mu_N=N^{-\zeta_\mu},\,\kappa^q_N=N^{-\zeta_q}$ then we only need the rates to satisfy $\zeta_\mu+\zeta_q>\frac12,\,\zeta_\mu\vee \zeta_q>0,\,\epsilon>0$.}

\subsection{Inefficiency under MDP}

The methods in this section could be applied to an MDP. In fact, many papers using DR-type methods such as $\hat{\rho}_{\mathrm{DR}}$ (equal to the adaptive version of $\hat\rho_{\mathrm{DRL(\cm_1)}}$) assume that the underlying distribution is MDP when estimating $q$-functions: \ie, they fit $q$-functions that depend only on $s_t,a_t$ and that are time-invariant. However, using this additional structure in order to produce better $q$-function estimates does \emph{not} improve the asymptotic variance. Indeed, even if we used the oracle $q$-functions and oracle density ratios, we still only obtain the efficiency bounds in \cref{thm:m1_inf,thm:m2_inf}.
Thus, even though we might use a total of $\bigO(NT)$ transition observations to get better $q$-function estimates, if we use standard DR-type methods, this will get washed out, at least asymptotically, and our variance will only vanish as $\bigO(1/N)$.

{\blockedit 
\section{Efficient Estimator for Markov Decision Process} \label{sec:dm_infinite}

In this section, we propose an estimator that is \emph{efficient} under the MDP model by leveraging the EIF obtained in \cref{thm:m3_bound}. To our knowledge it is the first such estimator. We consider both the transition-sampling and trajectory-sampling settings and show that, under appropriate conditions in each setting, we achieve the \emph{same} efficiency bound derived in \cref{thm:m3_bound} asymptotically. Specifically, the conditions in the trajectory-sampling setting include certain sufficient mixing so that dependent-data observations that sufficiently far apart appear near-independent. We nonetheless need to develop a special sample-splitting procedure to handle the dependent data in this setting.

For brevity, we focus here on the case where the behavior policy is known, which is more relevant in RL. That is, we have that $\eta(s,a)$ is known. Our results can easily be extended to the unknown behavior policy case as well (see \cref{remark:unknownbehaviorpolicy} below).

\subsection{Efficient Estimation Under Transition Sampling}\label{sec:simulation}

The key to our estimator is the following estimating function, defined for a given $w$- and $q$-function:
\begin{align*}
\psi(s,a,r,s';w',q')=(1-\gamma)\rE_{p_{\pi_e}^{(0)}}[v'(s_0)]+w'(s)\eta(s,a)\prns{r+\gamma v'(s')-q'(s,a)},
\end{align*}
where we use $w',\,q'$ to denote dummy such functions and use the shorthand that, given any $q'$, we let $v'(s)=\rE_{\pi_e}[q'(s,a)\mid s]=\int_a q'(s,a)\pi_e(a\mid s) d\lambda_{\mathcal A}(a)$, which is computable as an integral of $q'$ wrt the known $\pi_e$ (a sum if $\mathcal A$ is finite).
Similarly, given $q'$, the first term above ($\rE_{p_{\pi_e}^{(0)}}[v'(s_0)]$) is also computable as both $p_{\pi_e}^{(0)}$ and $\pi_e$ are known. Notice this term is also constant wrt $(s,a,r,s')$.
This estimating function is derived from the EIF in \cref{thm:m3_bound}: when $q'=q,\,w'=w$, we have $\psi(s,a,r,s';w,q)=\rho^{\epol}+\phi_\text{eff}(s,a,r,s')$.

Based on this estimating function, our estimator is 
{\newedit 
\begin{align}%
  \hat{\rho}_{\mathrm{DRL(\cm_3)}}&=\P_n[\psi(s,a,r,s';\{\hat w^{\langle i \rangle }\}_{i=1}^n,\{\hat q^{\langle i \rangle }\}_{i=1}^n)]
 \nonumber \\
  &=
   \frac1{n}\sum_{i=1}^n
  {(1-\gamma)}\rE_{s_0\sim p_{\pi_e}^{(0)}}[\hat v^{\langle i \rangle }(s_0)] \nonumber \\
  &\phantom{=}+\frac1{n}\sum_{i=1}^n
  \hat w^{\langle i \rangle }(s^{\langle i \rangle })\eta(a^{\langle i \rangle },s^{\langle i \rangle })({r^{\langle i \rangle }+\gamma \hat v^{\langle i \rangle }(s'^{\langle i \rangle })-\hat q^{\langle i \rangle }(s^{\langle i \rangle },a^{\langle i \rangle })})\label{eq:martingaledifference},
\end{align}
}
where $\hat w^{\langle i \rangle },\,\hat q^{\langle i \rangle }$ are some plug-in estimates of $w,\,q$ to be used for the $i^\text{th}$ observation.
Recall $\hat v^{\langle i \rangle }$ is defined in terms of $\hat q^{\langle i \rangle }$ by taking expectations over $a\sim\pi_{e}(\cdot\mid s)$. Again, we consider \emph{two} cases. First, we consider an \emph{adaptive version}, where we let $\hat w^{\langle i \rangle }=\hat w,\,\hat q^{\langle i \rangle }=\hat q$ be shared among all data points and be estimated on the whole dataset of $n$ observations of $(s,a,r,s')$. Second, we consider a cross-fitting estimator, where we split the $n$ observations into two even folds and $\hat w^{\langle i \rangle },\,\hat q^{\langle i \rangle }$ are shared by all points $i$ in the same fold and are estimated on data only on the opposite fold. {\newedit The specific steps of the cross-fitting procedure are as in \cref{sec:nmdp}. Namely, we have four estimators: $\hat w^{[0]},\,\hat q^{[0]},\,\hat w^{[1]},\,\hat q^{[1]}$.
The first two are fit on one half of the data and the latter two on the other, and $\hat w^{\langle i \rangle },\hat q^{\langle i \rangle }$ are set to those fit on the half not containing $i$.
Unless otherwise specified, we always refer to the cross-fitting version.}

The key to showing efficiency of $\hat{\rho}_{\mathrm{DRL(\cm_3)}}$ is establishing the doubly robust (or, mixed bias) structure of $\psi(s,a,r,s';w',q')$, namely, that its expectation remains $\rho^{\epol}$ whether just $w'=w$ \emph{or} just $q'=q$.
Suppose that $q'=q$. Then,
\begin{align}
    \E[\P_n[\psi(s,a,r,s';w',q)]]\nonumber
    &=(1-\gamma)\rE_{p_{\pi_e}^{(0)}}\bracks{v(s_0)}+\rE_{p_{\pi_b}}[w'(s)\eta(s,a)\{r-q(s,a)+\gamma v(s')\}]\\\label{eq:dbm3_q}
    &=(1-\gamma)\rE_{p_{\pi_e}^{(0)}}\bracks{v(s_0)}=\rho^{\epol}.
\end{align}
Heuristically, this suggests that if $\hat q^{\langle i \rangle }\to q$ and $\hat w^{\langle i \rangle }\to w'$, where generally $w'\neq w$, then we expect that $\hat{\rho}_{\mathrm{DRL(\cm_3)}}\to\rho^{\epol}$. %
This viewpoint paints the estimator $\hat{\rho}_{\mathrm{DRL(\cm_3)}}$ as given by taking the direct method and adding a control variate term. 

On the other hand, 
if $w'=w$, 
then we have that
{\newedit 
\begin{align}
   \E[\P_n[\psi(s,a,r,s';w,q')]]
    &=\rE_{p_{\pi_b}}\left[w(s)\eta(s,a)r\right]+ \rE_{p_{\pi_b}}[w(s)\{-\eta(s,a)q'(s,a)+\gamma \eta(s,a)v'(s')\}]\nonumber\\&\phantom{=}+(1-\gamma)\rE_{p_{\pi_e}^{(0)}}\bracks{v'(s_0)} \label{eq:dbm3_1}\\
    &= \rE_{p_{\pi_b}}\left[w(s)\eta(s,a)r\right]
    + \rE_{p_{\pi_b}}[w(s)\{-\eta(s,a)q'(s,a)+v'(s)\}]\label{eq:dbm3_2}\\
    &=\rE_{p_{\pi_b}}\left[w(s)\eta(s,a)r\right]=\rho^{\epol}. \label{eq:dbm3_w}
\end{align}
}
Note that from \cref{eq:dbm3_1} to \cref{eq:dbm3_2}, we have used that for any $f_{w}(s)$ (see \cref{lem:ratio-estimation}): 
\begin{align*}
    \rE_{p_{\pi_b}}[\gamma w(s)\eta(s,a)f_{w}(s')-w(s)f_{w}(s)]+(1-\gamma)\rE_{p_{\pi_b}^{(0)}}[f_{w}(s)]=0
\end{align*}
Heuristically, this suggests that if $\hat w^{\langle i \rangle }\to w$ and $\hat q^{\langle i \rangle }\to q'$, where generally $q'\neq q$, then we expect that $\hat{\rho}_{\mathrm{DRL(\cm_3)}}\to\rho^{\epol}$. %
Together, \cref{eq:dbm3_q,eq:dbm3_w} show that $\P_n[\psi(s,a,r,s';w',q')]$ has zero G\^ateaux derivative in $w',q'$ in any direction at $w'=w,q'=q$, a property known as Neyman orthogonality \citep{ChernozhukovVictor2018Dmlf}.

We now proceed to prove formally the efficiency and double robustness of our estimator. Note the L2 error such as $\|\hat w^{[j]}-w\|_2$ is defined on the offline data $p_{\pi_b}(s,a,r,s')$. 

\begin{theorem}[Efficiency of $\hat{\rho}_{\mathrm{DRL(\cm_3)}}$ under transition sampling: cross-fitting]\label{thm:m3_inf}
{\neweditf Define $\kappa^w_n,\kappa^q_n$ such that $\|\hat{w}^{[j]}-w\|_2\leq\kappa^w_n$ and $\|\hat{q}^{[j]}-q\|_2\leq\kappa^q_n$ for $j=0,1$.
Assume (\ref{thm:m3_inf}a) there exists constants $C_w,C_{S'}>0$ such that $w\leq C_w$ and $p_{b,S'}(\cdot)/p_{b,S}(\cdot)\leq C_{S'}$, where $p_{b,S'}(\cdot)$ and $p_{b,S}(\cdot)$  are marginal densities of $p_{\pi_b}(s,a,r,s')$ wrt $s'$ and $s$, (\ref{thm:m3_inf}b) 
$0\leq\hat q^{[j]}\leq (1-\gamma)^{-1}R_{\max}$ and $0\leq\hat w^{[j]}\leq C_w$ for $j=0,1$, 
(\ref{thm:m3_inf}c) $\kappa^w_n\vee \kappa^q_n=\op(1)$, and (\ref{thm:m3_inf}d) $\kappa^w_n\kappa^q_n=\op(n^{-1/2})$.
Then, $\hat{\rho}_{\mathrm{DRL(\cm_3)}}$ is RAL and efficient; in particular, $\sqrt{n}(\hat{\rho}_{\mathrm{DRL(\cm_3)}}-\rho^{\epol})\stackrel{d}{\rightarrow} \mathcal{N}(0,\mathrm{EB}({\cm_3}))$. }
\end{theorem}

The result essentially follows by showing that $|\hat{\rho}_{\mathrm{DRL(\cm_3)}}-\P_n[\psi(s,a,r,s';w,q)]|= \Op(\kappa^w_n\kappa^q_n)+\op(n^{-1/2})$. Under the above rate assumptions, the right-hand side is $\op(n^{-1/2})$ and the result is immediately concluded from the central limit theorem (CLT). %
Here, using cross-fitting, we are able to completely avoid any restriction on our plug-in estimators, except for requiring a slow rate. In particular, the rate can be \emph{subparametric}, that is, slower than square-root. Crucially, this allows us to potentially use any nonparametric black-box machine learning method, whether we can ensure good metric entropy conditions or not. 

The adaptive version requires additional metric entropy conditions on the estimators. Let $\mathcal{N}(\tau,\Fcal,\|\cdot\|_{\infty})$ be the $\tau$-covering number of $\Fcal$ wrt $L_{\infty}$ norm.
\begin{theorem}[Efficiency of $\hat{\rho}_{\mathrm{DRL(\cm_3)}}$ under transition sampling: adaptive]\label{thm:m3_inf_adaptive}
Define $\kappa^w_n,\kappa^q_n$ such that $\|\hat{w}-w\|_2\leq\kappa^w_n$ and $\|\hat{q}-q\|_2\leq\kappa^q_n$. Suppose the conditions of \cref{thm:m3_inf} hold and that in addition 
$\hat w\in \mathcal{F}_w,\hat q\in \mathcal{F}_q$ s.t. $\log \mathcal{N}(\tau,\mathcal{F}_w,\|\cdot\|_{\infty})=O(1/\tau^2),\,\log \mathcal{N}(\tau,\mathcal{F}_q,\|\cdot\|_{\infty})=O(1/\tau^2)$.
Then, $\hat{\rho}_{\mathrm{DRL(\cm_3)}}$ is RAL and efficient; in particular, $\sqrt{n}(\hat{\rho}_{\mathrm{DRL(\cm_3)}}-\rho^{\epol})\stackrel{d}{\rightarrow} \mathcal{N}(0,\mathrm{EB}({\cm_3}))$. 
\end{theorem}

Next, we formalize the notion of double robustness, which ensures our estimate is consistent even if we inconsistently estimate one of the components.
\begin{theorem}[Double robustness of $\hat{\rho}_{\mathrm{DRL(\cm_3)}}$]\label{thm:db_m3}
Assume only conditions (\ref{thm:m3_inf}a)--(\ref{thm:m3_inf}b) of \cref{thm:m3_inf} hold.
Assume further that
$\|\hat{w}^{[j]}-w^{\dagger}\|_2=\op(1)$ and $\|\hat{q}^{[j]}-q^{\dagger}\|_2=\op(1)$ for some $w^\dagger,q^\dagger$. Then, as long as either $w^\dagger=w$ or $q^{\dagger}=q$, then we have that $\plim_{n \to \infty} \hat{\rho}_{\mathrm{DRL(\cm_3)}}=\rho^{\epol}$. 
The same holds for the adaptive version, if we further assume the metric entropy condition in \cref{thm:m3_inf_adaptive}.
\end{theorem}

\Cref{thm:db_m3} does not provide a rate or an asymptotic distribution. We next strengthen the result (and, correspondingly, the conditions) to ensure a rate. This kind of double robustness is sometimes called model double robustness because the rates needed essentially correspond to parametric estimation and therefore the conditions essentially refer to whether these parametric models are well-specified \citep{SmuclerEzequiel2019Auaf}.
\begin{theorem}[Model double robustness of $\hat{\rho}_{DRL(\cm_3)}$]\label{thm:db_n}
Assume only conditions (\ref{thm:m3_inf}a)--(\ref{thm:m3_inf}b) of \cref{thm:m3_inf} hold. If either $\|\hat{q}^{[j]}-q^{\dagger}\|_{2}=\op(1),\,\|\hat{w}^{[j]}-w\|_{2}=\bigO_{p}(n^{-1/2})$ or $\|\hat{q}^{[j]}-q\|_{2}=\bigO_{p}(n^{-1/2}),\,\|\hat{w}^{[j]}-w^{\dagger}\|_{2}=\op(1)$ holds, then $\hat{\rho}_{\mathrm{DRL(\cm_3)}}=\rho^{\epol}+\Op(n^{-1/2})$. 
The same holds for the adaptive version, if we further assume the metric entropy condition in \cref{thm:m3_inf_adaptive}.
\end{theorem}

\begin{remark}[Unknown behavior policy]\label{remark:unknownbehaviorpolicy}
All of results are easily extended to the case where the behavior policy is unknown by replacing $\hat{w}(s)\eta(s,a)$ with $\hat{w}(s)\hat{\eta}(s,a)$, where $\hat{\eta}(s,a)$ is some estimator for $\eta(s,a)$, \eg, $\epol(a|s)/\hat{\pi}^{b}(a|s)$, where $\hat{\pi}^{b}(a|s)$ is some estimator for the behavior policy. 
All of the results stay the same where conditions on $\|\hat{w}- w\|_{2}$ are simply replaced with the same conditions on $\|\hat{w}\hat{\eta}-w\eta\|_{2}=\bigO(\|\hat{w}-w\|_{2}+\|\hat{\eta}-\eta\|_{2})$ instead.
\end{remark}

\begin{remark}
After the first posted version of this paper, 
\citet{tang2019harnessing} proposed a doubly-robust-style estimator for the infinite-horizon MDP setting, which is given by taking a sample average of $\tilde \psi(s,a,r,s';\hat w,\hat v)$, where
\begin{align*}
\tilde \psi(s,a,r,s';w',v')=(1-\gamma)\rE_{p_{\pi_e}^{(0)}
}[v'(s_0)]+w'(s)\eta(s,a)\prns{r+\gamma v'(s')-v'(s)},
\end{align*}
and $\hat w,\hat v$ are adaptively estimated. {\newedit The asymptotic behavior was not fully characterized, but following our work, \citet[Theorem 19]{KallusNathan2020EEoN} proved that if we impose Donsker conditions or if we use cross-fold estimates and under appropriate estimation rates (or, even if we plug-in oracle $w,v$), we can obtain that it is asymptotically normal with variance $\var[\tilde \psi(s,a,r,s';w,v)]$. This, however, is \emph{larger} than $\operatorname{EB}(\cm_3)$ by $\rE[w^2(s)\var[\eta(s,a)\{r+\gamma v(s')\}\mid s]]$ \citep[see Section 6.3][]{KallusNathan2020EEoN}. }
That is, this estimator is \emph{not} efficient, even in ideal oracle-nuisance settings.
Moreover, it is only partially doubly robust in that it requires that $\bpol$ be well-specified. In comparison, our estimator is in fact efficient and fully doubly robust.
\end{remark}

\subsection{Efficient Estimation Under Trajectory Sampling}\label{sec:dependent}

We next study the trajectory-sampling setting and show that we can achieve the very same efficiency bound even though the transition data is dependent. All of our results apply to the asymptotic regime $T\to\infty$, where $N\geq1$ is arbitrary, bounded or growing. In particular, we can consider just a single, long trajectory ($N=1$). 
Since the data is dependent, the standard notions of regular estimation do not apply; therefore, our ``efficiency'' statements are phrased solely in terms of showing that we can achieve the same asymptotic distribution of centered normal with variance equal to the efficiency bound corresponding to iid observations from the same stationary distribution.
Indexing the data as $\{(s^{\langle j \rangle }_t,a^{\langle j \rangle }_t,r^{\langle j \rangle }_t,s'^{\langle j \rangle }_{t+1})\}_{j=1,t=0}^{N,T}$ and identifying each $(j,t)$ with a corresponding $i=1,\dots,n$, where $n=NT$, we define our estimator $\hat{\rho}_{\mathrm{DRL(\cm_3)}}=\P_N\P_T[\psi(s,a,r,s';\{\hat w^{\langle i \rangle }\}_{i=1}^n,\{\hat q^{\langle i \rangle }\}_{i=1}^n)]$. That is, the same as in \cref{eq:martingaledifference}, taking an average of $\psi$ over transitions with estimated $w$- and $q$-functions, but the transitions now are actually \emph{dependent} observations.
Because of this, we restrict our attention to the case where there is nonetheless sufficient mixing. 
We also need to be more careful when constructing cross-fitting estimates.

Letting {\newedit $x^{\langle j \rangle }_t=(s^{\langle j \rangle }_t,a^{\langle j \rangle }_t,r^{\langle j \rangle }_t,s^{\langle j+1 \rangle}_t)$, }recall that we assume that $x^{\langle j \rangle }_0,x^{\langle j \rangle }_1,\dots$ forms a stationary process for each $j=1,\dots,N$, that is, while these are dependent, the marginal distribution of each has an identical distribution. In the results below we further assume that far-apart observations are less dependent, that is, the effect of earlier states gets washed away the farther ahead we look. To measure the level of such dependence we use the standard \emph{mixing coefficients} $\alpha_m,\beta_m,\phi_m,\rho_m$, each of which measures the dependence between $x^{\langle j \rangle }_0,\dots,x^{\langle j \rangle }_t$ and $x^{\langle j \rangle }_{t+m},x^{\langle j \rangle }_{t+m+1},\dots$ using different metrics of dependence (taking worst-case over $t$). For example, $\alpha_m$ is the total variation distance between the joint distribution of the two subsequences and the product of their marginals. Since these are standard we relegate their definitions to \cref{sec:appendixdefs}. The coefficients are related via $2\alpha_m\leq\beta_m\leq \phi_m$, $4\alpha_m\leq\rho_m\leq 2\phi_m^{1/2}$, so $\alpha_m$ is weakest and $\phi_m$ is (almost) strongest \citep{bradley2005basic}.

Before we proceed to discuss feasible estimators, we show that despite dependent data, our estimating function retains its efficiency structure under sufficient mixing. 
\begin{theorem}[Efficiency structure under mixing] \label{thm:ideal_m3}
{\neweditf 
Suppose $\sum_{m=1}^\infty\alpha_m<\infty$ and $w\leq C_w$ for some $C_w>0$.
Then we have $\sqrt{NT}\prns{\P_N\P_T[\psi(s,a,r,s';w,q)]-\rho^{\epol}}\stackrel{d}{\rightarrow}\mathcal{N}(0,{\mathrm{EB}(\cm_3)})$.}
\end{theorem}
The $\alpha$-mixing condition above is used in order to invoke a stationary-process CLT \citep[Theorem 18.5.4]{ibragimov1975independent}. However, such a CLT still involves covariances across time, which would inflate the asymptotic variance.
The key structural aspect of $\P_N\P_T[\psi(s,a,r,s';w,q)]$ that enables the result is that, when we use the \emph{oracle} $q$ function, the variables being time-averaged in the second term in \cref{eq:martingaledifference} form a \emph{martingale difference sequence}, which ensures zero covariances across time. This occurs by virtue of the fact that the conditional expectation of the term inside the parentheses is zero by the definition of $q$. This essentially yields the result after some algebra. In terms of showing efficiency of a feasible (rather than oracle) estimator, what remains is to show that our estimator is equal to the above oracle average up to errors that are $o_p((NT)^{-1/2})$.

\begin{remark}[Relaxing Stationarity by Ergodicity]\label{remark:stationarity}
Assuming that $p_{\pi_b}^{(0)}$ is invariant so that $x^{\langle j \rangle }_0,x^{\langle j \rangle }_1,\dots$ is stationary is purely technical. It can easily be replaced by assuming ergodicity instead, so that the initial state distribution is irrelevant and we only approach stationarity. 
Namely, note $x^{\langle j \rangle }_0,x^{\langle j \rangle }_1,\dots$ forms a Markov chain.
If it is a positive Harris chain \citep[for definition, see][p. 231]{MeynS.P.SeanP.2009Mcas} then Proposition 17.1.6 in \citet{MeynS.P.SeanP.2009Mcas} guarantees that any CLT that holds when the initial state distribution is invariant also holds for \emph{any} initial state distribution. This is simply because ergodicity means the initial state distribution gets washed away, asymptotically. All our results in this section proceed by showing $\hat{\rho}_{\mathrm{DRL(\cm_3)}}=\P_N\P_T[\psi(s,a,r,s';w,q)]+\op((NT)^{-1/2})$ and then applying a mixing-process CLT on the dependent but stationary process in the first term. Each time, per that proposition, we can assume a positive Harris chain \emph{instead of} stationarity, let the denominator of $w$ be the invariant distribution, and define all mixing coefficients wrt the chain starting from the invariant distribution, and then this CLT will still hold and our characterizations of the asymptotic distribution of the estimator will still hold \citep[see also][Remark 6]{JonesGalinL.2004OtMc}. Since this can always be done, we focus our analysis on stationary processes for
generality. 
\end{remark}

We next analyze such feasible estimators, considering three cases: adaptive, cross-fitted with $N>1$, and cross-fitted with $N=1$.
The difficulty with the latter case is that the data consists of a single, long trajectory, so any way we split the data, we will still have some dependence between the folds, undermining the standard cross-fitting technique.
For each cross-fitting estimator, we define a segmentation of our $n$ observations into folds and estimate $w$- and $q$-functions separately in each fold.
If $N\geq2$, we can split our data into folds across trajectories. Let $\mathcal D_0,\,\mathcal D_1$ be a random even partition of $\{1,\dots,N\}$ and fit $\hat w^{[j]},\,\hat q^{[j]}$ in each fold separately (see \cref{fig:foldsN}). We then set $\hat w^{\langle i \rangle },\,\hat q^{\langle i \rangle }$ to the estimates $\hat w^{[1-j]},\,\hat q^{[1-j]}$ fit only on $\mathcal D_{1-j}$ where $j$ is such that $i=t\in \mathcal D_j$. We refer to this case as \emph{cross-trajectory-fitting}. The benefit of this approach is that we have perfect independence across the folds because trajectories are independent. {\newedit Recall that we used a similar strategy in the transition-sampling setting.} Unfortunately, this is not possible when $N=1$. In this case, we propose the following alternative. 
Let $\mathcal T_0,\,\mathcal T_1,\,\mathcal T_2,\,\mathcal T_3$ be a random even partition of $\{0,\dots,T\}$ and fit $\hat w^{[j]},\,\hat q^{[j]}$ in each fold separately (see \cref{fig:foldsT}). We then set $\hat w^{\langle i \rangle },\,\hat q^{\langle i \rangle }$ to the estimates $\hat w^{[{(j+2)\;\operatorname{mod}\;4}]},\,\hat q^{[{(j+2)\;\operatorname{mod}\;4}]}$ fit only on $\mathcal T_{(j+2)\;\operatorname{mod}\;4}$ where $j$ is such that $t\in \mathcal T_j$. Thus, we always use nuisances estimated on a fold that is not adjacent to the $t^\text{th}$ data point.
We refer to this case as \emph{cross-time-fitting}.
Although we do not have perfect independence between folds, under sufficient mixing, non-adjacent folds will be sufficiently near-independent, asymptotically.}

\begin{figure}[t!]%
\hspace{0.05\textwidth}\begin{subfigure}[b]{0.4\textwidth}%
\hspace{-0.15\textwidth}\begin{tikzpicture}
    \matrix (M) [matrix of nodes,
        nodes={minimum height = 2cm, minimum width = 5cm, outer sep=0, anchor=center, draw},
        row 1/.style={nodes={draw=none,minimum height = 0.4cm}},
        column 1/.style={nodes={draw=none,minimum width = 1.4cm}},
        row sep=-\pgflinewidth, column sep=-\pgflinewidth, nodes in empty cells,
        e/.style={fill=yellow!10}
      ]
      {
        &\\
        &$\mathcal D_0$\\
        &$\mathcal D_1$\\      
      };
      \node[fit=(M-1-2)(M-1-2)]{$t=0,$\hfill$\dots$\hfill$T$};
      \node[fit=(M-2-1)(M-3-1)]{\raisebox{-4cm}{
      \rotatebox[origin=r]{-90}{\parbox{4cm}{
      \rotatebox[origin=r]{90}{$j=1,$}
      \hfill$\dots$\hfill\rotatebox[origin=r]{90}{$N$}
      }}}
      };
  \end{tikzpicture}
  \caption{Two folds over $N\geq2$ trajectories.}\label{fig:foldsN}
  \end{subfigure}\hspace{0.1\textwidth}\begin{subfigure}[b]{0.4\textwidth}%
  \hspace{-0.05\textwidth}\begin{tikzpicture}
    \matrix (M) [matrix of nodes,
        nodes={minimum height = 1cm, minimum width = 1.2cm, outer sep=0, anchor=center, draw},
        row 1/.style={nodes={draw=none,minimum height = 0.4cm}},
        column 1/.style={nodes={draw=none,minimum width = 1.4cm}},
        row 3/.style={nodes={draw=none}},
        row 4/.style={nodes={draw=none}},
        row 5/.style={nodes={draw=none}},
        row sep=-\pgflinewidth, column sep=-\pgflinewidth, nodes in empty cells,
        e/.style={fill=yellow!10}
      ]
      {
        &&&&\\
        $N=1$&$\mathcal T_0$&$\mathcal T_1$&$\mathcal T_2$&$\mathcal T_3$\\
        \\
        \\
        \\
      };
      \node[fit=(M-1-2)(M-1-5)]{$t=0,$\hfill$\dots$\hfill$T$};
  \end{tikzpicture}\caption{Four folds over a single trajectory.}\label{fig:foldsT}\end{subfigure}%
  \caption{Arrangement of folds for cross-fitting of nuisances for DRL in $\cm_3$.}%
\end{figure}

{\blockedit First, we analyze the case of the cross-trajectory-fitting version, where we can avoid complex metric entropy assumptions by virtue of the unique structure of our estimator.
\begin{theorem}[Efficiency of $\hat{\rho}_{\mathrm{DRL(\cm_3)}}$ with cross-trajectory-fitting]\label{thm:m3_eff_cro}
Define $\kappa^w_n,\kappa^q_n$ such that $\|\hat{w}^{[j]}-w\|_2\leq\kappa^w_n$ and $\|\hat{q}^{[j]}-q\|_2\leq\kappa^q_n$ for $j=0,1$.
Assume 
(\ref{thm:m3_eff_cro}a) $\sum_{k=1}^{\infty}\rho_k<\infty$,
{\neweditf (\ref{thm:m3_eff_cro}b) $w\leq C_w$ for some $C_w>0$,}
(\ref{thm:m3_eff_cro}c) $0\leq\hat q^{[j]}\leq (1-\gamma)^{-1}R_{\max}$ and $0\leq\hat w^{[j]}\leq C_w$,
(\ref{thm:m3_eff_cro}d) $\kappa^w_n\vee \kappa^q_n=\op(1)$, 
(\ref{thm:m3_eff_cro}e) $\kappa^w_n\kappa^q_n=\op(n^{-1/2})$.
Then, $\sqrt{NT}(\hat{\rho}_{\mathrm{DRL(\cm_3)}}-\rho^{\epol})\stackrel{d}{\rightarrow}\mathcal{N}(0,{\mathrm{EB}(\cm_3)})$. 
\end{theorem}
Notice that the condition (\ref{thm:m3_eff_cro}a) is slightly stronger than the mixing condition in \cref{thm:ideal_m3}. The other conditions match \cref{thm:m3_inf}.

Cross-trajectory-fitting is only feasible for $N\geq 2$ (although $N$ need not grow). 
If $N=1$, we instead proposed cross-time-fitting, which we analyze next.

{\newedit \begin{theorem}[Efficiency of $\hat{\rho}_{\mathrm{DRL(\cm_3)}}$ with cross-time-fitting]\label{thm:m3_eff_cro2}
Define $\kappa^w_n,\kappa^q_n$ such that $\|\hat{w}^{[j]}-w\|_2\leq\kappa^w_n$ and $\|\hat{q}^{[j]}-q\|_2\leq\kappa^q_n$ for $j=0,1,2,3$.
Assume 
(\ref{thm:m3_eff_cro2}a) $\phi^{1/2}_t=O(1/t^{2+\epsilon})$ for some $\epsilon>0$,
(\ref{thm:m3_eff_cro2}b) $w\leq C_w$ {\neweditf for some $C_w>0$},
(\ref{thm:m3_eff_cro2}c) $0\leq\hat q^{[j]}\leq (1-\gamma)^{-1}R_{\max}$ and $0\leq\hat w^{[j]}\leq C_w$,
(\ref{thm:m3_eff_cro2}d) $\kappa^w_n\vee \kappa^q_n=\op(1)$, 
(\ref{thm:m3_eff_cro2}e) $\kappa^w_n\kappa^q_n=\op(n^{-1/2})$.
Then, $\sqrt{NT}(\hat{\rho}_{\mathrm{DRL(\cm_3)}}-\rho^{\epol})\stackrel{d}{\rightarrow}\mathcal{N}(0,{\mathrm{EB}(\cm_3)})$.  
\end{theorem}
}
In both \cref{thm:m3_eff_cro,thm:m3_eff_cro2} we are able to avoid strong conditions on the plug-in estimators we use aside from requiring a slow, subparametric convergence rate. We only require slightly stronger mixing conditions than the oracle case in \cref{thm:ideal_m3}.

Finally, for the adaptive version of our estimator, we need to control the metric entropy of our plug-in estimators. In particular, we suppose that we are given some class $\mathcal F_\psi$ that almost surely contains $\psi(\cdot,\cdot,\cdot,\cdot;\hat w,\hat q)$. We let $J_{[]}(\infty,\mathcal F_\psi, L_p)$ be the bracketing integral wrt the $L_p$ norm \citep[for definition, see][p.~17]{KosorokMichaelR2008ItEP}. 
{\newedit 
\begin{theorem}[Efficiency of $\hat{\rho}_{\mathrm{DRL(\cm_3)}}$ with in-sample fitting]\label{thm:m3_eff}
Define $\kappa^w_n,\kappa^q_n$ such that $\|\hat{w}-w\|_2=\kappa^w_n$ and $\|\hat{q}-q\|_2=\kappa^q_n$ and fix some $p>2$.
Assume 
(\ref{thm:m3_eff}a) $\sum_{m=1}^\infty m^{2/(p-2)}\beta_m<\infty$,
(\ref{thm:m3_eff}b) $w\leq C_w$ {\neweditf for some $C_w>0$},
(\ref{thm:m3_eff}c) $0\leq\hat q\leq (1-\gamma)^{-1}R_{\max}$ and $0\leq\hat w\leq C_w$,
(\ref{thm:m3_eff}d) $\kappa^w_n\vee \kappa^q_n=\op(1)$, 
(\ref{thm:m3_eff}e) $\kappa^w_n\kappa^q_n=\op(n^{-1/2})$,
(\ref{thm:m3_eff}f) $J_{[]}(\infty,\mathcal F_\psi, L_p(p_{\pi_b}^{\infty}))<\infty$.
Then, $\sqrt{NT}(\hat{\rho}_{\mathrm{DRL(\cm_3)}}-\rho^{\epol})\stackrel{d}{\rightarrow}\mathcal{N}(0,{\mathrm{EB}(\cm_3)})$.  
\end{theorem}
}
To prove this, we invoke a uniform central limit theorem for $\beta$-mixing sequences \citep[Theorem 11.24]{KosorokMichaelR2008ItEP}. 
Because of in-sample fitting, we require
condition (\ref{thm:m3_eff}f) in order to control a term corresponding to a stochastic equicontinuity condition.
}

{\newedit 
\begin{remark}[When Stationarity Fails]\label{rem:finite} In this section, we assumed the data is stationary, or at least eventually stationary as in \cref{remark:stationarity}. But such may not apply to problems with absorbing states, as we study in \cref{sec: cart pole}. But even without stationarity, we can still view the data as transitions $(s^{\langle i \rangle },a^{\langle i \rangle },r^{\langle i \rangle },s'^{\langle i \rangle })$, $i=1,\dots,NT$, drawn (non-independently) from: 
\begin{align*}
    \prns{\frac1T\sum_{t=1}^T p^{(t)}_{b}(s,a)}  p(r|s,a)p(s'|s,a). 
\end{align*}
If the effective state-action distribution ${\frac1T\sum_{t=1}^T p^{(t)}_{b}(s,a)}$ has good coverage and $N\to\infty$ we should still expect convergence, and our DRL estimator is still using the ``best'' estimating function in the sense that it is still the least-norm gradient of the estimand, as a function of the $T$-long trajectories. Nonetheless, due to the dependence of transitions in the same trajectory and without stationarity and mixing, it is difficult to theoretically characterize the rate of the MSE in $T$.
\end{remark}
}

The remaining question is how to consistently estimate $q$ and $w$, especially from a single trajectory. We discuss
how to estimate $w$ in \cref{sec:density}
and how to estimate $q$ in 
\cref{sec:q-learning}. {\blockedit We first discuss how our results above lend themselves directly to constructing confidence intervals.

\section{Asymptotically valid confidence intervals}\label{sec:inference}

We are often interested in confidence intervals in addition to point estimates. Our asymptotic normality results lend themselves directly to the construction of such. Namely, all we have to do is consistently estimate the asymptotic variance. If an estimator $\hat\rho_n$ satisfies $\sqrt{n}(\hat\rho_n-\rho^{\epol})\to\mathcal N(0,V)$ and we have a consistent variance estimator $\hat V_n\to V$ then we will always have that $\P\prns{\abs{\hat\rho_n-\rho^{\epol}}\leq \Phi^{-1}(1-\alpha/2)\sqrt{\hat V/n}}\to1-\alpha$, where $\Phi^{-1}$ is the inverse cumulative distribution function of the standard normal (\eg, for $\alpha=0.05$, $\Phi^{-1}(1-\alpha/2)\approx 1.96$). This means that the confidence interval $[\hat\rho_n-\Phi^{-1}(1-\alpha/2)\sqrt{\hat V/n},\,\hat\rho_n+\Phi^{-1}(1-\alpha/2)\sqrt{\hat V/n}]$ has asymptotic coverage exactly $1-\alpha$. By \cref{thm:m1_inf,thm:m2_inf,thm:m3_inf,thm:m3_eff_cro,thm:m3_eff_cro2,thm:m3_eff}, it then suffices to estimate $\operatorname{EB}(\cm_1),\operatorname{EB}(\cm_2),\operatorname{EB}(\cm_3)$ to construct asymptotically valid confidence intervals.

{\newedit Focusing on $\operatorname{EB}(\cm_3)$ and the transition-sampling setting, we propose the following estimator:
\begin{align*}
\widehat{\operatorname{EB}}(\cm_3)=
\P_n[(\psi(s,a,r,s';\{\hat w^{\langle i \rangle }\}_{i=1}^n,\{\hat q^{\langle i \rangle }\}_{i=1}^n)-\hat{\rho}_{\mathrm{DRL(\cm_3)}})^2],
\end{align*}
that is, the sample variance of $\psi(s^{\langle i \rangle },a^{\langle i \rangle },r^{\langle i \rangle },s'^{\langle i \rangle };\hat w^{\langle i \rangle },\hat q^{\langle i \rangle })$. This estimate is consistent under the same conditions as in \cref{thm:m3_inf}:
\begin{theorem}\label{thm:variance est}
Under the conditions of \cref{thm:m3_inf},
$$
\widehat{\operatorname{EB}}(\cm_3)\stackrel{p}{\rightarrow}{\operatorname{EB}}(\cm_3).
$$
\end{theorem}
}

A similar result holds in $\cm_1$ and $\cm_2$. In each case, our estimators, $\hat{\rho}_{\mathrm{DRL(\cm_1)}}$ and $\hat{\rho}_{\mathrm{DRL(\cm_2)}}$, were constructed as sample averages of cross-fitted estimates of the corresponding EIF plus the estimand. {\newedit Taking the sample variance corresponding to this sample average, we again obtain a consistent variance estimator that we can use to construct asymptotically valid confidence intervals.}

Note that since our estimators are efficient, one cannot improve on the above confidence intervals, asymptotically. {\newedit More formally, a test based on an efficient estimator is automatically locally uniformly powerful in the sense that the power function defined in a neighborhood of the true data-generating process attains the upper bound (see \citealp[Lemma 25.45]{VaartA.W.vander1998As}).  
}
}

\section{Modeling the Ratio of Average Visitation Distributions}\label{sec:density}

Our DRL estimator in $\cm_3$ relied on having an estimator for the ratio of average visitation distributions, $w(s)$. In this section, we discuss its estimation from semiparametric inference perspective. These estimates can then be plugged into $\hat\rho_{\mathrm{DRL(\cm_3)}}$.

\subsection{Importance Sampling Using Stationary Density Ratios}\label{sec:isestmdp}

{\blockedit Before discussing how to estimate $w(s)$, we consider an IS-type estimator for MDPs using $w(s)$. We can transform our DRL estimator to an IS-type estimator by simply choosing $\hat q^{[I]}=0$. This leads to the marginalized importance sampling (MIS) estimator 
\begin{align}{\label{eq:m3_is}}
   \hat{\rho}_{\mathrm{MIS}}= \P_{n}\left [\eta(s,a)\hat{w}(s) r\right] ,\quad \hat w(s)\approx w(s).
\end{align}
where ``$\approx$'' above means ``estimating.'' 
Note that this is different from the IS estimator proposed by \cite{Liu2018}, which is defined as an empirical approximation of
\begin{align}{\label{eq:m3_is_li}}
    \rE_{p^{(\infty)}_{\bpol,\gamma}}\left [\eta(s,a)\hat{\tilde{w}}(s) r\right],\quad\hat{\tilde{w}}(s)\approx\tilde{w}(s) = \frac{p^{(\infty)}_{\epol,\gamma}(s)}{p^{(\infty)}_{\bpol,\gamma}(s)}.
\end{align}
The difference between the two methods is that we use $p^{(0)}_{\bpol}(s)$ instead of $p^{(\infty)}_{\bpol,\gamma}(s)$ in the denominator of the density ratio. In the transition-sampling setting, $p^{(0)}_{\bpol}(s)$ in \cref{eq:m3_is} can be \emph{anything}. In the trajectory-sampling setting, the denominator is an invariant distribution, or is the stationary distribution $p^{(\infty)}_{\bpol}(s)$ if we consider the ergodic case (see \cref{remark:stationarity}), which is still different from $p^{(\infty)}_{\bpol,\gamma}(s)$. There are a few benefits to this. Intuitively, since we see samples from $p^{(\infty)}_{\bpol}$, using \cref{eq:m3_is} can be more efficient because, to get a sample from the distribution $p^{(\infty)}_{\bpol,\gamma}$, we would essentially have to throw away $(1-\gamma)$ fraction of our samples. Indeed, the performance of \cref{eq:m3_is_li} behaves badly when $\gamma<1$ \citep[Figure 3(d)]{Liu2018}.} %

Nonetheless, unlike $\hat{\rho}_{\mathrm{DRL(\cm_3)}}$ as in \cref{sec:dependent}, the estimator $\hat{\rho}_{\mathrm{MIS}}$ does not have a martingale difference structure. This means that the covariance terms across the time in the CLT do \emph{not} drop out, potentially inflating the variance of the $\P_T$ average in the trajectory-sampling setting. Moreover, because it lacks a doubly robust structure, there is an inflation term due to the plug-in of an estimate, $\hat w$, of $w$, unlike $\hat{\rho}_{\mathrm{DRL(\cm_3)}}$. This occurs even if the estimate has a parametric rate, $\|\hat{w}-w\|_{2}=\bigO_{p}(n^{-1/2})$, because there is no mixed-bias structure to cancel it out. These two reasons make it difficult to analyze the asymptotic MSE of $\hat{\rho}_{\mathrm{MIS}}$. They also suggest the estimator is not efficient.

\subsection{Efficient Semiparametric Estimation}\label{sec:westsemi}

{\blockedit 
The remaining question is how to estimate $w(s)=p^{(\infty)}_{\epol,\gamma}(s)/p^{(0)}_{\bpol}(s)$. Here, we take a semiparametric approach. First, we consider a characterization of $w(s)$ by modifying Theorem 4 in \cite{Liu2018}. We obtain the following lemma.

\begin{lemma}[Characterization of $w(s)$]\label{lem:ratio-estimation}
{\newedit 
Define 
\begin{align} \label{eq:estimating}
    L(w',f_w)=\rE[\gamma w'(s)\eta(s,a)f_{w}(s')-w'(s)f_{w}(s)]+(1-\gamma)\rE_{p_{\pi_e}^{(0)}}[f_{w}(s)].
\end{align}
Then, for $w'=w$, we have $L(w',f_{w})=0$ for any $f_{w}$. 
Conversely, if $L(w',f_{w})=0$ for all $\lambda_{\mathcal S}$-square-integrable functions $f_{w}$ and there is a unique solution $g$ to the integral equation 
\begin{align*}
    0=\gamma \int p(s'|s)g(s)\mathrm{d}\lambda_{\mathcal S}(s)-g(s')+(1-\gamma)p^{(0)}_{\epol}(s'),
\end{align*}
then, $w'(s)=w(s)$. 
}
\end{lemma}
Again, this holds for \emph{any} $p_{\pi_b}^{(0)}(s)$. This is the difference from  \citet[Theorem 4]{Liu2018}, which only holds for $p_{\pi_b}^{(0)}(s)=p^{(\infty)}_{\bpol,\gamma}(s)$. 
When $p_{\pi_b}^{(0)}(s)$ is an invariant distribution, as in the trajectory-sampling setting, $L(w',f_w)$ is equal to 
\begin{align}\label{eq:estimating2}
    \rE[\gamma w'(s)\eta(s,a)f_{w}(s')-w'(s')f_{w}(s')]+(1-\gamma)\rE_{p_{\pi_e}^{(0)}}[f_{w}(s)].
\end{align}
Thus, in this case, the condition that $L(w',f_w)=0$ for all $f_w$ is equivalent to the conditional moment equation.: 
\begin{align}\label{eq:inf_gamma1}
    \rE\bracks{{w(s)\eta(s,a)-w(s')}+(1-\gamma)\frac{p_{\pi_e}^{(0)}(s')}{p_{\pi_b}^{(0)}(s')}\mid s'}=0. 
\end{align} 
Note this is not a standard moment equation since it still depends on the unknown quantity $p_{\pi_b}^{(0)}(s)$.
}
This is closely related to a similar key relation of $\mu_{k}(s_k)$ used in \cref{sec:tmdp}, namely, $\rE[\nu_{k-1}\mid s_k]=\mu_{k}(s_k)$, which implies
\begin{align}\label{eq:recursive}
    \rE[\mu_{k-1}(s_{k-1})\eta(a_{k-1},s_{k-1})-\mu_{k}(s_k)\mid s_k]=0. 
\end{align}
{\newedit For derivation, refer to \citet[Section 3]{NathanUehara2019}.}
Heuristically,
taking a limit as $k\to \infty$, replacing $\lim_{k\to \infty}\mu_{k}(s)$ with $w(s)$, and setting $\gamma=1$, we get \cref{eq:inf_gamma1}. 
Notice that in \cref{eq:recursive}, we obtain $\mu_k$ from $\mu_{k-1}$, whereas in \cref{eq:inf_gamma1} we obtain $w$ from itself, \ie, it solves a fixed-point equation.
This change is analogous to the change in $q$-equations between the time-variant finite-horizon problem and the time-invariant infinite-horizon problem. 

Suppose first that we assume a parametric model $w(s)=w(s;\beta^*)$. Then, $\beta^*$ can be estimated as a solution to an empirical approximation of \cref{eq:estimating}, that is, 
\begin{align} \label{eq:estimating2_w}
\edit{\P_n[\gamma w(s;\beta)\eta(s,a)f_{w}(s')-w(s;\beta)f_w(s)]+(1-\gamma)\rE_{p_{\pi_e}^{(0)}}[f_w(s)]=0,}
\end{align}
for some \emph{vector}-valued function $f_w$. We denote the estimator as $\hat{\beta}_{f_w}$. Note $\rE_{p_{\pi_e}^{(0)}}[f_w(s)]$ can be exactly calculated because $p_{\pi_e}^{(0)}$ is known. 

\begin{example}[Linear regression approach] \label{ex:linear1}
Consider a case when our model is linear in some features of $s$, \ie, $w(s;\beta)=\beta^{\top}\psi(s)$. Then, as in linear regression, a natural choice for $f_{w}(s)$ is $\psi(s)$. The estimator of $\hat\beta_\psi$ is constructed as the solution to
\begin{align*}
   \edit{\frac{1}{n}\sum_{i=1}^n\psi(s^{\langle i \rangle })\left(\gamma \eta(s^{\langle i \rangle },a^{\langle i \rangle }) \psi^{\top}(s'^{\langle i \rangle })-  \psi^{\top}(s^{\langle i \rangle })\right)\beta + (1-\gamma)\rE_{p_{\pi_e}^{(0)}}[\psi(s)]=0.}
\end{align*}
In the finite-state-space setting, we can use $\psi(s)=(\mathrm{I}(s^{*1}=s),\cdots,\mathrm{I}(s^{*d}=s))^{\top}$, where $\mathcal{S}=\{s^{*1},\cdots,s^{*d}\}$. 
\end{example}

{\blockedit More generally, for a linear or non-linear model, under the correct specification assumption, that is, there exists $\beta^{*}$ such that $w(s)=w(s;\beta^{*})$, we have the following efficient estimation result. 
{\newedit We focus on the transition-sampling setting.  }
}
{\newedit
\begin{theorem}[Efficient estimation of $w(s;\beta^*)$ under transition-sampling]
\label{thm:bound_ratio}
Define 
\begin{align*}
    \Delta_{f_w}(s,a,s';\beta)= w(s;\beta)\{\gamma \eta(s,a)f_w(s')-f_w(s)\}.  
\end{align*}
Suppose $\E\sup_{\beta\in \Theta_{\beta}}\|\Delta_{f_w}(s,a,s';\beta)\|<\infty$, where $\Theta_{\beta}$ is a parameter space for $\beta$. Assume $w(s)=w(s;\beta^*)$ for some $\beta^*\in\Theta_\beta$ and that a vector-valued $f_w$ is given such that $L(w(s;\beta),f_w)=0\iff\beta=\beta^*$.  
Further assume standard regularity conditions: $\Theta_\beta$ is compact, $\beta^{*}$ is in its interior, $w(s;\beta)$ is a $C^{2}$-function with respect to $\beta$ with first and second derivatives uniformly bounded, and for any $\alpha$ with $\|\alpha\|=1$ we have $\rE[|\alpha^{\top}\Delta_{f_w}(s,a,s';\beta)  |^{2+\epsilon}]\big|_{\beta=\beta^{*}}<\infty$ for some $\epsilon>0$. Then, the asymptotic variance of $\hat\beta_{f_w}$ is 
\begin{align*}
   \rE[\nabla_{\beta^{\top}}\Delta_{f_w}(s,a,s': \beta)]^{-1}\var[\Delta_{f_w}(s,a,s': \beta)]\{\rE[\nabla_{\beta^{\top}}\Delta_{f_w}(s,a,s': \beta)]^{\top}\}^{-1}\vert_{\beta=\beta^{*}}. 
\end{align*}
\end{theorem}
}

Importantly, regardless of the choice of $f_{w}$, the rate of $\|{w}(s;\hat \beta_{f_w})-w(s)\|_{2}$ will be $\bigO_{p}(n^{-1/2})$. {\newedit Compared to the usual conditional moment equation setting \citep{ChenXiaohong2007C7LS}, the efficient choice of $f_w$ to minimize the asymptotic variance here is unclear because $p^{(0)}_b(s)$ is unknown.}

Because of the doubly robust structure of $\hat\rho_{\mathrm{DRL(\cm_3)}}$, it did not matter how we estimated $w$ as long as we had a (subparametric) rate. This is not true for $\hat{\rho}_{\mathrm{MIS}}$. We can, however, derive its asymptotics for the particular estimation approach above. 
{\newedit 
\begin{theorem}[Asymptotic property of $\hat{\rho}_{\mathrm{MIS}}$]\label{thm:bound_ratio2}
Suppose the conditions of \cref{thm:bound_ratio} hold and that $\bG_{n}[r\eta(s,a)w(s;\hat{\beta}_{f_w})]-\bG_{n}[r\eta(s,a)w(s;\beta^{*})] =\op(1)$, where $\bG_{n}=\sqrt{n}(\P_n-\rE)$ is the empirical process. Then, $\sqrt{n}(\hat{\rho}_{\mathrm{MIS}}-\rho^{\epol})\stackrel{d}{\rightarrow}\mathcal{N}(0,V_{\mathrm{MIS}})$ where
\begin{align}\label{eq:is1}
V_{\mathrm{MIS}}=\mathrm{var}[w(s;\beta)\eta(s,a)r+\rE[\nabla_{\beta^{\top}}w(s;\beta)\eta(s,a)r]\rE[\nabla_{\beta^{\top}}\Delta_{f_w}(s,a,s';\beta)]^{-1}\Delta_{f_w}(s,a,s';\beta)]\vert_{\beta=\beta^{*}}.
\end{align}
\end{theorem}
}

Note the technical condition {\newedit $\bG_{n}[r\eta(s,a)w(s;\hat{\beta}_{f_w})]-\bG_{n}[r\eta(s,a)w(s;\beta^{*})] =\op(1)$} can potentially be verified as in the proofs of \cref{thm:m3_eff,thm:m3_eff_cro}.

\section{Modeling the $q$-function}\label{sec:q-learning}

In this section, we discuss from a semiparametric inference perspective how to estimate the $q$-function in an off-policy manner, potentially from only one trajectory. Our approach can be seen as a generalization of LSTDQ \citep{LagoudakisMichail2004LPI}. The estimated $q$-function we obtain can be used in our estimator, $\hat{\rho}_{\mathrm{DRL(\cm_3)}}$. 

{\blockedit By definition, the $q$-function is characterized as a solution to 
\begin{align*}
   q(s,a)=\mathrm{E}[r\mid s,a]+\gamma \mathrm{E}[\mathrm{E}_{a'\sim \epol}[ q(s',a')\mid s'] \mid s,a]. 
\end{align*}

Assume a parametric model for the $q$-function, $q(s,a)=q(s,a;\beta)$. Then, the parameter $\beta$ can be estimated using the following recursive estimating equation:
\begin{align*}
    &\mathrm{E}\bracks{e_{\text{q}}(s,a,r,s';\beta)|s,a}=0,\\ &\text{where}\quad e_{\text{q}}(s,a,r,s';\beta)=r +\gamma \rE_{a'\sim\epol}\bracks{q(s',a';\beta)|s'}-q(s,a;\beta). 
\end{align*}
This implies that for any function $f_{q}(s,a)$, 
\begin{align}\label{eq:q-z-pop}
    \rE[f_{q}(s,a)e_{\text{q}}(s,a,r,s';\beta)]=0. 
\end{align}
More specifically, given a vector-valued $f_{q}(s,a)$, we can define an estimator $\hat{\beta}_{f_q}$ as the solution to}
\begin{align}\label{eq:q-z}
    \edit{\P_n[f_{q}(s,a)e_{\text{q}}(s,a,r,s';\beta)]=0.}
\end{align}

\begin{example}[LSTDQ]\label{ex:lstdq}
When $q(s,a;\beta)=\beta^{\top}\psi(s,a)$ and $f_{q}(s,a)=\psi(s,a)$, this leads to the LSTDQ method \citep{LagoudakisMichail2004LPI}:
\begin{align*}{\blockedit
 \left(\sum_{i=1}^{n}\psi(s^{\langle i \rangle },a^{\langle i \rangle })[\psi^{\top}(s^{\langle i \rangle },a^{\langle i \rangle })-\gamma \mathrm{E}_{a \sim \epol}\{\psi^{\top}(s'^{\langle i \rangle },a)|s^{\langle i \rangle }\}]\right)^{-1}
    \left\{\sum_{i=1}^{n}r^{\langle i \rangle }\psi(s^{\langle i \rangle },a^{\langle i \rangle })\right\}=0.
}\end{align*}
\end{example}

More generally, for a linear or non-linear model,
under the correct specification assumption, that is, that there exists some $\beta^{*}$ such that $q(s,a)=q(s,a;\beta^{*})$, we have the following result. {\newedit We again focus on the transition-sampling setting. 
\begin{theorem}[Efficient estimation of $q(s,a;\beta)$ under transition sampling]
\label{thm:bound_q2}
Suppose $\E\sup_{\beta\in \Theta_{\beta}}\|e_q(s,a,r,s';\beta)f_q(s,a)\|<\infty$, where $\Theta_{\beta}$ is a parameter space for $\beta$.  Assume $q(s,a)=q(s,a;\beta)$ for some $\beta\in\Theta_\beta$ and that a vector-valued $f_q$ is given such that $(\text{\cref{eq:q-z-pop} holds})\iff\beta=\beta^*$. Further, assume standard regularity conditions: $\Theta_\beta$ is compact, $\beta^{*}$ is in its interior, $q(s,a;\beta)$ is $C^{2}$-function with respect to $\beta$ with first and second derivatives uniformly bounded, and for any $\alpha$ with $\|\alpha\|=1$ we have $\rE[|e_q(s,a,r,s';\beta)\alpha^{\top}f_q(s,a)|^{2+\epsilon}]\big|_{\beta=\beta^{*}}>0$ for some $\epsilon>0$. 
The lower bound for the asymptotic MSE for estimating $\beta^*$ scaled by $n$ is
\begin{align*}
    V_{\beta}=\rE[\nabla_{\beta}m_{q}(s,a;\beta)v^{-1}_{q}(s,a;\beta)\nabla_{\beta^{\top}}m_{q}(s,a;\beta)]^{-1}\big|_{\beta=\beta^{*}},
\end{align*}
where $m_q(s,a;\beta)=\rE[e_q(s,a,r,s';\beta)|s,a],\,v_q(s,a)=\mathrm{var}[e_q(s,a,r,s';\beta)|s,a]$. 

This bound is achieved when 
\begin{align}\label{eq:eff-ifq}
    f_{q}(s,a)= \nabla_{\beta}m_{q}(s,a;\beta)v^{-1}_{q}(s,a;\beta)\big|_{\beta=\beta^{*}}. 
\end{align}
\end{theorem}

Importantly, regardless of the choice of $f_{q}$, the rate $\|q(\cdot,\cdot;\hat{\beta}_{f_q})-q\|_{2}$ is $\bigO_{p}(n^{-1/2})$. 
Nonetheless, efficient estimation is preferred. Practically, we do not know the efficient $f_{q}$ in \cref{eq:eff-ifq}. One way is parametrically estimating it and another way is a sieve generalized method of moments (GMM) estimator, using a basis expansion for $f_{q}$ \citep{HahnJinyong1997Eeop}.}

We can also extend the approach to achieve nonparametric estimation of $q$. This is most easily done by extending the LSTDQ approach in \cref{ex:lstdq}. We simply let $q(s,a;\beta_N)=\sum_{j=1}^{d_N}\beta_j\psi_{j}(s,a)$ where $\psi_1,\psi_2,\dots$ is a basis expansion of $L^2$ and $d_N\to\infty$ as we collect more data. 
Given regularity conditions and smoothness conditions on $q$, we can obtain rates on $\|q(\cdot,\cdot;\hat\beta_N)-q\|_2$ without assuming correct parametric specification \citep{ChenXiaohong1998Seef}.
This provides a means to estimate $q$ for $\hat\rho_{\mathrm{DRL(\cm_3)}}$, either parametrically or nonparametrically.

If we use $q$ as estimated parametrically above, we can also establish the asymptotic behavior of $\hat{\rho}_{\mathrm{DM}}$. Again, as in the case of $\hat{\rho}_{\mathrm{MIS}}$, because $\hat{\rho}_{\mathrm{DM}}$ lacks the doubly robust structure, we must have parametric rates on $q$-estimation in order to achieve $1/n$ MSE scaling in the below, unlike the case of $\hat{\rho}_{\mathrm{DRL(\cm_3)}}$ where $q$-estimation can have slow nonparametric rates.
\begin{theorem}[Asymptotic property of $\hat{\rho}_{\mathrm{DM}}$]
\label{thm:bound_q2b}
Let
$\hat{\rho}_{\mathrm{DM}}=(1-\gamma)\rE_{p^{(0)}_{\epol}}[\rE_{a\sim \epol(s)}\{q(s,a;\hat{\beta}_{f_q})\mid s\}]$.
Suppose the assumptions of \cref{thm:bound_q2} hold.
Then $\sqrt{n}(\hat{\rho}_{\mathrm{DM}}-\rho^{\epol})\stackrel{d}{\rightarrow}\mathcal{N}(0,V_{\mathrm{DM}})$ where
\begin{align*}
   V_{\mathrm{DM}}=(1-\gamma)^{2} \rE_{p^{(0)}_{\epol}}[\rE_{a\sim \epol(s)}\bracks{\nabla_{\beta^{\top}}q(s,a;\beta)|s}]  V_{\beta} \rE_{p^{(0)}_{a \sim \epol(s)}}[\rE_{\epol}\bracks{\nabla_{\beta}q(s,a;\beta)|s}]\big|_{\beta=\beta^{*}}. 
\end{align*}
\end{theorem}

{\newedit\label{pageref: parameteric q comment}
Interestingly, this is smaller than or equal to the efficiency bound in $\cm_3$. This is not a contradiction since the above $\hat{\rho}_{\mathrm{DM}}$ is not regular wrt $\cm_3$ as it assumes the well-specification of the parametric model $q(s,a;\beta)$, which leads to the smaller model than $\cm_3$. 

\begin{lemma}\label{eq:relation}
$V_{\mathrm{DM}}\leq \mathrm{EB}(\cm_3)$. 
\end{lemma}
This result is well-known in the bandit setting when we use a binary deterministic policy \citep{TanZhiqiang2007CUoP}. Our result can be seen as its generalization to the more complex MDP setting.  }

 \begin{remark}\label{rem:literature}
 \citet{Ueno2011,LuckettDanielJ.2018EDTR} considered related semiparametric estimation techniques for the $v$-function. 
 Compared with that, our focus is a $q$-function estimation rather than a value function estimation. {\neweditf Note many traditional TD-type methods \citep{SuttonRichardS1998Rl:a}, including LSTD($\lambda$) \citep{NediCA2003LSPE,boyan1999least}, Gradient Temporal Difference learning (GTD) \citep{Sutton2009}, Temporal Difference learning with Gradient Correction (TDC) \citep{Sutton2009b}, and Off-Policy LSTD \citep{yu2012least} are also defined as the solution to estimating equations as in \cref{eq:q-z-pop}. For details, refer to \citet{Ueno2011,yu2018generalized}. The asymptotic MSEs of these methods can be calculated as in \cref{thm:bound_q2}.}
\end{remark}

\section{Experimental Results}\label{sec:numerical}

\edit{In this section, we conduct experiments to compare our method with existing off-policy evaluation methods. We consider a simpler setting that perfectly fits the theory and a more challenging setting that requires some function approximation.}

\subsection{Taxi Environment}

\edit{First we consider the Taxi environment and focus on simple $w$- and $q$-estimators in order to illustrate the doubly robust property of our method.} For detail on this environment, see \citet{Liu2018}.

We set our target evaluation policy to be the final policy $\epol=\pi^{*}$ after running $q$-learning for 1000 iterations. We set another policy $\pi_{+}$ as the result after 150 iterations. The behavior policy is then defined as $\bpol=\alpha \pi^{*}+(1-\alpha)\pi_{+}$, where we range $\alpha$ to vary the overlap.
We show results for $\alpha=0.2,\,0.6$ here and provide additional results for $\alpha=0.4,\,0.8$ in \cref{appendix:extraresults}. We consider the case with the behavior policy known and set $\gamma = 0.98$. Note that this $\pi^{*},\,\pi_{+}$ are fixed in each setting.

\begin{figure}[!ht]
\minipage{0.5\textwidth}%
  \includegraphics[width=\linewidth]{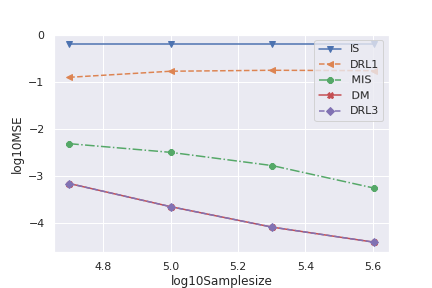}%
  \caption{Setting (1) with $\alpha=0.2$}\label{fig:2_1}%
\endminipage\hfill%
\minipage{0.5\textwidth}%
  \includegraphics[width=\linewidth]{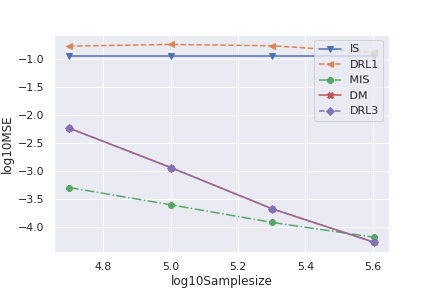}%
  \caption{Setting (1) with $\alpha=0.6$}\label{fig:6_1}%
\endminipage\\%
\minipage{0.5\textwidth}%
  \includegraphics[width=\linewidth]{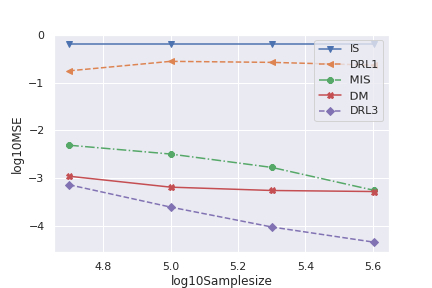}%
  \caption{Setting (2) with $\alpha=0.2$}\label{fig:2_2}%
\endminipage\hfill%
\minipage{0.5\textwidth}%
  \includegraphics[width=\linewidth]{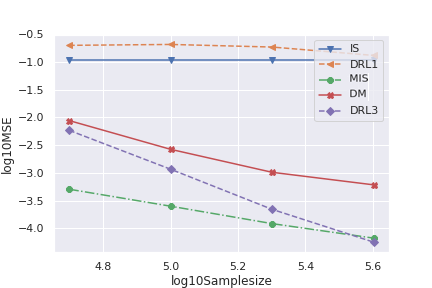}%
  \caption{Setting (2) with $\alpha=0.6$}\label{fig:6_2}%
\endminipage\\%
\minipage{0.5\textwidth}%
  \includegraphics[width=\linewidth]{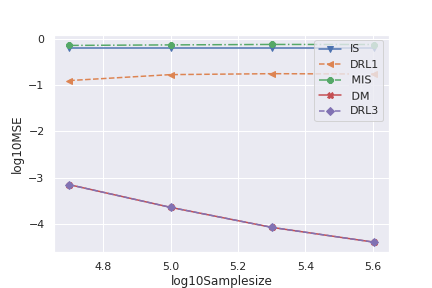}%
  \caption{Setting (3) with $\alpha=0.2$}\label{fig:2_3}%
\endminipage\hfill%
\minipage{0.5\textwidth}%
  \includegraphics[width=\linewidth]{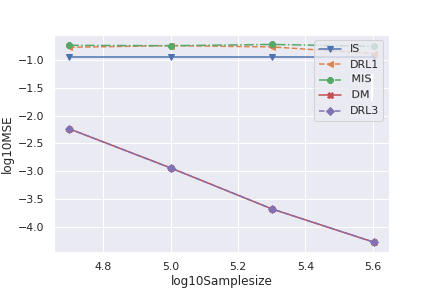}%
  \caption{Setting (3) with $\alpha=0.6$}\label{fig:6_3}%
\endminipage%
\end{figure}

We estimate all $w$-functions following \cref{ex:linear1}. For $q$-functions, we use a value iteration for the approximated MDP based on the empirical distribution. Then, we compare 
$\hat{\rho}_{\mathrm{IS}}$, $\hat{\rho}_{\mathrm{DRL(\cm_1)}}$, $\hat{\rho}_{\mathrm{MIS}}$, $\hat{\rho}_{\mathrm{DM}}$, and $\hat{\rho}_{\mathrm{DRL(\cm_3)}}$. We consider observing a single trajectory ($N=1$) of increasing length $T$, $T\in[50000,100000,200000,400000]$. For each, we consider $200$ replications. Note that we use adaptive (in-sample) fitting and not cross-fitting because $N=1$. In addition, we do not compare to a marginalized importance sampling estimator or to $\hat{\rho}_{\mathrm{DRL(\cm_2)}}$ because $\mu_t$ cannot be estimated with $N=1$ (\eg, the empirical estimated marginal importance $\hat{\mu}_t$ is just $\nu_t$). 

To study the effect of doubly robust property, we consider three settings. 
\begin{enumerate} [(1)]
    \item Both $w$-model and $q$-model are correct.
    \item Only $w$-model is correct: we add noise $\mathcal{N}(1.0,\,1.0)$ to $\hat q(s,a)$. 
    \item Only $q$-model is correct: we add noise $\mathcal{N}(1.0,\,1.0)$ to $\hat w(s)$. 
\end{enumerate}

\textbf{Results and Discussion}: We report the resulting MSE over the replications for each estimator in each setting in \cref{fig:2_1,fig:6_1,fig:2_2,fig:6_2,fig:2_3,fig:6_3}. 

First, we note that the estimator $\hat{\rho}_{\mathrm{DRL(\cm_3)}}$ handily outperforms the standard IS and DR estimators, $\hat{\rho}_{\mathrm{IS}}$, $\hat{\rho}_{\mathrm{DR}}$, in every setting. This is owed to the fact that these do not leverage the MDP structure. The competitive comparison is of course to DM and MIS.

We find that, in the large-sample regime, $\hat{\rho}_{\mathrm{DRL(\cm_3)}}$ dominates all other estimators across all settings. First, for $T=400000$, it has the lowest MSE among all estimators for each setting. Second, while in some settings it has MSE similar to another method, it beats it handily in another setting. Compared to DM, the MSE is similar when the $q$-function is well-specified but $\hat{\rho}_{\mathrm{DRL(\cm_3)}}$ does much better when $q$ is ill-specified. Compared to MIS, the MSE is similar when \emph{both} the $w$-function is well-specified \emph{and} there is good overlap but $\hat{\rho}_{\mathrm{DRL(\cm_3)}}$ performs much better when \emph{either} specification \emph{or} overlap fails. This is of course owed to the doubly robust structure and the efficiency of $\hat{\rho}_{\mathrm{DRL(\cm_3)}}$.

In the small-to-medium sample regime, $\hat{\rho}_{\mathrm{DRL(\cm_3)}}$ performs the best among all estimators \emph{except} when overlap is good ($\alpha=0.6$) and $w$ is well specified (settings (2) and (3)). In these cases, for the small-to-medium sample regime, MIS performs better. However, as in the large-sample regime, it performs much worse in small-to-medium samples too when overlap is bad or when $w$ is misspecified. In particular, in setting (2) with $\alpha=0.2$, $\hat{\rho}_{\mathrm{DRL(\cm_3)}}$ has performance much better than all other estimators across the sample-size regimes.

Because having either parametric misspecification or nonparametric rates for $\hat w$ and $\hat q$ is unavoidable in practice (for continuous state-action spaces), the estimator $\hat{\rho}_{\mathrm{DRL(\cm_3)}}$ is superior. This is doubly true when overlap can be weak.

{\blockedit\subsection{CartPole Environment}\label{sec: cart pole}

We next conduct an experiment in the CartPole environment based on the implementation of OpenAI Gym \citep{brockman2016openai}. In the CartPole environment, the state space is continuous and four-dimensional and the action space is binary. Thus, we require flexible models for $w$ and $q$ and may not be able to guarantee their precise convergence. Moreover, the environment has an absorbing state and therefore our trajectories are highly non-stationary, yet we show our method still works in practice as suggested by \cref{rem:finite}.

We set the target and behavior policy in the following way. First, we run Deep Q-Network (DQN) in an \emph{online} interaction with the environment to learn $q^*$, following OpenAI's default implementation.\footnote{\url{https://github.com/openai/baselines}.}
Then, based on $q^*$, we define a range softmax policies given by a temperature parameter $\tau$: $\pi(a\mid s:\tau)\propto \exp(Q(s,a)/\tau)$. We then set the behavior policy as $\pi_b(a\mid s)=\pi(a\mid s:1.0)$, and we consider a variety of evaluation policies $\epol(a\mid s)=\pi(a\mid s: \tau)$ for $\tau\in [0.7,0.9,1.1,1.3]$.
The training dataset is generated by executing the behavior policy with a fixed horizon length $T=1000$. Specifically, if the agent visits the terminal absorbing states before $1000$ steps, the rest of the trajectory will consist of repeating the last state.
We consider observing $N\in [50,100,200,400]$ trajectories, \ie, $n\in [50,100,200,400]\times 1000$ transitions.

We estimate $w$ using a minimax approach leveraging \cref{eq:estimating}. Namely, we consider a model $w(s;\beta)$ given a neural network with $32$ units in each, ReLU activations for hidden layers, and a softplus activation for the output to ensure nonnegative output. Then, we fit the weights $\beta$ by minimizing the maximum of the left-hand-side of \cref{eq:estimating2_w} over all $f_w$ in the unit ball of the reproducing kernel Hilbert space (RKHS) with the Gaussian kernel $k(x_i,x_j)=\exp(-\|x_i-x_j\|^2/(2\sigma^2))$. We similarly estimate $q$ leveraging \cref{eq:q-z}. We again use the same neural network architecture for $q(s,a;\beta)$ except that the input has one more dimension and we do not apply an activation to the output. We again consider $f_q$ in the same RKHS unit ball (but with one more input dimension). For both $w$- and $q$-estimation, we normalize all data to have mean zero and unit variance and set the length-scale parameter $\sigma$ to the median of pairwise distances in the data. We use Adam to optimize the neural networks and set the learning rate to $0.005$.  

We compare MIS ($\hat \rho_{\operatorname{MIS}}$), DM  ($\hat \rho_{\operatorname{DM}}$) and DRL3  ($\hat \rho_{\cm_3}$) using the above $w$- and $q$-estimators. We also these to DualDICE \citep{ChowYinlam2019DBEo}, which is a variant of the MIS estimator. In DualDICE, the $w$ estimator is based on a different minimax objective function using two neural networks. We choose hyperparameters to be the same as in the implementation of \citet{UeharaMasatoshi2020MWaQ}.

\begin{figure}[t!]
\minipage{0.5\textwidth}%
  \includegraphics[width=\linewidth]{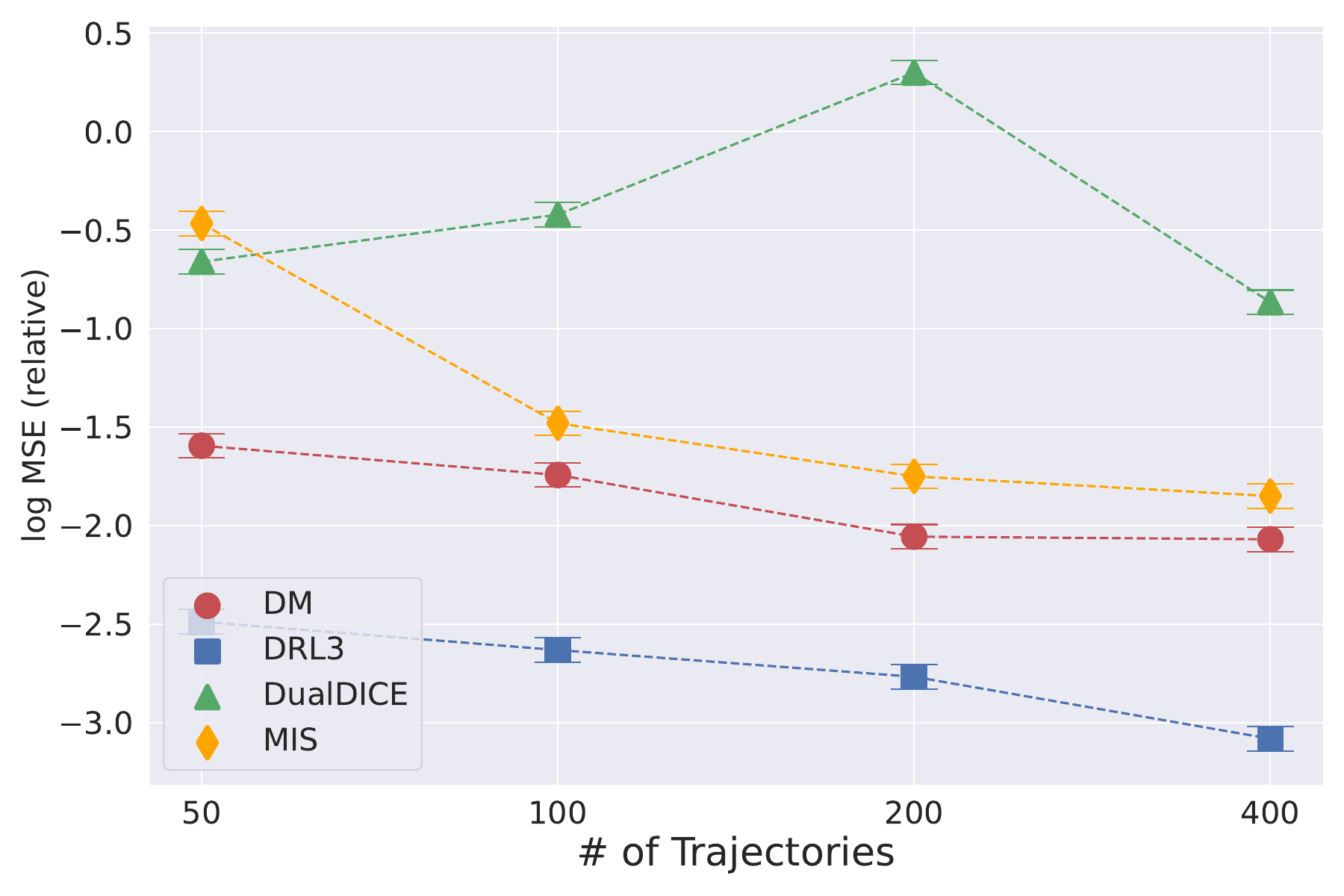}%
  \caption{CartPole: $\tau=1.3$ and $N$ varying.}\label{fig:exp1}%
\endminipage\hfill%
\minipage{0.5\textwidth}%
  \includegraphics[width=\linewidth]{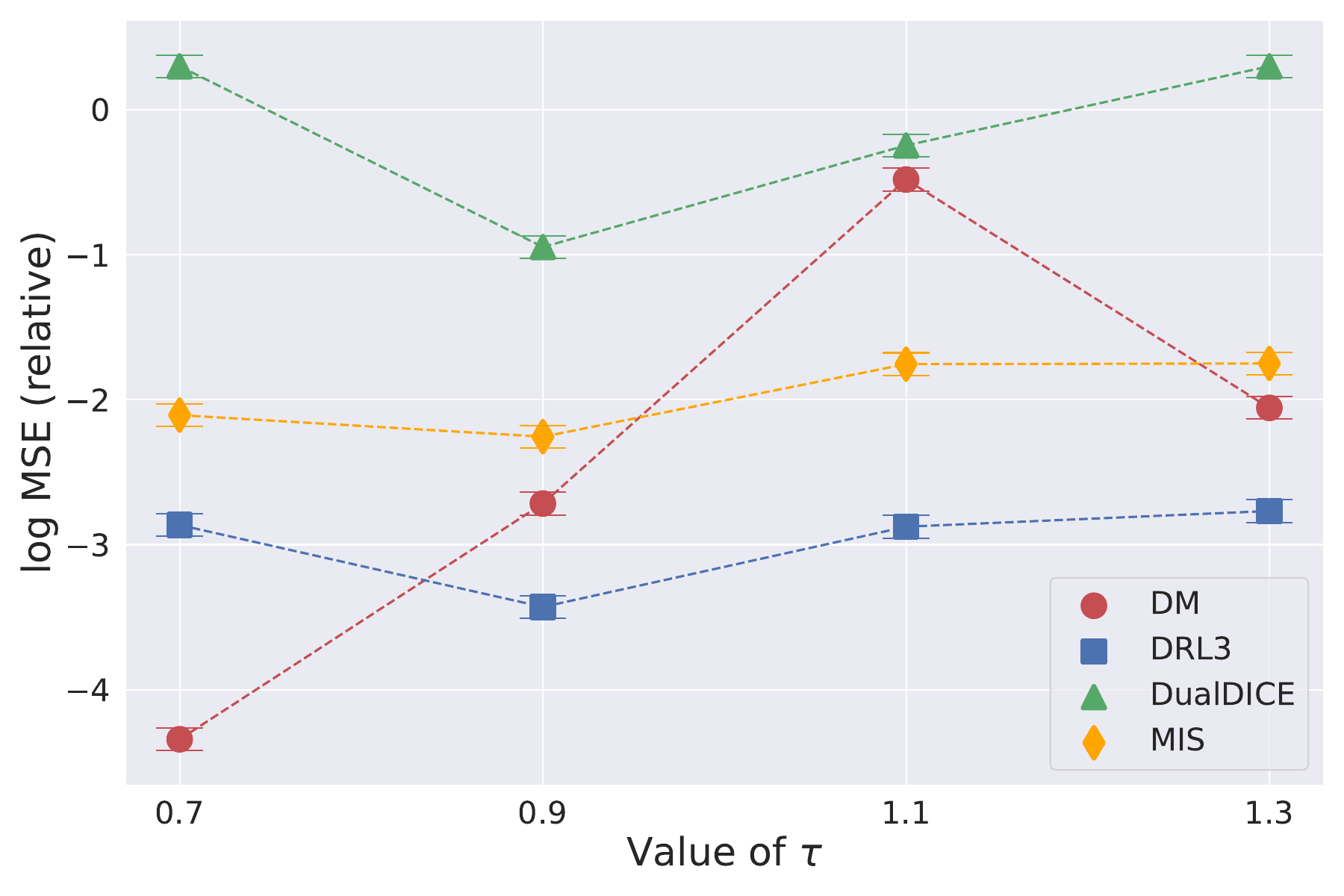}%
  \caption{CartPole: $N=200$ and $\tau$ varying.}\label{fig:exp2}%
\endminipage
\end{figure}

\textbf{Results and Discussion}: We run 40 replications of the experiment for each $\tau$ and $N$ and consider the MSE of each algorithm relative to $(\rho^{\epol}-\rho^{\bpol})^2$. To estimate the latter normalizer, we estimate each of $\rho^{\epol},\rho^{\bpol}$ as a simple sample average using $1000$ \emph{on-policy} trajectories.
This normalization enhances interpretability as we vary $\tau$.

In \cref{fig:exp1} we report the results for varying $N$ and fixing $\tau=1.3$. We show the relative MSEs on a logarithmic scale with $90\%$-confidence intervals. We observe that DRL clearly outperforms the other estimators. This can be attributed to the fact that both $w$- and $q$-estimators are flexible, and hence have high variance, which influences the variance of both MIS and DM, respectively, while DRL is largely insensitive to the particular $w$- and $q$-estimators.
One exception is $\tau=0.7,N=200$, where we see DM performs better than DR. On the other hand, MIS always performs worse than DR. This would suggest that $w$-estimation is more difficult than $q$-estimation in this environment, possibly because of the non-stationarity of the data.  Finally, we note DualDICE performs consistently badly across the settings, which can be attributed to the instability of the minimax optimization of two neural networks involved in its $w$-estimation.}

\section{Conclusions}\label{sec: conclusions}

We established the efficiency bound for OPE in a time-invariant Markov decision process in the regime where $N$ is (potentially) finite and $T\to \infty$. This novel lower bound quantifies how fast one could hope to estimate policy value in a model usually assumed in RL.
According to our results, many IS and DR OPE estimators used in RL are in fact not leveraging this structure to the fullest and are inefficient. This leads to MSE that is suboptimal in \emph{rate}, not just in leading coefficient.
We instead proposed the first efficient estimator achieving the efficiency bound, while also enjoying a double robustness property at the same time. 
We hope our work inspires others to further develop estimators that build on ours by leveraging MDP structure as we have here and perhaps combining this with ideas like balancing \citep{kallus2018balanced}, stability \citep{Kallus2019IntrinsicallyES}, or blending \citep{thomas2016} that can improve the finite-sample performance in addition to our asymptotic efficiency. Finally, we remark that although we focus on parametric estimation of nuisance functions, recent works address nonparametric estimation. For details, refer to \citet{huang2022beyond,uehara2021finite}.

\theendnotes

\bibliographystyle{informs2014}
\bibliography{pfi}

\newpage 
\begin{APPENDIX}{}

\section{Notation}

We first summarize the notation we use in \cref{tab:pre}. Following empirical process theory literature, in the proofs, we also use $\mathbb P$ to denote expectations interchangeably with $\rE[\cdot]$.

\begin{table}[h!]\label{tab:abbr}\SingleSpacedXI
    \centering
      \caption{Notation} \vspace{0.3cm}
    \begin{tabular}{l|l}
     $\cm_1,\cm_{1,b}$ & NMDP model, see \cref{def:nmdpmodel} \\
     $\cm_2,\cm_{2,b}$ & TMDP model, see \cref{def:tmdpmodel} \\
     $\cm_3,\cm_{3,b}$ & MDP model, see \cref{def:mdpmodel} \\
    $p_{\pi_e}^{(0)}(s)$ & Known initial distribution we want to evaluate\\ 
  $p_{\pi_b}^{(0)}(s)$ & Marginal distribution of offline data over the state space\\ 
    $\cj$ & $(s_0,a_0,r_0,s_1,a_1,r_1,\dots)$ \\
     $\cj_{s_t}$, $\cj_{a_t}$ & History up to $s_t$ or to $a_t$, respectively \\
     $\ch_{s_t}$, $\ch_{a_t}$ & History up to $s_t$ or to $a_t$, respectively, excluding reward variables \\
    $r_t,s_t,a_t,$  &  Reward, state, and action at $t$  \\
    $\eta_t$ & $\pi_{e,t}/\pi_{b,t}$ \\ 
    $\rho^\pi$  & Policy value, $\lim_{T\to \infty}c_T(\gamma)\;\rE_{\pi}[\sum_{t=0}^{T}\gamma^{t}r_t]$\\
    $\rho^\pi_{T}$  & $c_T(\gamma)\;\rE_{\pi}[\sum_{t=0}^{T}\gamma^{t}r_t]$\\
    $p^{(\infty)}_{\pi,\gamma}(s)$ & Average visitation distribution of the policy $\pi$ with discount rate $\gamma$ \\
    $p^{(\infty)}_{\pi}(s)$ & Stationary distribution \\
    $w(s)$ & $p^{(\infty)}_{\epol,\gamma}(s)/p^{(0)}_{b}(s)$ \\
    $\nabla_{\beta}$ & Differentiation with respect to $\beta$ \\
    $\epol(a|s)$, $\bpol(a|s)$ &  Target and behavior policies respectively \\
$v(s)$ & Value function   \\
 $q(s,a)$ & $q$-function  \\ 
    $\nu_{t}(\ch_{a_t}) $  &  Cumulative density ratio $\prod_{k=0}^{t}\pi_{e,k}/\pi_{b,k}$ \\ 
    $\mu_{t}(s_t,a_t)$  &  Marginal density ratio $\rE[\nu_t\mid s_t,a_t]$ \\
    $\eta(s,a)$ & Instantaneous density ratio $\epol(a|s)/\bpol(a|s)$ \\ 
    $C, R_{\mathrm{max}}$ & Upper bound of density ratio and reward, respectively \\
    $\|\cdot \|_{p}$ & $L^{p}$-norm $\rE[f^{p}]^{1/p}$\\
    $\rE_{\pi}[\cdot],\P_\pi$ & Expectation with respect to a sample from a policy $\pi$ \\
    $\rE[\cdot],\P$ & Same as above for $\pi=\bpol$   \\
    $\rE_{N}[\cdot],\P_{N},\P_{T}$ & Empirical or time average (based on sample from a behavior policy) \\
    $\mathcal D_0,\mathcal D_1$& The split samples when using cross-fitting, $\mathcal D_0\cup\mathcal D_1=\{1,\dots,N\}$\\
    $N_j$& The size of $\mathcal D_j$\\
    $\rE_{N_j},\P_{N_j}$& Empirical expectation on $\mathcal D_j$\\
    $\bG_N$ & Empirical process $\sqrt{N}(\P_{N}-\P)$  \\
    $A_N=\smallo_{p}(a_N)$ & The term $A_N/a_N$ converges to zero in probability \\ 
    $A_N=\bigO_{p}(a_N)$  & The term $A_N/a_N$ is bounded in probability \\
    \end{tabular}
    \label{tab:pre}
\end{table}
\newpage

{\blockedit 
\section{Technical Background}

\subsection{Some Definitions}\label{sec:appendixdefs}
\begin{definition}[$\P$-Donsker]\label{def:donsker}
A class $\Fcal$ of measurable functions $f$ is called $\P$-Donsker when the sequence of processes $\sqrt{n}(\P_n-\P)$ converges in distribution to a tight limit process in the space $l^{\infty}(\Fcal)$ as $n\to\infty$. 
\end{definition}
If the class $\Fcal$ has a square-integrable envelope and a finite uniform entropy integral, then $\Fcal$ is $\P$-Donsker \citep[Theoreem 19.14]{VaartA.W.vander1998As}. 

Next we review the definitions of the standard mixing coefficients \citep[Chapter 14]{DavidsonJamesE.H1994Slta}.

\begin{definition}[Mixing coefficients]
Consider a stationary stochastic process $X_1,X_2,\dots$, \ie, identically distributed but possibly dependent. Let $P$ denote the joint measure of the process, define $\mathcal{Q}_{a}^b=\sigma(X_a,\cdots,X_b)$ as the sigma algebra corresponding to subsequences of variable, and let $L^2(Q_a^b)$ be the corresponding space of square-integrable functions. Then we have the following definitions:
\begin{align*}
    \alpha_m&=\sup_t\sup_{G\in \mathcal{Q}_{1}^t , H\in  \mathcal{Q}_{t+m}^{\infty}}  |P(G\cap H) -P(G)P(H)|,\\
    \beta_m&=\sup_t\sup_{H\in  \mathcal{Q}_{t+m}^{\infty}}  |P(H|\mathcal{Q}_{1}^t) -P(H)|,\\
    \phi_m&=\sup_t\sup_{G\in \mathcal{Q}_{1}^t , H\in  \mathcal{Q}_{t+m}^{\infty},P(G)>0}  |P(H|G) -P(H)|,\\
    \rho_m&=\sup_t\sup_{g\in L^2(\mathcal{Q}_{1}^t) , h\in  L^2(\mathcal{Q}_{t+m}^{\infty})} 
    \abs{\operatorname{Corr}(g,h)}
    .
\end{align*}
\end{definition}

Each of these coefficients measures the amount of dependence when we consider subsequences that are separated by $m$ jumps. The more ergodic the process, the faster these shrink toward zero as $m$ grows. An independent process has all of these always equal zero.

These coefficients satisfy the following relationships \citep{bradley2005basic}:
\begin{align}\label{eq:mixing relationships}
2\alpha_m\leq\beta_m\leq \phi_m,\quad
4\alpha_m\leq\rho_m\leq 2\phi_m^{1/2}.
\end{align}

\subsection{Semiparametric Theory}\label{appendix:semiparam}

In this section, we expand on \cref{sec:semiparam} to briefly review the precise definitions and results of semiparametric theory that we use following \citep{VaartA.W.vander1998As,bickel98}. We focus on scalar estimands for simplicity since our OPE problem involves a scalar estimand. We denote the all of the data $\{O_{i}\}_{i=1}^{n}$ as $O^{n}$, the estimand as $R:\cm \to \mathbb{R}$, and the estimator as $\hat R(O^{n})$. We let $\mu$ be a dominating measure for $\cm$ such that $\mu\gg F$ for $F\in\cm$.

\subsubsection{Definitions}

\begin{definition}[Influence function of estimators]
An estimator $\hat R(O^{n})$ is asymptotically linear (AL) with influence function (IF) $\psi(O)$ if 
\begin{align*}
    \sqrt{n}(\hat R(O^{n})-R(F))=\frac{1}{\sqrt{n}}\sum_{i=1}^{n}\psi(O_{i})+\op(1/\sqrt{n}). 
\end{align*}
\end{definition}

{\newedit 
\begin{definition}[One-dimensional submodel and its score function]
A one-dimensional submodel of $\cm$ passing through $F$ is a set of distributions $\{F_{\epsilon}:\epsilon \in(-1,1)\}\subset\cm$ such that: (a) $F_{0}=F$, (b) the score function
\begin{align*}
    s(O;\epsilon)=\frac{\mathrm{d}}{\mathrm{d}\epsilon}\log (\mathrm{d}F_{\epsilon}/\mathrm{d}\mu)(O)
\end{align*}
exists, (c) $\E s^2(O;0)<\infty$, and (d) $\E\sup_{\epsilon\in(-1,1)}\abs{\mathrm{d}F_{\epsilon}/\mathrm{d}\mu(O)}<\infty$.

The score of the submodel at $F$ is defined as $s(O)=s(O;0)$. 
By definition it belongs to $L_2$, the Hilbert space of square-integrable functions wrt $F$.
\end{definition}

\begin{definition}[Tangent space]
The tangent space at $F$ wrt $\cm$ is the linear closure of the score functions at $F$ over all one-dimensional submodels wrt $L_2$.
\end{definition}
Note that the tangent space is always a cone.

\begin{definition}[Pathwise differentiability]\label{def:pathwise}
A functional $R(F)$ is pathwise differentiable at $F$ wrt $\cm$ if there exists a function $D_{F}(O)$ such that any one-dimensional submodel $\{F_{\epsilon}\}$ of $\cm$ passing through $F$ with score function $s(O)$ satisfies
\begin{align*}
    \frac{d R(F_{\epsilon})}{d\epsilon}\mid_{\epsilon=0}=\E[D_{F}(O)s(O)].
\end{align*}
The function $D_{F}(O)$ is called a gradient of $R(F)$ at $F$ wrt $\cm$. The efficient IF (EIF, or canonical gradient) of $R(F)$ wrt  $\cm$ is the unique gradient $\tilde D_F(O)$ of $R(F)$ at $F$ wrt $\cm$ that belongs to the tangent space at $F$ wrt $\cm$.

\end{definition}

Next, we define regular estimators, which are those whose limiting distribution is insensitive to local changes to the data-generating process. It excludes, for example, super-efficient pathologies such as the Hodge estimator. 
}  

\begin{definition}[Regular estimators]\label{def:regular}
An estimator sequence $\hat R$ is called regular at $F$ for $R(F)$ wrt $\cm$, if there exists a probability measure $L$ such that, for any one-dimensional submodel $\{F_{\epsilon}\}$ of $\cm$ passing through $F$, we have
\begin{align*}
    \text{$\sqrt{n}\{\hat R(O^n)-R(F_{1/\sqrt{n}})\}$ with $O^n\sim F_{1/\sqrt{n}}^n$, converges in distribution to $L$ as $n\to\infty$.}
\end{align*}
\end{definition}

\subsubsection{Characterizations}
The following theorems show that influence functions of asymptotically linear estimators for $R(F)$ and gradients of $R(F)$ correspond to one another and how to compute the EIF. These are both based on Theorem 3.1 of \citet{vaart1991}.

\begin{theorem}[Influence functions are gradients]
\label{thm:gradients}
Suppose $\hat R(O^n)$ is an asymptotically linear estimator of $R(F)$ with influence function $D_F(O)$ and that $R(F)$ is pathwise differentiable at $F$ wrt $\cm$. Then, $\hat R(O^n)$ is a regular estimator of $R(F)$ wrt $\cm$ if and only if and $D_F(O)$ is a gradient of $R(F)$ at $F$ wrt $\cm$. 
\end{theorem}

\begin{corollary}[Characterization of EIF]\label{cor:EIF}
The EIF wrt $\cm$ is the projection of any gradient wrt $\cm$ onto the tangent space wrt $\cm$. 
\end{corollary}

\subsubsection{Strategy to calculate the EIF}

{\newedit
With the above definitions and theorems in mind, our general strategy to construct EIF is as follows. 
\begin{enumerate}
    \item Calculate a gradient $D_F(O)$ of the target functional $R(F)$.
    \item Calculate the tangent space wrt $\cm$.
    \item Either:
    \begin{enumerate}
        \item Show that $D_F(O)$ already lies in the tangent space and is thus the EIF, or
        \item Project $D_F(O)$ onto the tangent space to obtain the EIF.
    \end{enumerate}
\end{enumerate}
}
\subsubsection{Optimalites}

The efficiency bound is defined as the variance of the EIF, $\var_{F}[\tilde D_F(O)]$.
It has the following interpretations. First, the efficiency bound is the lower bound in a local asymptotic minimax sense \citep[Thm.~25.20]{VaartA.W.vander1998As}.
\begin{theorem}[Local Asymptotic Minimax theorem]
\label{thm:lam}
Let $R(F)$ be pathwise diffentiable at $F$ wrt $\cm$ with the EIF $\tilde D_F(O)$. Then, for any estimator sequence $\hat R(O^n)$, and symmetric quasi-convex loss function $l:\mathbb{R}\to [0,\infty)$,
\begin{align*}
    \sup_{m\in\mathbb N,\,\{F_{\epsilon}^{(1)}\},\dots,\,\{F_{\epsilon}^{(m)}\}} \lim_{n\to \infty }\sup_{k=1,\dots,m}\rE_{F^{(k)}_{1/\sqrt{n}}}[l(\sqrt{n}\{\hat R(O^n)-R(F^{(k)}_{1/\sqrt{n}})\})]\geq \int l(u)\mathrm{d}\mathcal N(0,\var_{F}[\tilde D_F(O)])(u), 
\end{align*}
where the first supremum is taken over all finite collections of one-dimensional submodels of $\cm$ passing through $F$.
\end{theorem}
The above suprema appear somewhat complicated. It also implies the following weaker but easier to interpret result.
\begin{corollary}
Under the same assumptions of \cref{thm:lam}, 
\begin{align*}
    \inf_{\delta>0}\liminf_{n\to\infty}\sup_{\|Q-F\|_\text{TV}\leq \delta}\rE_{Q}[l(\sqrt{n}\{\hat R(O^n)-R(Q)\})]\geq \int l(u)\mathrm{d}\mathcal{N}(0,\var_{F}[\tilde D_F(O)])(u),  
\end{align*}
where $\|\cdot\|_\text{TV}$ is the total variation distance. 
\end{corollary}

Second, the efficiency bound can be seen as a pointwise lower bound bound at the specific instance $F$, if we restrict to regular estimators
\citep[Thm.~25.21]{VaartA.W.vander1998As}. 

\begin{theorem}[Convolution theorem] \label{thm:conv}
Let $l:\mathbb{R}\to [0,\infty)$ be a symmetric quasi-convex loss function.
Let $R(F)$ be pathwise differentiable at $F$ wrt $\cm$ with EIF $\tilde D_F(O)$. Let $\hat R(O^n)$ be a regular estimator sequence at $F$ wrt $\cm$ with limiting distribution $L$. Then, 
\begin{align*}
    \int l(u)\mathrm{d}L(u)\geq \int l(u)\mathrm{d}\mathcal{N}(0,\var_{F}[\tilde D_F(O)])(u).
\end{align*}
\end{theorem}

Equality holds obviously holds when $L=\mathcal{N}(0,\var_{F}[\tilde D_F(O)])$. We therefore say an estimator is efficient if it is regular and its limiting distribution is $\mathcal{N}(0,\var_{F}[\tilde D_F(O)])$. Note that such would be implied if it were regular and AL with influence function $\tilde D_F(O)$. The following implies such is in fact both necessary and sufficient \citep[Lemma~25.23]{VaartA.W.vander1998As}.
\begin{theorem}
Let $R(F)$ be pathwise differentiable at $F$ wrt $\cm$ with the EIF $\tilde D_F(O)$. Then an estimator sequence is efficient (regular wrt $\cm$ and with limiting distribution $\mathcal{N}(0,\var_{F}[\tilde D_F(O)])$) if and only if is AL with influence function $\tilde D_F(O)$.
\end{theorem}
}

\section{Proofs}
\begin{myproof}[Proof of \cref{thm:m1_bound}]
{\newedit 
We omit the proof since as it can (almost) be concluded as corollary of Theorem 1 of \citet{NathanUehara2019} where we simply set $T=\infty$ and redefine rewards by multiplying by $\gamma^t$. There are only two subtle differences. They prove the efficiency bound when $p(r_t|\cj_{a_t}),p(a_t|\cj_{s_t}),p(s_t|\cj_{r_t})$ are replaced with $p(r_t|\ch_{a_t}),p(a_t|\ch_{s_t}),p(s_t|\ch_{a_t})$, and the initial density is unknown. The former change simply means our conditioning sets are in the efficiency bound are slightly different. The latter change simply eliminates the $k=0$ term in the efficiency bound.
}
\end{myproof}

\begin{myproof}[Proof of \cref{thm:m2_bound}]
{\newedit 
Again we omit the proof as it can (almost) be concluded as a corollary of Theorem 3 of \citep{NathanUehara2019}. There is a single subtle difference. They prove the efficiency bound when the initial density is unknown. This change simply eliminates the $k=0$ term in the efficiency bound.
}

\end{myproof}

\begin{myproof}[Proof of \cref{cor:m1m2ebexist}]
We prove the statement for NMDP. For TMDP, the statement is confirmed similarly.
We have 
\begin{align*}
     &\lim_{T \to \infty }(1-\gamma)^{2}\sum_{k=1}^{T}\mathrm{E}[\gamma^{2(k-1)}\nu^{2}_{k-1} \mathrm{var}\{r_{k-1}+v(s_{k})|s_{k-1},a_{k-1}\}] \\
     &  \lnapprox \lim_{T \to \infty }(1-\gamma)^{2}\sum_{k=1}^{T}\gamma^{2(k-1)} \|\nu_k\|^{2}_{\infty}<\infty. 
\end{align*}
The final statement is clear because TMDP is included in NMDP. 
\end{myproof}

{\newedit 
\begin{myproof}[Proof of \cref{thm:m3_bound}]
Let $o=(s,a,r,s')$ and let $L_2=\{f(o):\E f^2(o)<\infty\}$.
Recall the model is 
\begin{align*}
    \cm_{3}=\{p(o)=p(s)p(a|s)p(r|s,a)p(s'|s,a) \;:\; p(s)p(a\mid s)>0\}. 
\end{align*}

\paragraph{\bf{Calculation of the tangent space}}  
We first prove that the corresponding tangent space is 
\begin{align*}
   \Tcal&=L_2\cap(\Tcal_1 +  \Tcal_2 +  \Tcal_3 + \Tcal_4), ~~\text{where}\\
   \Tcal_1 &=\braces{f(s): \E[f]=0 }\,,\Tcal_2=\braces{f(s,a): \E[f(s,a)|s]=0},\\
   \Tcal_3 &=\braces{f(s,a,s'): \E[f(s,a,s')|s,a]=0},\\
   \Tcal_4 &=\braces{f(s,a,r): \E[f(s,a,r)|s,a]=0}. 
\end{align*}
Notice that $\Tcal_j$ are orthogonal so the above sum is a direct sum.

For any submodel
\begin{align*}
    \{p_{\theta}(s)p_{\theta}(a\mid s)p_{\theta}(s'|s,a)p_{\theta}(r|s,a) \},
\end{align*}
passing through $p$ at $\theta=0$,
the score function can be decomposed as  
\begin{align*}
    g(o)
     &=g_{s}(s)+g_{a|s}(s,a)+g_{r|s,a}(s,a,r)+g_{s'|s,a}(s,a,s'),~\text{where}\\
    g_s(s) &\coloneqq \frac{d}{d\theta}\log p_{\theta}(s),\,
    g_{a|s}(s,a)\coloneqq \frac{d}{d\theta}\log p_{\theta}(a|s),\\
    g_{r|s,a}(a,s,r)&\coloneqq \frac{d}{d\theta}\log p_{\theta}(r|s,a),\,
    g_{s'|s,a}(s,a,a')\coloneqq \frac{d}{d\theta}\log p_{\theta}(s'|s,a),
\end{align*}
where these satisfy
$$
\E[g_s(s)]=0,\,
\E[g_{a|s}(s,a)\mid s]=0,\,
\E[g_{r|s,a}(s,a,r)\mid s,a]=0,\,
\E[g_{s'|s,a}(s,a,s')\mid s,a]=0.
$$
Therefore, $\Tcal$ contains the tangent space.

Next, consider any $g(o)\in\Tcal$. Since $g$ is in the direct sum $\Tcal_j$ we can write $g(o)=g_{s}(s)+g_{a|s}(s,a)+g_{r|s,a}(s,a,r)+g_{s'|s,a}(s,a,s')$. 
Let $k(x)=2(1+e^{-2x})^{-1}$. Notice that $k$ is bounded by $2$ and $k(0)=k'(0)=1$.
Consider the submodel 
\begin{align*}
    p_{\theta}(s)p_{\theta}(a|s)p_{\theta}(r|s,a)p_{\theta}(s'|s,a),\quad \theta \in (-\epsilon,\epsilon)
\end{align*}
where 
\begin{align*}
    & p_{\theta}(s)=p(s)k(\theta g_s(s)),\, p_{\theta}(a|s)=p(a|s)k(\theta  g_{a|s}(s,a)), \\ &p_{\theta}(s'|s,a)=p(s'|s,a)k(\theta g_{s'|s,a}(s,a,s')),\, p_{\theta}(r|s,a)=p(r|s,a)k(\theta g_{r|s,a}(s,a,r)). 
\end{align*}
Taking the score of this submodel we see it exactly coincides with $g$, completing the argument.

\paragraph{\bf{Calculation of gradient}}
We show the statement in two steps: (1) calculating some gradient, (2) showing this gradient is in the tangent space of $\cm_3$. Then, we can conclude that this gradient is actually the EIF. 

To calculate a gradient of the target functional $J(\epol)$ wrt the model ($\cm_3$) %
we seek $\phi$ satisfying 
\begin{align*}
   \nabla_{\theta}\rho^{\epol}(\theta)|_{\theta=0}&=\rE_{p_{b}}[\phi(s,a,r,s')g(s,a,r,s')]|_{\theta=0},\\
\rho^{\epol}(\theta) & = \lim_{T\to \infty}\rE[c_{T}(\gamma)\sum_{t=0}^{T}\gamma^{t}r_t|s_0 \sim p_{e}(s_0),a_0 \sim \epol(a_1|s_1),s_1\sim p_{\theta}(s_1|s_0,a_0),r_1\sim p_{\theta}(r_1|s_1,a_1),\cdots ], 
\end{align*}
where $p_{\theta}(s)p_{\theta}(a\mid s)p_{\theta}(s'|s,a)p_{\theta}(r|s,a)$ is any submodel, whose score function belongs to the tangent set $\Tcal$. 
Note
\begin{align*}
 \rho^{\epol}(\theta)%
  &=\lim_{T\to \infty} c_{T}(\gamma)\sum_{t=0}^T \int \gamma^{t} r_tp_{\theta}(r_t|s_t,a_t)\left\{\prod_{k=0}^t \epol(a_k|s_k)p_{\theta}(s_{k+1}|s_k,a_k)\right\}p^{(0)}_{\epol}(s_0)\mathrm{d}\lambda(\cj_{s_{t+1}})\\
  &=\int \int  c_{t}(\gamma)\gamma^{t} r_t p_{\theta}(r_t|s_t,a_t)\left\{\prod_{k=0}^t \epol(a_k|s_k)p_{\theta}(s_{k+1}|s_k,a_k)\right\}p^{(0)}_{\epol}(s_0)\mathrm{d}\lambda(\cj_{s_{t+1}})\mathrm{d}\mu'(t)  
\end{align*}
where $\mu'$ is a counting measure on $\mathbb{Z}$. This is only to emphasize that $\lim_{T\to\infty}\sum_{t=0}^{T}$ is an integral wrt the counting measure. 

Then, we have 
\begin{align} \label{eq:exchange}
      \nabla_{\theta} \rho^{\epol}(\theta)|_{\theta=0}&=C(\theta)|_{\theta=0},\\
      C(\theta)&=\lim_{T\to \infty}c_{T}(\gamma)\sum_{t=0}^{T}\int \gamma^{t} r_t \nabla p_{\theta}(r_t|s_t,a_t)\braces{\prod_{k=0}^t \epol(a_k|s_k)p_{\theta}(s_{k+1}|s_k,a_k)}p^{(0)}_{\epol}(s_0)\mathrm{d}\lambda(\cj_{s_{t+1}})  \nonumber \\
        &+\lim_{T\to \infty}c_{T}(\gamma)\sum_{t=0}^{T}\int \gamma^{t} r_tp_{\theta}(r_t|s_t,a_t) \braces{\prod_{k=0}^t \epol(a_k|s_k)\nabla p_{\theta}(s_{k+1}|s_k,a_k)}p^{(0)}_{\epol}(s_0)\mathrm{d}\lambda(\cj_{s_{t+1}}).   \nonumber
\end{align}

This exchange of integration and differentiation is justified by showing
\begin{align}\label{eq:check1}
     &\lim_{T\to \infty}c_{T}(\gamma)\sum_{t=0}^{T}\int \gamma^{t} r_t |\nabla p_{\theta}(r_t|s_t,a_t)|\braces{\prod_{k=0}^t \epol(a_k|s_k)p_{\theta}(s_{k+1}|s_k,a_k)}p^{(0)}_{\epol}(s_0)\mathrm{d}\lambda(\cj_{s_{t+1}})\\
        &+\lim_{T\to \infty}c_{T}(\gamma)\sum_{t=0}^{T}\int \gamma^{t} r_tp_{\theta}(r_t|s_t,a_t) \sum_{k=0}^t \braces{\prod_{k=0}^t \epol(a_k|s_k)|\nabla p_{\theta}(s_{k+1}|s_k,a_k)|}p^{(0)}_{\epol}(s_0)\mathrm{d}\lambda(\cj_{s_{t+1}}). \label{eq:check2}
\end{align}
is uniformly upper-bounded by some value around some neighborhood of $\theta=0$. The first term \eqref{eq:check1} is equal to 
\begin{align*}
    &\lim_{T\to \infty}\int c_{T}(\gamma)\sum_{t=0}^{T} \rE_{\epol,\theta}[\gamma^{t} r_t |g_{R|S,A}(r_t|s_t,a_t)|]=\rE_{p^{(\infty)}_{e,\gamma}(s:\theta)\epol(a|s)p_{\theta}(r|s,a) }[r |g_{R|S,A}(r|s,a:\theta)|] \tag{Fubini}\\
    &\leq R_{\max}\rE_{p^{(\infty)}_{e,\gamma}(s:\theta)\epol(a|s) }[\rE_{p_{\theta}(r|s,a)}[|g_{R|S,A}(r|s,a:\theta)|\mid s,a]]\\
        &\leq R_{\max}\rE_{p^{(\infty)}_{e,\gamma}(s:\theta)\epol(a|s) }[\int h(r|s,a)\mathrm{d}\mu(r)]<\infty, 
\end{align*}
where $h(r|s,a)$ is some integrable function. Here we leveraged condition (d) in the definition of a one-dimensional submodel. 

The second term \eqref{eq:check2} is equal to 
\begin{align*}
     &\lim_{T\to \infty}\gamma \int c_{T}(\gamma)\sum_{c=1}^{T}\gamma^c \sum_{t=c+1}^{T}\rE_{\epol,\theta}[\gamma^{t-c-2}r_t |g_{S'|S,A}(s_{t+1}|s_t,a_t:\theta)| ]
\end{align*}
by changing an index and recalling $$|\nabla p_{\theta}(s_{t+1}\mid s_t,a_t)|= |g_{S'\mid S,A}(s_{t+1} \mid s_t,a_t)| p_{\theta}(s_{t+1}\mid s_t,a_t).$$ Then, it can be upper-bounded as follows: 
\begin{align*}  
     & \lim_{T\to \infty}\gamma \int c_{T}(\gamma)\sum_{c=1}^{T}\gamma^c \rE_{\epol,\theta}[v(s_{t+1}) |g_{S'|S,A}(s_{t+1}|s_t,a_t:\theta)| ] \tag{Tower property} \\ 
     & = \gamma \rE_{p^{(\infty)}_{e,\gamma}(s:\theta)\epol(a|s)p_{\theta}(s'|s,a) }[v(s')|g_{S'|S,A}(s'|s,a:\theta)|] \tag{Definition of discounted occupancy measure} \\ 
     &\leq (1-\gamma)^{-1}R_{\max}\gamma \rE_{p^{(\infty)}_{e,\gamma}(s:\theta)\epol(a|s)}[\rE_{p_{\theta}(s'|s,a)}[|g_{S'|S,A}(s'|s,a:\theta)|]] \\
      &\leq (1-\gamma)^{-1}R_{\max}\gamma \rE_{p^{(\infty)}_{e,\gamma}(s:\theta)\epol(a|s)}[\int h(s'|s,a)\mathrm{d}(s')]<\infty, 
\end{align*}
where $h(s'|s,a)$ is some integrable function. 
Combing all together, the exchange of integration and differentiation \eqref{eq:exchange} is justified. 

Now, we calculate $\nabla\rho^{\epol}(\theta)|_{\theta=0}$ in order to derive the influence function, we continue the calculation: 
\begin{align*}
     &\nabla\rho^{\epol}(\theta)|_{\theta=0}\\
     &= \rE_{p^{(\infty)}_{e,\gamma}}[r g_{R|S,A}(r|s,a)]+\gamma \rE_{p^{(\infty)}_{e,\gamma}}[v(s')g_{S'|S,A}(s'|s,a)] \tag{We already checked this to prove the exchange of integration and differentiation}\\ 
     &=\rE_{p^{(\infty)}_{e,\gamma}}[\{r-\rE[r|s,a]\}g_{R|S,A}(r|s,a)]+\gamma \rE_{p^{(\infty)}_{e,\gamma}}[\{v(s')-\rE[v(s')|s,a]\}g_{S'|S,A}(s'|s,a)] \tag{Means of score functions are $0$}\\
     &= \rE_{p^{(\infty)}_{e,\gamma}}\left[\left\{r-\rE[r|s,a]+\gamma v(s')-\gamma \rE[v(s')|s,a]\right\} g(s,a,r,s')\right] \\ 
     &=\rE_{p^{(\infty)}_{e,\gamma}}[\{r+\gamma v(s')-q(s,a)\} g(s,a,r,s') ] \tag{Use $\rE[r+\gamma v(s')\mid s,a]=q(s,a)$ }\\
    &=\rE_{p_{b}(s,a,r,s')}[(w(s)\eta(s,a)\{r+\gamma v(s')-q(s,a)\})g(s,a,r,s') ]. \tag{Importance sampling}
\end{align*}
Therefore, the following function is a gradient, if the $L_2$-norm is finite:
\begin{align*}
   w(s)\eta(s,a)\{r+\gamma v(s')-q(s,a)\}
\end{align*}
In addition, if it exists, it belongs to $\Tcal_1+ \Tcal_2+ \Tcal_3+ \Tcal_4$ since 
\begin{align*}
   &  w(s)\eta(s,a)\{r-\rE[r|s,a]\}\in \Tcal_4,\, w(s)\eta(s,a)\{v(s')-\rE[v(s')|s,a]\}\in \Tcal_3,\\
    &  w(s)\eta(s,a)\{r+\gamma v(s')-q(s,a)\}= w(s)\eta(s,a)\{ r-\rE[r|s,a\}+\gamma\{v(s')-\rE[v(s')|s,a]\}\}. 
\end{align*}
Thus, it is the EIF. The efficiency bound is therefore
\begin{align*}
   \rE_{p_{b}}[ w^2(s)\eta^2(a,s)\{r+\gamma v(s')-q(s,a)\}^2]. 
\end{align*}

Finally, we show that the EIF is still the same under $\cm_{3b}$. First, the above EIF under $\cm_3$ is still a gradient under $\cm_{3b}$.  Again, what we have to prove is that this function belongs to the tangent space. This is obvious since the tangent space of $\cm_{3b}$ is 
\begin{align*}
    L_2\cap(\Tcal_1 + \Tcal_3 +   \Tcal_4),
\end{align*}
and the gradient above belongs to  $L_2\cap(\Tcal_3 + \Tcal_4)\subseteq L_2\cap(\Tcal_1 + \Tcal_3 +   \Tcal_4)$. 
\end{myproof}
}

{\newedit 
\begin{myproof}[Proof of \cref{thm:m1_inf}]

In this proof, we write $q^{\omega_N}_k$ as $q_k$ to simplify the notation. Define:
\begin{align*}
\psi(\{\hat{\nu}_{k}\},\{\hat{q}_k\})=\{1-\gamma\}^{-1}\bracks{\rE_{a_0\sim \epol(s_0), s_0\sim p^{(0)}_e(s_0)}[\hat{q}_0(s_0,a_0)]
    + \sum_{k=0}^{\omega_N}\hat{\nu}_{k}
    \{r_k- \hat{q}_k(s_k,a_k) + \gamma \rE_{\epol}[\hat{q}_{k+1}(s_{k+1},a_{k+1})|s_{k+1}]\}}. 
\end{align*}
The estimator $\hat{\rho}^{\cm1}_{\mathrm{DRL}(\cm_1)}$ is then given by
\begin{align*}
\frac{N_0}{N}\P_{\dm_0}\psi(\{\hat{\nu}^{[1]}_{k}\},\{\hat{q}^{[1]}_k\})+\frac{N_1}{N}\P_{\dm_1}\psi(\{\hat{\nu}^{[0]}_{k}\},\{\hat{q}^{[0]}_k\}),
\end{align*}
where $\P_{\dm_j}$ is the empirical average over $\dm_j$ and $N_j$ is the sample size of each.
Then, we have 
\begin{align}
    \sqrt{N}(\P_{\dm_0}\psi(\{\hat{\nu}^{[1]}_{k}\},\{\hat{q}^{[1]}_k\})-\rho^{\epol}_{\omega_N})&= \sqrt{N/N_0}\bG_{N_0}[\psi(\{\hat{\nu}^{[1]}_{k}\},\{\hat{q}^{[1]}_k\})-\psi(\{\nu_{k}\},\{q_k\})]
    \label{eq:term1_drl}
    \\
    &+\sqrt{N/N_0}\bG_{N_0}[\psi(\{\nu_{k}\},\{q_k\}) ] 
       \label{eq:term1_dr2}
    \\
    &+\sqrt{N}(\rE[\psi(\{\hat{\nu}^{[1]}_{k}\},\{\hat{q}^{[1]}_k\})|\{\hat{\nu}^{[1]}_{k}\},\{\hat{q}^{(1)}_k\} ]-\rho^{\epol}_{\omega_N} ).    \label{eq:term1_dr3}
\end{align}
We analyze each term. To do that, we use the following relation:
\begin{align*}
&\psi(\{\hat{\nu}_{k}\},\{\hat{q}_k\})-\psi(\{\nu_{k}\},\{{q}_k\})=D_1+D_2+D_3,\quad\text{where}\\
&D_1 =\sum_{k=0}^{w_N}(\gamma^k \hat{\nu}_{k}-\gamma^k \nu_{k})(-\hat{q}_k+q_k)+\gamma^k(\hat{\nu}_{k-1}-\nu_{k-1})(-\hat{v}_k+v_k), \\
&D_2 =\sum_{k=0}^{\omega_N}\gamma^k \nu_{k}(\hat{q}_k-q_k)+\gamma^{k} \nu_{k-1}(\hat{v}_k-v_k), \\
&D_3 =\sum_{k=0}^{\omega_N}\gamma^k(\hat{\nu}_{k}-\nu_{k} )(r_k-q_k+v_{k+1}).
\end{align*}

First, we show the term \eqref{eq:term1_drl} is $\smallo_{p}(1)$. 

\paragraph*{\bf{The term \eqref{eq:term1_drl} is $\smallo_{p}(1)$.}}

If we can show that for any $\epsilon>0$, 
\begin{align}
\label{eq:process_m1}
\lim_{n\to \infty}\sqrt{N_0}P[&\P_{\dm_0}[\psi(\{\hat{\nu}^{[1]}_{k}\},\{\hat{q}^{[1]}_k\})-\psi(\{\nu_{k}\},\{{q}_k\})]\\&\notag- 
 \rE[\psi(\{\hat{\nu}^{[1]}_{k}\},\{\hat{q}^{[1]}_k\})-\psi(\{\nu_{k}\},\{{q}_k\})|\{\hat{\nu}^{[1]}_{k}\},\{\hat{q}^{[1]}_k\} ]>\epsilon|\mathcal{D}_1]=0, %
\end{align}
Then, by bounded convergence theorem, we would have 
\begin{align*}
  \lim_{n\to \infty}\sqrt{N_0}P[&\P_{\dm_0}[\psi(\{\hat{\nu}^{[1]}_{k}\},\{\hat{q}^{[1]}_k\})-
  \psi(\{\nu_{k}\},\{{q}_k\})]\\&-\rE[\psi(\{\hat{\nu}^{[1]}_{k}\},\{\hat{q}^{[1]}_k\})-\psi(\{\nu_{k}\},\{{q}_k\})|\{\hat{\nu}^{[1]}_{k}\},\{\hat{q}^{[1]}_k\}]>\epsilon]=0,
\end{align*}
yielding the statement. 

To show \eqref{eq:process_m1}, we show that the conditional mean is $0$ and conditional variance is $\smallo_p(1)$. 
The conditional mean is 
\begin{align*}
&\rE[\P_{\dm_0}[\psi(\{\hat{\nu}^{[1]}_{k}\},\{\hat{q}^{[1]}_k\})-\psi(\{\nu_{k}\},\{{q}_k\})]\\&\qquad\qquad\qquad\qquad\qquad-\rE[\psi(\{\hat{\nu}^{[1]}_{k}\},\{\hat{q}^{[1]}_k\})-\psi(\{\nu_{k}\},\{{q}_k\})\mid \{\hat{\nu}^{[1]}_{k}\},\{\hat{q}^{[1]}_k\}]|\mathcal{D}_1] \\
&=0.
\end{align*}
Here, we leverage the sample splitting construction, that is, $\hat{\nu}^{[1]}_{k}$ and $\hat{q}^{[1]}_{k}$ only depend on $\mathcal{D}_1$. The conditional variance is 
\begin{align*}
&\mathrm{var}[\sqrt{N_0}\P_{\dm_0}[\psi(\{\hat{\nu}^{[1]}_{k}\},\{\hat{q}^{[1]}_k\})-\psi(\{\nu_{k}\},\{{q}_k\})]|\mathcal{D}_1]\\
&\qquad\qquad= \rE[D^{2}_1+D^{2}_2+D^{2}_3+2D_{1}D_{2}+2D_{2}D_{3}+2D_{2}D_{3} \mid \mathcal{D}_1] \\ 
&\qquad\qquad=\omega^{2}_N \max\{\Op((\kappa^\nu_N)^2),\Op((\kappa^q_N)^2),\Op(\kappa^\nu_N\kappa^q_N)\}=\smallo_p(1).
\end{align*}
Here, we used the convergence rate assumptions (\ref{thm:m1_inf}c) and the relation $\|\hat{v}^{[1]}_{k}-v_{k}\|_{2}\leq C\|\hat{q}^{[1]}_{k}-q_{k}\|_{2}$ arising from the fact that the former is the marginalization of the latter over $\pi_{e,k}$ and Jensen's inequality. Then, from Chebyshev's inequality:
\begin{align*}
&\sqrt{N_0}P[\P_{\dm_0}[\psi(\{\hat{\nu}^{[1]}_{k}\},\{\hat{q}^{[1]}_k\})-\psi(\{\nu_{k}\},\{{q}_k\})]-\rE[\psi(\{\hat{\nu}^{[1]}_{k}\},\{\hat{q}^{[1]}_k\})-\psi(\{\nu_{k}\},\{{q}_k\})|\{\hat{\nu}^{[1]}_{k}\},\{\hat{q}^{[1]}_k\} ]>\epsilon|\mathcal{D}_1] \\
&\leq \frac{1}{\epsilon^{2}}\mathrm{var}[\sqrt{N_0}\P_{\dm_0}[\psi(\{\hat{\nu}^{[1]}_{k}\},\{\hat{q}^{[1]}_k\})-\psi(\{\nu_{k}\},\{{q}_k\})]|\mathcal{D}_1]=\smallo_{p}(1).\qedhere
\end{align*}

\paragraph*{ \bf{The term \cref{eq:term1_dr3} is $\smallo_{p}(1)$.}}

\begin{align*}
&\sqrt{N}\rE[\psi(\{\hat{\nu}^{[1]}_{k}\},\{\hat{q}^{[1]}_k\})-\rE[\psi(\{\nu_{k}\},\{{q}_k\})]|\{\hat{\nu}^{[1]}_{k}\},\{\hat{q}^{[1]}_k\}]\\ 
&=\sqrt{N}\rE[\sum_{k=0}^{\omega_N}\gamma^k (\hat{\nu}^{[1]}_{k}-\nu_{k} )(-\hat{q}^{[1]}_k+q_k)+\gamma^{k}(\hat{\nu}^{[1]}_{k-1}-\nu_{k-1})(-\hat{v}_k+v_k)|\{\hat{\nu}^{[1]}_{k}\},\{\hat{q}^{[1]}_k\}]\\
    &+\sqrt{N}\rE[\sum_{k=0}^{\omega_N}\gamma^k \nu_{k}(\hat{q}^{[1]}_k-q_k)+\gamma^k \nu_{k-1}(\hat{v}^{[1]}_k-v_k)|\{\hat{\nu}^{[1]}_{k}\},\{\hat{q}^{[1]}_k\}] \\
    &+\sqrt{N}\rE[\sum_{k=0}^{\omega_N}\gamma^k(\hat{\nu}^{[1]}_{k}-\nu_{k} )(r_k-q_k+v_{k+1})|\{\hat{\nu}^{[1]}_{k}\},\{\hat{q}^{[1]}_k\}] \\
&=\sqrt{N}\rE[\sum_{k=0}^{\omega_N}\gamma^k(\hat{\nu}^{[1]}_{k}-\nu_{k} )(-\hat{q}^{[1]}_k+q_k)+\gamma^k(\hat{\nu}^{[1]}_{k-1}-\nu_{k-1})(-\hat{v}^{[1]}_k+v_k)|\{\hat{\nu}^{[1]}_{k}\},\{\hat{q}^{[1]}_k\}] \\
&= \sqrt{N}\sum_{k=0}^{\omega_N} \mathcal{O}(\gamma^k \|\hat{\nu}^{[1]}_{k}-\nu_{k}\|_{2}\|\hat{q}^{[1]}_{k}-q_{k}\|_{2}+\gamma^k \|\hat{\nu}^{[1]}_{k-1}-\nu_{k-1}\|_{2}\|\hat{q}^{[1]}_{k}-q_{k}\|_{2} ) =\sqrt{N}\sum_{k=0}^{\omega_N}\kappa^\nu_N \kappa^q_N=\smallo_{p}(1).  %
\end{align*}
Here, we have used the assumption (\ref{thm:m1_inf}e). 

\paragraph*{\bf{Combining all things}}

Finally, we get 
\begin{align*}
    \sqrt{N}(\P_{\dm_0}\psi(\{\hat{\nu}^{[1]}_{k}\},\{\hat{q}^{[1]}_k\})-\rho^{\epol}_{\omega_N})=\sqrt{N/N_0}\bG_{\dm_0}[\psi(\{\nu_{k}\},\{q_k\}) ]+\smallo_{p}(1).
\end{align*}
By flipping the role, 
\begin{align*}
   \sqrt{N}(\P_{\dm_1}\psi(\{\hat{\nu}^{[0])}_{k}\},\{\hat{q}^{[0])}_k\})-\rho^{\epol}_{\omega_N})=\sqrt{N/N_1}\bG_{\dm_1}[\psi(\{\nu_{k}\},\{q_k\}) ]+\smallo_{p}(1).
\end{align*}
Therefore, 
\begin{align*}
    &\sqrt{N}(\hat{\rho}_{\mathrm{DRL(\cm_1)}}-\rho^{\epol}_{\omega_N})\\
    &=  N_0/N\times \sqrt{N}\psi(\{\hat{\nu}^{[1]}_{k}\},\{\hat{q}^{[1]}_k\})-\rho^{\epol}_{\omega_N})+ N_1/N\times \sqrt{N}(\P_{\dm_1}\psi(\{\hat{\nu}^{[0]}_{k}\},\{\hat{q}^{[0]}_k\})-\rho^{\epol}_{\omega_N}) \\
    &=\sqrt{N_0/N}\bG_{N_0}[\psi(\{\nu_{k}\},\{q_k\})  ]+\sqrt{N_1/N}\bG_{N_1}[\psi(\{\nu_{k}\},\{q_k\}) ]+\smallo_{p}(1)\\
    &=\bG_{N}[\psi(\{\nu_{k}\},\{q_k\}) ] +\smallo_{p}(1).
\end{align*}
Finally, from the assumption (\ref{thm:m1_inf}a),(\ref{thm:m1_inf}b)(\ref{thm:m1_inf}d) and CLT, the efficiency is concluded. This estimator is also regular from \cref{thm:gradients}.

\end{myproof}
}

\begin{myproof}[Proof of \cref{thm:m2_inf}]
The proof is similar to that of \cref{thm:m1_inf}
\end{myproof}

{\newedit 
\begin{myproof}[Proof of \cref{thm:m3_inf,thm:m3_inf_adaptive}]
First, We provide the proof the cross-fitting version. Then, we provide the proof of the adaptive version. 

In the cross-fitting version, the estimator is given by 
\begin{align*}
\frac{n_0}{n}\P_{\Dcal_0}\psi(\hat w^{[1]},\hat q^{[1]})+\frac{n_1}{n}\P_{\Dcal_1}\psi(\hat w^{[0]},\hat q^{[0]}) 
\end{align*}
where $\P_{\Dcal_0}$ is a sample average over a set of samples in one fold, and  $\P_{\Dcal_1}$ is a sample average over a set of samples in another fold. Then, we have 
\begin{align}
    \sqrt{n}(\P_{\dm_0}\psi(\hat w^{[1]},\hat q^{[1]})-\rho^{\epol})&= \sqrt{n/n_0}\bG_{n_0}[\psi(\hat w^{[1]},\hat q^{[1]})-\psi(w^{[1]}, q^{[1]})]
    \label{eq:term3_drl}
    \\
    &+\sqrt{n/n_0}\bG_{n_0}[\psi(w^{[1]}, q^{[1]})] 
       \label{eq:term3_dr2}
    \\
    &+\sqrt{n}(\rE[\psi(\hat w^{[1]},\hat q^{[1]})\mid \hat w^{[1]},\hat q^{[1]} ]-\rho^{\epol} ).    \label{eq:term3_dr3}
\end{align}

We analyze each term. Here, we have 
\begin{align*}
\psi(\hat{w},\hat{q})-\psi(w,q)&=D_1+D_2+D_3, \\
  D_1 & =\{\hat{w}(s)-w(s)\}\eta(s,a)\{r-q(s,a)+\gamma v(s')\}, \\
D_2 &= w(s)\eta(s,a)\{q(s,a)-\hat{q}(s,a)+\gamma \hat{v}(s')-\gamma v(s')\}+(1-\gamma)\rE_{p^{(0)}_{\epol}}[v'(s)-v(s)], \\
 D_3 &=\{\hat{w}(s)-w(s)\}\eta(s,a)\{q(s,a)-\hat{q}(s,a)+\gamma \hat{v}(s')-\gamma v(s')\}.
\end{align*}

\paragraph{\bf{Term \eqref{eq:term3_drl} is $\op(1)$ }}

We show that the conditional variance given $\Dcal_1$ is  $\op(1)$. The rest of the argument is the same as the proof of \cref{thm:m1_inf}. 
The conditional variance is 
\begin{align*}
    \var[\sqrt{n_0}\P_{\Dcal_0}[ \psi(\hat{w},\hat{q})-\psi(w,q)    ]|\Dcal_1 ]&\leq \E[D^2_1+D^2_2+D^3_2+2D_1D_2+2D_2D_3+2D_1D_3 |\Dcal_1 ] \\ 
    &=\op(1). 
\end{align*}
Here, we use 
\begin{align*}
     \|\hat v(s')-v(s')\|_2 \leq \sqrt{C_{S'}}\|\hat q(s,a)-q(s,a)\|_2. 
\end{align*}
from Jensen's inequality. 

\paragraph{\bf{Term \eqref{eq:term3_dr3} is $\op(1)$ }}

\begin{align*}
   | \E[D_1+D_2+D_3|\Dcal_1]|&\leq \|\hat w-w\|_2\|\eta\|_2\|\hat q-q\|_2+ \|\hat w-w\|_2\|\eta\|_2\|\hat v(s')-v(s)\|_2 \\ 
   &\leq \|\hat w-w\|_2\|\eta\|_2\|\hat q-q\|_2+\sqrt{\gamma C_S'} \|\hat w-w\|_2\|\eta\|_2\|\hat q-q\|_2 \\ 
   &=\kappa^w_n\kappa^q_n=\op(n^{-1/2}). 
\end{align*}

\paragraph{\bf{Proving efficiency} }

\begin{align*}
    &\sqrt{n}(\P_{\dm_0}\psi(\hat w^{[1]},\hat q^{[1]})+\P_{\dm_1}\psi(\hat w^{[0]},\hat q^{[0]})-\rho^{\epol})\\
    &=\sqrt{n/n_0}\bG_{n_0}[\psi(w^{}, q^{})]+ \sqrt{n/n_1}\bG_{n_1}[\psi(w^{}, q^{})]+\op(n^{-1/2})\\
    &=\bG_{n}[\psi(w^{}, q^{})] +\op(n^{-1/2}). 
\end{align*}
Finally, from CLT, the final statement is concluded. 

{\newedit 
\paragraph*{\bf{Proof of adaptive version}}

The adaptive version is similarly proved. 
\begin{align}
    \sqrt{n}(\P_{n}\psi(\hat w^{[1]},\hat q^{[1]})-\rho^{\epol})&=\bG_{n}[\psi(\hat w^{[1]},\hat q^{[1]})-\psi(w^{[1]}, q^{[1]})]
    \label{eq:term3_drl_ada}
    \\
    &+\bG_{n}[\psi(w^{[1]}, q^{[1]})] 
       \label{eq:term3_dr2_ada}
    \\
    &+\sqrt{n}(\rE[\psi(\hat w^{[1]},\hat q^{[1]})\mid \hat w^{[1]},\hat q^{[1]} ]-\rho^{\epol} ).    \label{eq:term3_dr3_ada}
\end{align}
The third term is $\op(1)$ following the similar logic as before. We prove the first term is $\op(1)$. From \citet[Lemma 19.24]{VaartA.W.vander1998As}, what we need to prove is 
\begin{align*}
    \mathbb{F}=\{\psi(w,q):w\in \Fcal_{w}, q\in \Fcal_q\}
\end{align*}
belongs to a Donsker class. If we can prove the uniform entropy integral:
\begin{align*}
    \int_{0}^1 \sqrt{\log \sup_{U}\mathcal{N}(\tau, \mathbb{F},L_2(U))}d(\tau), 
\end{align*}
is finite, from \citet[Theorem 19.14]{VaartA.W.vander1998As}, the class $\mathbb{F}$ belongs to a Donsker class. Here, 
\begin{align*}
    \log \sup_{U}\mathcal{N}(\tau, \mathbb{F},L_2(U)) & \leq    \log \sup_{U}\mathcal{N}(\tau, \mathbb{F},L_{2}(U))\\
    &\leq C \{\log \mathcal{N}(\tau,\Fcal_{w},L_{\infty}(U))+\log \sup_{U}\mathcal{N}(\tau, \Fcal_q,L_{\infty}(\cdot)) \} \\ 
    &\leq C (1/\tau)^{\beta}< C (1/\tau)^{2}.  
\end{align*}
From the second line to the third line, we use \citet[Leemma 9]{UeharaMasatoshi2021FSAo}. Thus, the uniform entropy integral is finite, which leads to the conclusion that first term \eqref{eq:term3_drl_ada} is $\op(1)$.
}

\end{myproof}
}

{\blockedit 
\begin{myproof}[Proof of \cref{thm:db_m3}]
We provide the proof the cross-fitting version. The estimator is given by 
\begin{align*}
\frac{n_0}{n}\P_{\Dcal_0}\psi(\hat w^{[1]},\hat q^{[1]})+\frac{n_1}{n}\P_{\Dcal_1}\psi(\hat w^{[0]},\hat q^{[0]}) 
\end{align*}
where $\P_{\Dcal_0}$ is a sample average over a set of samples in one fold, and  $\P_{\Dcal_1}$ is a sample average over a set of samples in another fold. Then, we have 
\begin{align}
  \P_{\dm_0}\psi(\hat w^{[1]},\hat q^{[1]})&= \sqrt{1/n_0}\bG_{n_0}[\psi(\hat w^{[1]},\hat q^{[1]})-\psi(w^{\dagger}, q^{\dagger})]
    \label{eq:term3_drl_m3}
    \\
    &+\rE[\psi(\hat w^{[1]},\hat q^{[1]})\mid \hat w^{[1]},\hat q^{[1]} ]-\rE[\psi( w^{\dagger}, q^{\dagger})]   \label{eq:term3_dr3_m3} \\ 
    &+  \P_{\dm_0}\psi( w^{\dagger}, q^{\dagger}). 
\end{align}
Then, we have 
\begin{align*}
\rE[\psi(\hat{w},\hat{q})|\hat{w},\hat{q}]&=\rE[\{\hat{w}(s)-w^{\dagger}(s)\}\eta(s,a)\{r-q^{\dagger}(s,a)+\gamma v^{\dagger}(s')\}]+ \\
 &+\rE[w^{\dagger}(s)\eta(s,a)\{q^{\dagger}(s,a)-\hat{q}(s,a)+\gamma \hat{v}(s')-\gamma v^{\dagger}(s')\}]+(1-\gamma)\rE_{p^{(0)}_{e}}[v'(s)-v^{\dagger}(s)] \\
 &+\rE[\{\hat{w}(s)-w^{\dagger}(s)\}\eta(s,a)\{q^{\dagger}(s,a)-\hat{q}(s,a)+\gamma \hat{v}(s')-\gamma v^{\dagger}(s')\}] \\
 &+\rE[\{\hat{v}(s')-v^{\dagger}(s')-\hat{q}(s,a)+q^{\dagger}(s,a)\}|\hat{w},\hat{q}]\\
&+\rE[\psi(w^{\dagger},q^{\dagger})|\hat{w},\hat{q}] \\
&=\mathcal{O}(\| \hat{w}(s)-w^{\dagger}(s)\|_2,\|\hat{q}(s,a)-q^{\dagger}(s,a)\|_2,\|\hat{v}(s')-v^{\dagger}(s') \|_{2})+\rE[\psi(w^{\dagger},q^{\dagger})] \\
&=\op(1)+\rE[\psi(w^{\dagger},q^{\dagger})].
\end{align*}
Thus, the term \eqref{eq:term3_dr3_m3} is $\op(1)$. Thus, the remaining part is proving when at least one model is well-specified, the term $\P_{\dm_0}\psi( w^{\dagger}, q^{\dagger})$ is consistent. 

\paragraph*{\bf{$q$-model is well-specified .}}

Consider the case where $q^{\dagger}(s,a)=q(s,a)$;
\begin{align*}
    &\rE[\psi(w^{\dagger},q^{\dagger})]=\rE[(1-\gamma)\rE_{p^{(0)}_{e}}[v(s)]+w^{\dagger}(s)\eta(s,a)\{r+\gamma v(s')-q(s,a)\}] \\
    &= (1-\gamma)\rE_{p^{(0)}_{\epol}}[v(s)]=\rho^{\epol}. 
\end{align*}
This implies when the q-model is consistent, the estimator $\hat{\rho}_{\mathrm{DRL(\cm_3)}}$ is also consistent. 

\paragraph*{\bf{$w$-model is well-specified.}}

Consider the case where $w^{\dagger}(s)=w(s)$;
\begin{align}
    \rE[\psi(w^{\dagger},q^{\dagger})]&=\rE[(1-\gamma)\rE_{p^{(0)}_{e}}[v^{\dagger}(s)]+w(s)\eta(s,a)\{r+\gamma v^{\dagger}(s')-q^{\dagger}(s,a)\}] \nonumber\\
    &=(1-\gamma)\rE_{p^{(0)}_{e}}[v^{\dagger}(s)]+\rE[w(s)\eta(s,a)r]+\rE[w(s)\eta(s,a)\gamma v^{\dagger}(s')] \label{eq:db1}\\
    &-\rE[w(s)\eta(s,a) q^{\dagger}(s,a)] \nonumber \\ 
    &=\rE[w(s)\eta(s,a)r]+\rE[w(s) v^{\dagger}(s)]-\rE[w(s)\eta(s,a) q^{\dagger}(s,a)] \label{eq:db2}\\ 
    &=\rE[w(s)\eta(s,a)r]=\rho^{\epol}.\label{eq:db3}
\end{align}
From \eqref{eq:db1} to \eqref{eq:db2}, we use a result 
\begin{align*}
    (1-\gamma)\rE_{p^{(0)}_{\epol}}[v^{\dagger}(s)]+\rE[w(s)\eta(s,a)\gamma v^{\dagger}(s')]=\rE[w(s) v^{\dagger}(s)]
\end{align*}
from \cref{lem:ratio-estimation}. From \eqref{eq:db2} to \eqref{eq:db3}, we use a result 
\begin{align*}
    \rE[w(s)\eta(s,a) q^{\dagger}(s,a)]= \rE[w(s)\rE[\eta(s,a) q^{\dagger}(s,a)|s]]=\rE[w(s)v^{\dagger}(s)]. 
\end{align*}
This implies when the ratio model is correct, the estimator $\hat{\rho}_{\mathrm{DRL(\cm_3)}}$ is also consistent.

\end{myproof}
}

{\blockedit 
\begin{myproof}[Proof of \cref{thm:db_n}]
We only provide the proof the cross-fitting version. The adaptive version is similarly proved. The estimator is given by 
\begin{align*}
\frac{n_0}{n}\P_{\Dcal_0}\psi(\hat w^{[1]},\hat q^{[1]})+\frac{n_1}{n}\P_{\Dcal_1}\psi(\hat w^{[0]},\hat q^{[0]}). 
\end{align*}
We have the following decomposition: 
\begin{align}
    \sqrt{n}(\P_{\Dcal_0}\psi(\hat w^{[1]},\hat q^{[1]})-\rho^{\epol}) 
    &= \sqrt{n/n_0}\{\bG_{n_0}\psi(\hat{w}^{[1]},\hat{q}^{[1]})-\bG_{n_0}\psi(w^{\dagger},q^{\dagger})\}+ \label{eq:db13}\\
    &+\sqrt{n}(\rE[\psi(\hat{w}^{[1]},\hat{q}^{[1]})|\hat{w}^{[1]},\hat{q}^{[1]}]-\rE[\psi(w^{\dagger},q^{\dagger})]) \label{eq:db23} \\
    &+\sqrt{n}( \rE[\psi(w^{\dagger},q^{\dagger})]-\rho^{\epol}) \label{eq:db33}\\
    &+\sqrt{n/n_0}\bG_{n_0}\psi(w^{\dagger},q^{\dagger}). \nonumber
\end{align}
The term \eqref{eq:db13} is $\op(1)$ as in the proof of \cref{thm:m3_inf}. The term \eqref{eq:db23} is $\bigO_p(1)$ as in the proof of \cref{thm:db_m3}:
\begin{align*}
\rE[\psi(\hat{w}^{[1]},\hat{q}^{[1]})|\hat{w}^{[1]},\hat{q}^{[1]}]&=\mathcal{O}(\| \hat{w}^{[1]}(s)-w^{\dagger}(s)\|_2,\|\hat{q}^{[1]}(s,a)-q^{\dagger}(s,a)\|_2,\|\hat{v}^{[1]}(s')-v^{\dagger}(s') \|_{2})+\rE[\psi(w^{\dagger},q^{\dagger})] \\
&=\bigO_p(n^{-1/2})+\rE[\psi(w^{\dagger},q^{\dagger})].
\end{align*}

The term \eqref{eq:db33} is $0$ following the argument in the proof of \cref{thm:db_m3}. Then, we obtain $\sqrt{n}(\hat{\rho}_{\mathrm{DRL(\cm_3)}}-\rho^{\epol})=\bG_{n}\psi(w^{\dagger},q^{\dagger})+\bigO_p(1)=\bigO_p(1)$.
\end{myproof}
}

{\blockedit\begin{myproof}[Proof of \cref{thm:ideal_m3}]
For the ease of the notation, we assume $N=1$. The extension to general $N$ is straightforward. 

Under the assumptions, $\psi(s,a,r,s';w,q)$ is bounded.
Invoking \citet[Theorem 18.5.4]{ibragimov1975independent}, we obtain
that $\sqrt{T}\prns{\P_T[\psi(s,a,r,s';w,q)]-\rho^{\epol}}$ converges to a normal distribution with zero mean and the variance: 
\begin{align}\notag
    \mathrm{var}[\phi_\text{eff}(s_0,a_0,r_0,s_1)]+2\sum_{i=1}^{\infty}\mathrm{cov}[\phi_\text{eff}(s_0,a_0,r_0,s_1),\phi_\text{eff}(s_i,a_i,r_i,s_{i+1})], 
\end{align}
where $\phi_\text{eff}$ is as in \cref{thm:m3_bound}. The first term is $\mathrm{EB}(\cm_3)$.
The second term is zero because 
\begin{align*}
    &\rE[\phi_\text{eff}(s_0,a_0,r_0,s_1)\phi_\text{eff}(s_i,a_i,r_i,s_{i+1})] \\
    &= \rE[\phi_\text{eff}(s_0,a_0,r_0,s_1)\rE[\phi_\text{eff}(s_i,a_i,r_i,s_{i+1})\mid s_{i-1},a_{i-1},r_{i-1},s_{i}]]=0. \qedhere
\end{align*}
\end{myproof}}

{\blockedit
\begin{myproof}[Proof of \cref{thm:m3_eff_cro}]
Using the short-hand $\psi(w',q')$ for $\psi(s,a,r,s';w',q')$
and $N_j=\abs{\dm_j}$,
the estimator $\hat{\rho}_{\mathrm{DRL(\cm3)}}$ is given by 
\begin{align*}
\frac{N_0}{N}\P_{\dm_0}\P_{T}\psi(\hat{w}^{[1]},\hat{q}^{[1]})+\frac{N_1}{N}\P_{\dm_1}\P_{T}\psi(\hat{w}^{[0]},\hat{q}^{[[0]}),
\end{align*}
where $\P_{\dm_j}$ is the empirical average on the samples in $\dm_j$. 

Then, we have 
\begin{align}
    \sqrt{NT}(\P_{\dm_0}\P_{T}\psi(\hat{w}^{[1]},\hat{q}^{[1]})-\rho^{\epol})&= \sqrt{N/N_0}\bG_{\dm_0,T}[\psi(\hat{w}^{[1]},\hat{q}^{[1]})-\psi(w,q)]
    \label{eq:term1_dr31}
    \\
    &\phantom{=}+\sqrt{N/N_0}\bG_{\dm_0,T}[\psi(w,q)] 
      \nonumber
    \\
    &\phantom{=}+\sqrt{NT}(\rE[\psi(\hat{w}^{[1]},\hat{q}^{[1]})|\hat{w}^{[1]},\hat{q}^{[1]} ]-\rho^{\epol} )   \label{eq:term1_dr33},
\end{align}
where $\bG_{\dm_0,T}$ is an empirical process defined over the all sample in the first fold. 
Here, we have 
\begin{align*}
&\P_{T}\psi(\hat{w},\hat{q})-\P_{T}\psi(w,q) \\
&=\P_{T}[\{\hat{w}(s)-w(s)\}\eta(s,a)\{r-q(s,a)+\gamma v(s')\}] \\
 &\phantom{=}+\P_{T}[w(s)\eta(s,a)\{q(s,a)-\hat{q}(s,a)+\gamma \hat{v}(s')-\gamma v(s')\}]+(1-\gamma)\rE_{p^{(0)}_{\epol}}[v'(s)-v(s)] \\
 &\phantom{=}+\P_{T}[\{\hat{w}(s)-w(s)\}\eta(s,a)\{q(s,a)-\hat{q}(s,a)+\gamma \hat{v}(s')-\gamma v(s')\}].
\end{align*}

We analyze each term. First, we show the term \cref{eq:term1_dr31} is $\smallo_{p}(1)$. 

\paragraph*{\textbf{The term \eqref{eq:term1_dr31} is $\smallo_{p}(1)$}}

Consider the case $N_0=1$. The case with $N_0>1$ similarly holds.
If we can show that for any $\epsilon>0$, 
\begin{align}
\label{eq:process_m1_1}
\lim_{T\to \infty}\sqrt{T}P[&\P_{T}[\psi(\hat{w}^{[1]},\hat{q}^{[1]})-\psi(w,q)]\\&\notag- 
 \rE[\psi(\hat{w}^{[1]},\hat{q}^{[1]})-\psi(w,q)|\hat{w}^{[1]},\hat{q}^{[1]} ]>\epsilon|\mathcal{D}_1]=0, %
\end{align}
then, by bounded convergence theorem, we would have 
\begin{align*}
  \lim_{T\to \infty}\sqrt{T}P[&\P_{T}[\psi(\hat{w}^{[1]},\hat{q}^{[1]})-
  \psi(w,q)]\\&-\rE[\psi(\hat{w}^{[1]},\hat{q}^{[1]})-\psi(w,q)|\hat{w}^{[1]},\hat{q}^{[1]}]>\epsilon]=0,
\end{align*}
which yields the statement. 

To show \cref{eq:process_m1_1}, we show that the conditional mean is $0$ and conditional variance is $\smallo_p(1)$. 
The conditional mean is 
\begin{align*}
&\rE[\P_{T}[\psi(\hat{w}^{[1]},\hat{q}^{[1]})-\psi(w,q)|\hat{w}^{[1]},\hat{q}^{[1]}]\\&\qquad\qquad\qquad\qquad\qquad-\P[\psi(\hat{w}^{[1]},\hat{q}^{[1]})-\psi(w,q)]|\mathcal{D}_1]
=0.
\end{align*}
Here, we leverage the sample-splitting construction, that is, $\hat{w}^{[1]}$ and $\hat{q}^{[1]}$ only depend on $\mathcal{D}_1$, and $\mathcal{D}_1,\,\mathcal{D}_0$ are independent. The conditional variance is 
\begin{align*}
&\mathrm{var}[\sqrt{T}\P_{T}[\psi(\hat{w}^{[1]},\hat{q}^{[1]})-\psi(w,q)]|\mathcal{D}_1]=\rE\left[T\left\{\P_{T}[\psi(\hat{w}^{[1]},\hat{q}^{[1]})-\psi(w,q)]\right\}^2|\mathcal{D}_1\right]\\
&=\frac{1}{T}\left[\sum_{i=0}^{T} \max\{\Op((\kappa^w_N)^2),\Op((\kappa^q_N)^2)\}+2\sum_{i<j}^{T}\rho_{\|i-j\|} \max\{\Op((\kappa^w_N)^2),\Op((\kappa^q_N)^2)\}\right]\\
&=\smallo_p(1).
\end{align*}
Then, from Chebyshev's inequality:
\begin{align*}
&\sqrt{T}P[\P_{T}[\psi(\hat{w}^{[1]},\hat{q}^{[1]})-\psi(w,q)]-\rE[\psi(\hat{w}^{[1]},\hat{q}^{[1]})-\psi(w,q)|\hat{w}^{[1]},\hat{q}^{[1]} ]>\epsilon|\mathcal{D}_1] \\
&\leq \frac{1}{\epsilon^{2}}\mathrm{var}[\sqrt{T}\P_{T}[\psi(\hat{w}^{[1]},\hat{q}^{[1]})-\psi(w,q)]|\mathcal{D}_1]=\smallo_{p}(1).\qedhere
\end{align*}

\paragraph*{\textbf{Second Term}}
We show 
$$\sqrt{NT}(\rE[\psi(\hat{w}^{[1]},\hat{q}^{[1]})|\hat{w}^{[1]},\hat{q}^{[1]}]-\rho^{\epol})=\op(1).$$

Assume $N_0=1$ for simplicity. The case with $N_0>1$ similarly holds. Noting $\rE[(\hat{q}(s,a)-q(s,a))^{2}]\geq \rE[(\hat{v}(s')-v(s'))^{2}]$ from Jensen's inequality, we have 
\begin{align*}
&|\sqrt{T}\rE[\psi(s,a,r,s';\hat{w}^{[1]},\hat{q}^{[1]})-\psi(s,a,r,s';w,q)|\hat{w}^{[1]},\hat{q}^{[1]}] |\\
=&|\sqrt{T}\rE[(\hat{w}(s)-w(s))\eta(s,a)(r-q(s,a)+\gamma v(s'))|\hat{w}^{[1]},\hat{q}^{[1]}]+ \\
 &\sqrt{T}\rE[w(s)\eta(s,a)\{q(s,a)-\hat{q}(s,a)+\gamma \hat{v}(s')-\gamma v(s')\}+(1-\gamma)\rE_{p^{(0)}_{\epol}}[\hat{v}(s)-v(s)]||\hat{w},\hat{q}]+ \\
 &\sqrt{T}\rE[\{\hat{w}(s)-w(s)\}\eta(s,a)\{q(s,a)-\hat{q}(s,a)+\gamma \hat{v}(s')-\gamma v(s')\}|\hat{w}^{[1]},\hat{q}^{[1]}]|\\
 =&|\sqrt{T}\rE[\{\hat{w}(s)-w(s)\}\eta(s,a)\{q(s,a)-\hat{q}(s,a)+\gamma \hat{v}(s')-\gamma v(s')\}|\hat{w}^{[1]},\hat{q}^{[1]}]|\\
\leq & \sqrt{T}\|\hat{w}^{[1]}(s)-w(s)\|_{2}\|\eta(s,a)\|_2\|q(s,a)-\hat{q}^{[1]}(s,a)\|_{2}=\smallo_{p}(1). 
\end{align*}

\paragraph*{\textbf{Combining all results}}

Combining all result,  
\begin{align*}
    \sqrt{NT}(\hat{\rho}_{\mathrm{DRL(\cm_3)}}-\rho^{\epol})=\mathbb{G}_{N,T}\psi(w,q)+\op(1).
\end{align*}
Then, from \cref{eq:mixing relationships} and \citet[Theorem 18.5.4]{ibragimov1975independent} the statement is concluded as in \cref{thm:ideal_m3}. 
\end{myproof}
}

{\newedit
\begin{myproof}[Proof of \cref{thm:m3_eff_cro2}]
We focus on $N=1$, which is most relevant for cross-time-fitting; also $N>1$ easily follows similarly.
Using the short-hand $\psi(w',q')$ for $\psi(s,a,r,s';w',q')$ and $T_j=\abs{\ct_j}$,
the estimator $\hat{\rho}_{\mathrm{DRL}(\cm3)}$ is given by 
\begin{align*}
\frac{T_0}{T}\P_{\ct_0}\psi(\hat{w}^{[2]},\hat{q}^{[2]})+\frac{T_1}{T}\P_{\ct_1}\psi(\hat{\nu}^{[3]},\hat{q}^{[3]})+\frac{T_2}{T}\P_{\ct_2}\psi(\hat{\nu}^{[0]},\hat{q}^{[0]})+\frac{T_3}{T}\P_{\ct_3}\psi(\hat{\nu}^{[1]},\hat{q}^{[1]}),
\end{align*}
where $\P_{\ct_j}$ is the empirical average on the samples in $\ct_j$.

The proof now proceeds as in \cref{thm:m3_eff_cro}. 
In particular, we have to deal with the dependence across folds more carefully. 

First, we show the analysis of the stochastic equicontinuity term. To prove it, we must leverage the fact that $\ct_0$ and $\ct_1$ are separated by $T/4$ time steps, rather than being independent, unlike \cref{thm:m3_eff_cro}. 

\paragraph*{\textbf{First part}: $\bG_{\ct_0}[\psi(\hat{w}^{[2]},\hat{q}^{[2]})-\psi(w,q)]=\op(1)$ }
\begin{myproof}
If we can show that for any $\epsilon>0$, 
\begin{align}
\label{eq:process_m1_}
\lim_{T\to \infty}\sqrt{T_0}P[&\P_{\ct_0}[\psi(\hat{w}^{[2]},\hat{q}^{[2]})-\psi(w,q)]\\&\notag- 
 \rE[\psi(\hat{w}^{[2]},\hat{q}^{[2]})-\psi(w,q)|\hat{w}^{[2]},\hat{q}^{[2]} ]>\epsilon|\mathcal{T}_2]=0, %
\end{align}
Then, by bounded convergence theorem, we would have 
\begin{align*}
  \lim_{T\to \infty}\sqrt{T_0}P[&\P_{\ct_0}[\psi(\hat{w}^{[2]},\hat{q}^{[2]})-
  \psi(w,q)]\\&-\rE[\psi(\hat{w}^{[2]},\hat{q}^{[2]})-\psi(w,q)|\hat{w}^{[2]},\hat{q}^{[2]}]>\epsilon]=0,
\end{align*}
yielding the statement. 

To show \eqref{eq:process_m1_}, we show that the conditional mean given $\mathcal{T}_2$ is $\smallo_p(1)$ and conditional variance given $\mathcal{T}_2$ is $\smallo_p(1)$. The conditional mean given $\mathcal{T}_2$ is 
\begin{align}
&|\E[\P_{\ct_0}[\psi(\hat{w}^{[2]},\hat{q}^{[2]})-\psi(w,q)]-\E[\psi(\hat{w}^{[2]},\hat{q}^{[2]})-\psi(w,q)]|\mathcal{T}_2]|  \nonumber  \\
&=|\E[\P_{\ct_0}[\psi(\hat{w}^{[2]},\hat{q}^{[2]})-\psi(w,q)]|\ct_2]-\E[\psi(\hat{w}^{[2]},\hat{q}^{[2]})-\psi(w,q) \mid \hat{w}^{[2]},\hat{q}^{[2]}] | \label{eq:show2} \\
&\leq |\P[\psi(\hat{w}^{[2]},\hat{q}^{[2]})-\psi(w,q)| \mid \hat{w}^{[2]},\hat{q}^{[2]}]-\E[\psi(\hat{w}^{[2]},\hat{q}^{[2]})-\psi(w,q) \mid \hat{w}^{[2]},\hat{q}^{[2]}]|+\op(1) \label{eq:show3} \\
&= \smallo_p(1).\nonumber 
\end{align}

Here, we leverage the sample splitting construction, that is, $\hat{w}^{[2]}_{k}$ and $\hat{q}^{[2]}_{k}$ only depend on $\mathcal{T}_2$, and the correlation between the train data and test data set is small.
To go from \cref{eq:show2} to  \cref{eq:show3}, we use the following argument.
First, note $\alpha_m \leq \phi_m$ by \cref{eq:mixing relationships}. 
In addition, from the definition of $\alpha$-mixing and its moment inequality, for any bounded function $f(x)$ with a first moment, based on \citet[Theorem 14.2]{DavidsonJamesE.H1994Slta}, we have 
\begin{align*}
    \|\E[f(s_t,a_t,s_{t+1},r_t)\mid s_0,a_0,s_1,r_0]-\E[f(s_t,a_t,s_{t+1},r_t)]\|_1\leq 6\|f(s_t,a_t,s_{t+1},r_t)\|_{1}. 
\end{align*}
Note that since $\alpha$-mixing is time-reversible \citep[Page 209]{DavidsonJamesE.H1994Slta}, the following also holds:
\begin{align*}
    \|\E[f(s_0,a_0,s_{1},r_1)\mid s_t,a_t,s_{t+1},r_t]-\E[f(s_0,a_0,s_{1},r_1)]\|_1\leq 6\|f(s_0,a_0,s_{1},r_1)\|_{1}. 
\end{align*}
By applying this moment inequality to \cref{eq:show2} and noting $\psi$ is a bounded function we have
\begin{align*}
    &|\E[\P_{\ct_0}[\psi(\hat{w}^{[2]},\hat{q}^{[2]})-\psi(w,q)]\mid \Tcal_2]-\E[\psi(\hat{w}^{[2]},\hat{q}^{[2]})-\psi(w,q) |\hat w^{[2]},\hat q^{[2]}]|\\
    &= |\frac{1}{T_0}\sum_{i\in T_0}\int \{\psi(z_i;\hat{w}^{[2]},\hat{q}^{[2]})-\psi(z_i;w,q)\} p(z_i\mid \Tcal_2)\mathrm{d}(z_i)
    -\int \braces{\psi(z_i;\hat{w}^{[2]},\hat{q}^{[2]})-\psi(z_i;w,q)} p(z_i)\mathrm{d}(z_i) |\\
    &\leq\frac{1}{T_0}\sum_{i\in T_0}|\int \{\psi(z_i;\hat{w}^{[2]},\hat{q}^{[2]})-\psi(z_i;w,q)\} p(z_i\mid \Tcal_2)\mathrm{d}(z_i) -\int \braces{\psi(z_i;\hat{w}^{[2]},\hat{q}^{[2]})-\psi(z_i;w,q)} p(z_i)\mathrm{d}(z_i) |\\
    &\leq 6 \int |\braces{\psi(z_i;\hat{w}^{[2]},\hat{q}^{[2]})-\psi(z_i;w,q)}| p(z_i)\mathrm{d}(z_i) \tag{$\alpha$-mixing moment inequality}\\
    &= 6\E[ |\psi(\hat{w}^{[2]},\hat{q}^{[2]})-\psi(w,q) | \mid \hat w^{[2]},\hat q^{[2]}]=\op(1).  
\end{align*}
Then, 
\begin{align*}
    \E[\P_{\ct_0}[\psi(\hat{w}^{[2]},\hat{q}^{[2]})-\psi(w,q)]\mid \Tcal_2]=\E[\psi(\hat{w}^{[2]},\hat{q}^{[2]})-\psi(w,q) |\hat w^{[2]},\hat q^{[2]}]+\op(1). 
\end{align*}

Besides, the conditional variance is 
\begin{align}
&\mathrm{var}[\sqrt{T_0}\P_{\ct_0}[\psi(\hat{w}^{[2]},\hat{q}^{[2]})-\psi(w,q)]|\mathcal{T}_2] \leq \mathrm{E}[T_0\{\P_{\ct_0}[\psi(\hat{w}^{[2]},\hat{q}^{[2]})-\psi(w,q)]\}^2|\mathcal{T}_2]  \nonumber \\
&\leq \frac{1}{T_0}\sum_{i,j\in \ct_0}\int | \{\psi(z_i;\hat{w}^{[2]},\hat{q}^{[2]})-\psi(z_i;w,q)\}\{\psi(z_j:\hat{w}^{[2]},\hat{q}^{[2]})-\psi(z_j;w,q)\}  p(z_i,z_j\mid \Tcal_2)\mathrm{d}(z_i,z_j) |   \label{eq:var2} \\
&=\frac{1}{T_0}\sum_{i,j\in \ct_0}\int | \{\psi(z_i;\hat{w}^{[2]},\hat{q}^{[2]})-\psi(z_i;w,q)\}\{\psi(z_j:\hat{w}^{[2]},\hat{q}^{[2]})-\psi(z_j;w,q)\}  p(z_i,z_j)\mathrm{d}(z_i,z_j) |+\op(1)  \label{eq:var5} \\
&=\frac{1}{T_0}\left[\sum_{i=0}^{T_0} \max\{\Op((\kappa^w_N)^2),\Op((\kappa^q_N)^2)\}+2\sum_{i<j}^{T_0}\rho_{\|i-j\|} \max\{\Op((\kappa^w_N)^2),\Op((\kappa^q_N)^2)\}\right]+\smallo_p(1)  \label{eq:var3}\\
&=\smallo_p(1). \label{eq:var4}
\end{align}
Here, we used a $\alpha$-mixing condition and its moment inequality from \eqref{eq:var2} to \eqref{eq:var5}. Then, we used a $\rho$-mixing condition based on $\rho_t\leq 2\sqrt{\phi_t}=2/t^{1+\epsilon}$ (see \cref{eq:mixing relationships}) from \eqref{eq:var3} to \eqref{eq:var4}. 

Finally, based on the obtained conditional mean and conditional variance, from Chebyshev's inequality, the rest of the proof is concluded. 
\end{myproof}

\paragraph{\textbf{Second part }}
We prove 
\begin{align*}
\rE[\P_{\ct_0}[\psi(s,a,r,s';\hat{w}^{[2]},\hat{q}^{[2]})]\mid \hat{w}^{[2]},\hat{q}^{[2]}]-\rE[\psi(s,a,r,s';w,q)] = \op(T^{-1/2}).
\end{align*}
Note from \citet[Theorem 14.2]{DavidsonJamesE.H1994Slta}, 
\begin{align*}
    \|\E[f(s_0,a_0,s_{1},r_1)\mid s_t,a_t,s_{t+1},r_t]-\E[f(s_0,a_0,s_{1},r_1)]\|_1\leq 6\alpha^{1/2}_{T_0}\|f(s_0,a_0,s_{1},r_1)\|_{2}. 
\end{align*}
Then, 
\begin{align*}
    &|\E[\P_{\ct_0}[\psi(\hat{w}^{[2]},\hat{q}^{[2]})-\psi(w,q)]\mid \Tcal_2]-\E[\psi(\hat{w}^{[2]},\hat{q}^{[2]})-\psi(w,q) |\hat w^{[2]},\hat q^{[2]}]|\\
    &=\left|\E\left[\frac{1}{T_0}\sum_{i\in T_0}\int \{\psi(z_i;\hat{w}^{[2]},\hat{q}^{[2]})-\psi(z_i;w,q)\} p(z_i\mid \Tcal_2)\mathrm{d}(z_i) -\int \braces{\psi(z_i;\hat{w}^{[2]},\hat{q}^{[2]})-\psi(z_i;w,q)} p(z_i)\mathrm{d}(z_i)\right] \right|\\
    &\leq\frac{1}{T_0}\sum_{i\in T_0}\E[|\int \{\psi(z_i;\hat{w}^{[2]},\hat{q}^{[2]})-\psi(z_i;w,q)\} p(z_i\mid \Tcal_2)\mathrm{d}(z_i) -\int \braces{\psi(z_i;\hat{w}^{[2]},\hat{q}^{[2]})-\psi(z_i;w,q)} p(z_i)\mathrm{d}(z_i) |] \\
    &\leq 6\int |\braces{\psi(z_i;\hat{w}^{[2]},\hat{q}^{[2]})-\psi(z_i;w,q)}| p(z_i)\mathrm{d}(z_i) \tag{$\alpha$-mixing moment inequality}\\
    &= 6\alpha^{1/2}_{T_0}\E[ |\psi(\hat{w}^{[2]},\hat{q}^{[2]})-\psi(w,q) |^2 \mid \hat w^{[2]},\hat q^{[2]}]=\op(1/\sqrt{T_0}) \tag{$\alpha_{T_0}\leq \phi_{T_0}$}.  
\end{align*}
To sum up, we have 
\begin{align}\label{eq:key_double}
    \E[\P_{\ct_0}[\psi(\hat{w}^{[2]},\hat{q}^{[2]})-\psi(w,q)]\mid \Tcal_2]=\E[\psi(\hat{w}^{[2]},\hat{q}^{[2]})-\psi(w,q) |\hat w^{[2]},\hat q^{[2]}]+\op(T^{-1/2}_0). 
\end{align}
Using this above, we have 
\begin{align}
&\sqrt{T}\rE[\P_{\ct_0}[\psi(s,a,r,s';\hat{w}^{[2]},\hat{q}^{[2]})]\mid \hat{w}^{[2]},\hat{q}^{[2]}] \nonumber \\
&=\sqrt{T}\rE[\rE[\P_{\ct_0}[\psi(s,a,r,s';\hat{w}^{[2]},\hat{q}^{[2]})] \mid \hat{w}^{[2]},\hat{q}^{[2]},\ct_2]\mid \hat{w}^{[2]},\hat{q}^{[2]}] \nonumber \\
&=\sqrt{T}\rE[\rE[\P_{\ct_0}[\psi(s,a,r,s';\hat{w}^{[2]},\hat{q}^{[2]})] \mid \ct_2]\mid \hat{w}^{[2]},\hat{q}^{[2]}] \nonumber\\
&=\sqrt{T}\rE[\rE[\P_{\ct_0}[\psi(s,a,r,s';\hat{w}^{[2]},\hat{q}^{[2]})]\mid \hat{w}^{[2]},\hat{q}^{[2]}]\mid \hat{w}^{[2]},\hat{q}^{[2]}]+\op(1) \tag{Use \cref{eq:key_double} } \\
&=\sqrt{T}\rE[\psi(s,a,r,s';\hat{w}^{[2]},\hat{q}^{[2]}) \mid \hat{w}^{[2]},\hat{q}^{[2]}]+\op(1) \nonumber  \\
&=\sqrt{T}\rE[\psi(s,a,r,s';w,q)]+\op(1).  \tag{The same as the proof of \cref{thm:m3_eff_cro}}
\end{align}
\paragraph{\textbf{Summary}}

Combining,  we have
\begin{align*}
    \sqrt{T}(\hat{\rho}_{\mathrm{DRL(\cm_3)}}-\rho^{\epol})=\mathbb{G}_{T}\psi(w,q)+\op(1). 
\end{align*}
The statement is concluded as in \cref{thm:ideal_m3}. 

\end{myproof}

\begin{myproof}[Proof of \cref{thm:m3_eff}]
For the ease of notation we assume $N=1$. The extension to general $N$ is straightforward. We use the same shorthand as in the proofs of \cref{thm:m3_eff_cro,thm:m3_eff_cro2}.

We have the following decomposition:
\begin{align}\label{eq:m3_eff_all}
    \sqrt{T}(\hat{\rho}_{\mathrm{DRL(\cm_3)}}-\rho^{\epol})&=\bG_{T}\psi(\hat{w},\hat{q})-\bG_{T}\psi(w,q)+\bG_{T}\psi(w,q)+\sqrt{T}(\rE[\psi(\hat{w},\hat{q})|\hat{w},\hat{q}]-\rho^{\epol}).
\end{align}

We again analyze each term.

\paragraph*{\textbf{First term}: $\bG_{T}\psi(\hat{w},\hat{q})-\bG_{T}\psi(w,q)=\op(1)$.}

From Theorem 11.24 of \citet{KosorokMichaelR2008ItEP} based on our assumptions, $\mathbb{G}_{T}\stackrel{d}{\rightarrow} \bh$ in $L^{\infty}(p_b)$, where $\bh$ is a tight mean zero Gaussian process with some covariance. By (\ref{thm:m3_eff}d), we have
$\|\psi(\hat{w},\hat{q})-\psi(w,q)\|_2=\op(1)$. Then by Lemma 18.5 of \citet{VaartA.W.vander1998As}, the statement is concluded.

\paragraph*{\textbf{Second term}: $\sqrt{T}(\rE[\psi(\hat{w},\hat{q})|\hat{w},\hat{q}]-\rho^{\epol})=\op(1).$}

The derivation is done as in the proof of \cref{thm:m3_eff_cro}. 

\paragraph*{\textbf{Summary}}
Combining,  we have
\begin{align*}
    \sqrt{T}(\hat{\rho}_{\mathrm{DRL(\cm_3)}}-\rho^{\epol})=\mathbb{G}_{T}\psi(w,q)+\op(1). 
\end{align*}
The statement is concluded as in \cref{thm:ideal_m3}. 
\end{myproof}}

{\newedit 

\begin{myproof}[Proof of \cref{thm:variance est}]
We provide the proof of cross-fitting version. The adaptive version is similarly proved. 
The estimator is given by 
\begin{align*}
\frac{n_0}{n}\P_{\Dcal_0}\psi^2_{eff}(\hat w^{[1]},\hat q^{[1]})+\frac{n_1}{n}\P_{\Dcal_1}\psi^2_{eff}(\hat w^{[0]},\hat q^{[0]}) 
\end{align*}
where $\P_{\Dcal_0}$ is a sample average over a set of samples in one fold, and  $\P_{\Dcal_1}$ is a sample average over a set of samples in another fold. Then, we have 
\begin{align}
 (\P_{\dm_0}\psi^2_{eff}(\hat w^{[1]},\hat q^{[1]})-\E[\psi^2_{eff}(w, q) ])&= \P_{\dm_0}[\psi^2_{eff}(\hat w^{[1]},\hat q^{[1]})-\psi^2_{eff}(w, q)]
    \label{eq:term3_drl_var}
    \\
    &+\sqrt{n/n_0}\bG_{n_0}[\psi^2_{eff}(w, q)]    \label{eq:term3_dr2_var}
    \\
    &+(\rE[\psi^2_{eff}(\hat w^{[1]},\hat q^{[1]})\mid \hat w^{[1]},\hat q^{[1]} ]-\E[\psi^2_{eff}(w, q) ] ).    \label{eq:term3_dr3_var}
\end{align}

We analyze each term. Here, we have 
\begin{align*}
|\psi^2_{eff}(\hat{w},\hat{q})-\psi^2_{eff}(w,q)|&=|\hat w^2\{r-\hat q(s,a)+\gamma \hat v(s')\}^2- w^2\{r- q(s,a)+\gamma v(s')\}^2|\\
&\lesssim \max\{|\hat w-w|,|\hat q-q|,|\hat v-v|\}
\end{align*}
for some constant $C$. 

\paragraph{\bf{Term \eqref{eq:term3_drl_var} is $\bigO_p(1/\sqrt{n})$ }}

The conditional mean given $\Dcal_1$ is $0$. The conditional variance given $\Dcal_1$ is  $\bigO_p(1/n)$. The rest of the argument is the same as the proof of \cref{thm:m3_inf}. 

\paragraph{\bf{Term \eqref{eq:term3_dr2_var} is $\bigO_p(1/\sqrt{n})$ }}

This is obvious. 

\paragraph{\bf{Term \eqref{eq:term3_dr3_var} is $\op(1)$ }}

\begin{align*}
    |\E[\psi^2_{eff}(\hat{w},\hat{q})-\psi^2_{eff}(w,q)\mid \Dcal_1]|&\leq C\max\{\|\hat w-w\|_2,\|\hat q-q\|_2,\|\hat v-v\|_2\}= \op(1).  
 \end{align*}

\end{myproof}
}

{\newedit 
\begin{myproof}[Proof of \cref{lem:ratio-estimation}]
Define 
\begin{align*}
    \delta(g,s')=\gamma \int p(s'|s)g(s)\mathrm{d}\lambda(s)-g(s')+(1-\gamma)p^{(0)}_{\epol}(s'),
\end{align*}
where $g(s)$ is any function and $p(s'|s)$ is a marginal distribution of $p(s'|s,a)\epol(a|s)$. Then, 
\begin{align*}
&L(w,f_w)\\
&= \rE_{p_b}[\{\gamma w(s)\eta(s,a)f_{w}(s')-w(s)f_{w}(s)\}]+(1-\gamma)\rE_{p_{\epol}^{(0)}}[f_{w}(s)] \\
    &= \rE_{p^{(\infty)}_{e,\gamma}}[(p^{(\infty)}_{e,\gamma}(s)/p_{b}(s))^{-1}\gamma w(s)\eta(s,a)f_{w}(s')]-\rE_{p^{(\infty)}_{e,\gamma}}[(p^{(\infty)}_{e,\gamma}(s)/p_{b}(s))^{-1} w(s)f_{w}(s)] \\
    &+(1-\gamma)\rE_{p_{\epol}^{(0)}}[f_{w}(s)] \\
    &= \int \delta(g,s')f_{w}(s')\mathrm{d}\lambda(s'),
\end{align*}
where we have $g(s)=p^{(\infty)}_{e,\gamma}(s)\{p^{(\infty)}_{e,\gamma}(s)/p_b(s)\}^{-1}w(s)$.

From the above, when $g(s)=p^{(\infty)}_{e,\gamma}(s)$, i.e.,  when $w(s)=p^{(\infty)}_{e,\gamma}(s)/p_{b}(s)$, $L(w,f_w)=0$. 

Conversely, $L(w,f_w)=0,\forall f_w\in L_2(S)$ means $\delta(g,s')=0$ from Riesz representation theorem. From the assumption, this means $w(s)=p^{(\infty)}_{e,\gamma}(s)/p_{b}(s)$. 
\end{myproof}
}

{\newedit 
\begin{myproof}[Proof of \cref{thm:bound_ratio}]

~
\paragraph*{\textbf{Consistency}}
We have $\hat \beta_{f_w} \stackrel{p}{\rightarrow} \beta^{*}$. We use \citet[Theorem 5.7]{VaartA.W.vander1998As}. We need to check two conditions: 
\begin{align*}
   &\sup_{\beta\in \Theta_{\beta}} |(\P_{N}-\P)[\Delta(s,a,s:\beta)] |\stackrel{p}{\rightarrow} 0,\\
   &\inf_{\|\beta-\beta_{*}\|>\epsilon}\|L(w(s:\beta,f_w))\|>0 
\end{align*}
for any $\epsilon>0$. The first condition is proved by \citet[Section 21.4]{DavidsonJamesE.H1994Slta} noting the first-order derivative of the map $\Theta_{\beta} \ni \beta \mapsto \Delta(s,a,s:\beta)$  is uniformly bounded. The second condition is proved by noting that 
\begin{align*}
    L(w(s:\beta),f_w)=0,\beta\in \Theta_{\beta}\iff \beta=\beta^{*}, 
\end{align*}
$\Theta_{\beta} \ni \beta \mapsto L(w(s:\beta)$ is continuous, and $\Theta_{\beta}$ is a compact space. 

\paragraph*{\textbf{Calculation of asymptotic variance}}

We calculate the asymptotic variance for general $f_{w}(s)$. We prove that the asymptotic MSE is given by
\begin{align}\label{eq:mainmain}
     \rE[\nabla_{\beta^{\top}}\Delta(s,a,s'; \beta)]^{-1}\rE[\Delta^{2}(s,a,s';\beta)]\{  \rE[\nabla_{\beta^{\top}}\Delta(s,a,s'; \beta)]^{-1}\}^{\top}|_{\beta^{*}}. 
\end{align}

For simplicity, we assume that $\beta$ is one-dimensional. Using mean value theorem, we have 
\begin{align}\label{eq:mainmain2}
    \sqrt{n}(\hat{\beta}_{f_w}-\beta^{*})=-\P_{n}[\nabla_{\beta^{\top}}\Delta(s,a,s'; \beta)]^{-1}|_{\beta^{\dagger}} \sqrt{n}\P_{n}[\Delta(s,a,s'; \beta)]|_{\beta^{*}},
\end{align}
where $\beta^{\dagger}$ is a value between $\hat{\beta}$ and $\beta^{*}$. The first term in right hand side of \eqref{eq:mainmain2} has the following property: 
\begin{align}
\label{eq:uni}
    \P_{n}[\nabla_{\beta^{\top}}\Delta(s,a,s'; \beta)]|_{\beta^{\dagger}}\stackrel{p}{\rightarrow} \rE[\nabla_{\beta^{\top}}\Delta(s,a,s'; \beta)]|_{\beta^{*}}.
\end{align}
This is proved by an uniform convergence condition from the fact the second-order derivative of the map $\Theta_{\beta}\ni \beta\mapsto \Delta(s,a,s';\beta)$ is uniformly bounded, and $\beta^{\dagger} \stackrel{p}{\rightarrow} \beta^{*}$.

Next, we calculate the second term in right hand size of \eqref{eq:mainmain2}. By CLT, we have 
\begin{align*}
    \sqrt{n}\P_{n}[\Delta(s,a,s'; \beta)]|_{\beta^{*}}\stackrel{d}{\rightarrow}\mathcal{N}(0,\var[\Delta(s,a,s'; \beta)]). 
\end{align*}

Finally, from Slutsky's theorem the asymptotic variance is 
\begin{align*}
   \rE[\nabla_{\beta^{\top}}\Delta(s,a,s'; \beta)]^{-1}\var[\Delta(s,a,s'; \beta)]\{\rE[\nabla_{\beta^{\top}}\Delta(s,a,s'; \beta)]^{\top}\}^{-1}|_{\beta^{*}}. 
\end{align*}

\end{myproof}
}

{\newedit
\begin{myproof}[Proof of \cref{thm:bound_ratio2}]

\begin{align*}
    \sqrt{n}(\hat{\rho}_{\mathrm{EIS}}-\rho^{\pi_e})&=  \bG_n[w(s;\hat{\beta}_{f_w})\eta(s,a)r]- \bG_n[w(s)\eta(s,a)r] \\ 
     &+\bG_n[w(s)\eta(s,a)r] + \\
     &+\sqrt{n}(\rE[w(s;\hat{\beta}_{f_w})\eta(s,a)r|\hat{\beta}_{f_w}]-\rho^{\pi_e}). 
\end{align*}
Here, from the standard argument, 
\begin{align*}
    &\sqrt{n}(\rE[w(s;\hat{\beta}_{f_w})\eta(s,a)r]-\rho^{\pi_e}) \\
    &=\sqrt{n}(\rE[w(s;\hat{\beta}_{f_w})\eta(s,a)r]-\rE[w(s;\beta^{*})\eta(s,a)r]) \\
     &=\sqrt{n}\{\rE[\nabla w(s;\beta^{*})\eta(s,a)r]|_{\beta^{*}}(\hat{\beta}_{f_w}- \beta^{*})+0.5(\hat{\beta}_{f_w}- \beta^{*})^{\top}\E[  \nabla_{\beta\beta^{\top}}\nabla w(s;\beta)\eta(s,a)r]|_{\beta^{\dagger}}(\hat{\beta}_{f_w}- \beta^{*})\} \tag{Taylor expansion} \\
    &=\rE[\nabla_{\beta^{\top}}w(s;\beta)\eta(s,a)r]\rE[\nabla_{\beta^{\top}}\Delta(s,a,s';\beta)]^{-1} \bG_n[\Delta(s,a,s';\beta)]|_{\beta^{*}}+\op(1). 
\end{align*}

Combining all together, we have
\begin{align*}
     \sqrt{n}(\hat{\rho}_{\mathrm{EIS}}-\rho^{\pi_e})= \bG_n[w(s;\beta)\eta(s,a)r+\rE[\nabla_{\beta^{\top}}w(s;\beta)\eta(s,a)r]\rE[\nabla_{\beta^{\top}}\Delta(s,a,s';\beta)]^{-1} \Delta(s,a,s';\beta)]|_{\beta^{*}}+\op(1). 
\end{align*}

\end{myproof}
}

{\newedit 
\begin{myproof}[Proof of \cref{thm:bound_q2,thm:bound_q2b}]
The proof is almost the same as the proof of \cref{thm:bound_ratio}. First, the asymptotic variance of $\hat \beta_{f_q}$ is 
\begin{align*}
    \E[f_q(s,a)\nabla_{\beta^{\top}}e_q(s,a,r,s';\beta)]^{-1}\E[f_q(s,a)f^{\top}_q(s,a)e^2_q(s,a,r,s';\beta) ]    \E[\nabla_{\beta}e_q(s,a,r,s';\beta)f^{\top}_q(s,a)]^{-1}|_{\beta^{*}} 
\end{align*}
This is equal to 
\begin{align*}
    \E[f_q(s,a)\nabla_{\beta^{\top}}m_q(s,a:\beta)]^{-1}\E[f_q(s,a)f^{\top}_q(s,a)v_q(s,a)]    \E[\nabla_{\beta}m_q(s,a:\beta)f^{\top}_q(s,a)]^{-1}|_{\beta^{*}} 
\end{align*}
Then, from \citet{matrix}, the lower bound is 
\begin{align*}
    \E[\nabla_{\beta}m(s,a;\beta)v^{-1}_q(s,a;\beta)\nabla_{\beta^{\top}}m(s,a;\beta)]|_{\beta^{*}}. 
\end{align*}
The asymptotic variance of the direct method is 
\begin{align*}
(1-\gamma)^2\E_{d_0}[\nabla_{\beta^{\top}} q(s,\epol)]  \E\left[ \frac{\otimes \{ \nabla_{\beta} \E[\gamma q(s',\pi;\beta)- q(s,a;\beta)|s,a]\}}{\mathrm{var}[r+\gamma q(s',\epol)|s,a] } \right]^{-1} \E_{d_0}[\nabla_{\beta} q(s,\epol)] .
\end{align*}
\end{myproof}

\begin{myproof}[Proof of \cref{eq:relation}]

Recall that the asymptotic variance of the direct method is 
\begin{align}\label{eq:variance}
(1-\gamma)^2\E_{d_0}[\nabla_{\beta^{\top}} q(s,\epol)]  \E\left[ \frac{\otimes \{ \nabla_{\beta} \E[\gamma q(s',\pi;\beta)- q(s,a;\beta)|s,a]\}}{\mathrm{var}[r+\gamma q(s',\epol)|s,a] } \right]^{-1} \E_{d_0}[\nabla_{\beta} q(s,\epol)] .
\end{align}
In addition, we have 
\begin{align*}
    (1-\gamma)\E_{d_0}[q(s_0,\epol)]=\E[-\gamma w(s,a)q(s',\epol)+w(s,a)q(s,a)]. 
\end{align*}
By differentiating this equation,  
\begin{align*}
    (1-\gamma)\nabla_{\beta} \E_{d_0}[q(s,\epol)]= \E[w(s,a)\nabla_{\beta}\E[-\gamma q(s',\epol)+q(s,a)|s,a]]. 
\end{align*}
According to CS-inequality, this immediately means that \eqref{eq:variance} is smaller than the efficient bound under the nonparametric model: 
\begin{align*}
    \E[w(s,a)^2\mathrm{var}[r+\gamma q(s',\epol)|s,a]]. 
\end{align*}
\end{myproof}
}

\begin{figure}[ht!]
\minipage{0.5\textwidth}%
  \includegraphics[width=\linewidth]{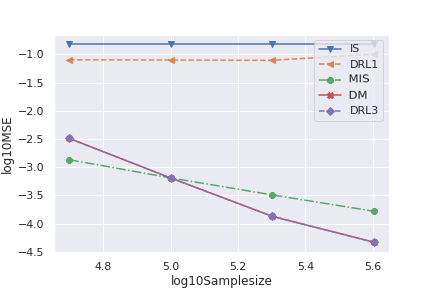}%
  \caption{Setting (1) with $\alpha=0.4$}\label{fig:4_1}%
\endminipage\hfill%
\minipage{0.5\textwidth}%
  \includegraphics[width=\linewidth]{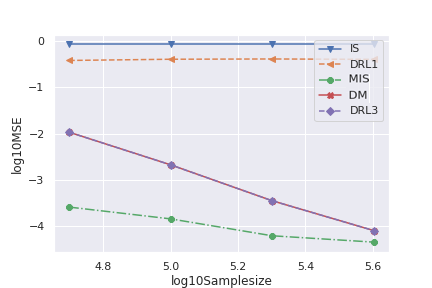}%
  \caption{Setting (1) with $\alpha=0.8$}\label{fig:8_1}%
\endminipage\\%
\minipage{0.5\textwidth}%
  \includegraphics[width=\linewidth]{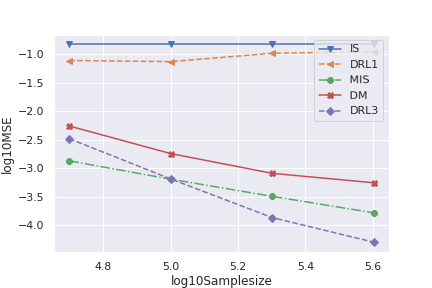}%
  \caption{Setting (2) with $\alpha=0.4$}\label{fig:4_2}%
\endminipage\hfill%
\minipage{0.5\textwidth}%
  \includegraphics[width=\linewidth]{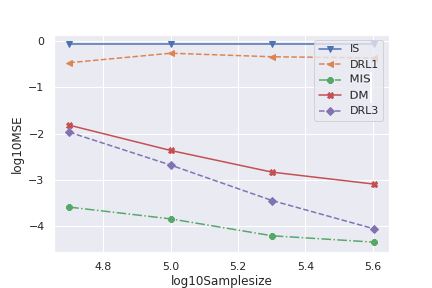}%
  \caption{Setting (2) with $\alpha=0.8$}\label{fig:8_2}%
\endminipage\\%
\minipage{0.5\textwidth}%
  \includegraphics[width=\linewidth]{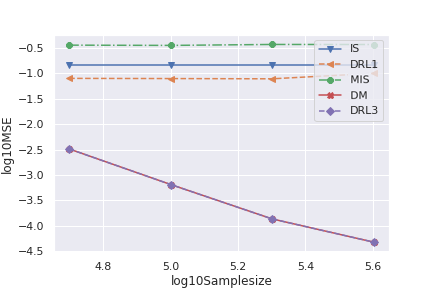}%
  \caption{Setting (3) with $\alpha=0.4$}\label{fig:4_3}%
\endminipage\hfill%
\minipage{0.5\textwidth}%
  \includegraphics[width=\linewidth]{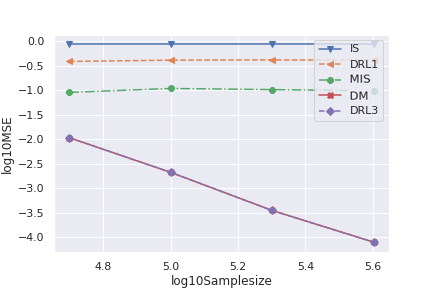}%
  \caption{Setting (3) with $\alpha=0.8$}\label{fig:8_3}%
\endminipage%
\end{figure}

\section{Additional Experimental Results}\label{appendix:extraresults}

Here, we provide additional results from the experiment in \cref{sec:numerical} with $\alpha=0.4,0.8$. 
The results are given in \cref{fig:4_1,fig:8_1,fig:4_2,fig:8_2,fig:4_3,fig:8_3}.
\end{APPENDIX}

\end{document}